\documentclass[11pt]{article}
\usepackage[final]{acl}
\usepackage{times}
\usepackage{latexsym}
\usepackage[T1]{fontenc}
\usepackage[utf8]{inputenc}
\usepackage{microtype}
\usepackage{inconsolata}
\usepackage{graphicx}
\usepackage{booktabs}
\usepackage{array}
\usepackage{amsmath}
\usepackage{amsthm}
\usepackage{amsfonts}
\usepackage{amssymb}
\usepackage{pifont}
\usepackage{enumitem}
\usepackage{float}
\usepackage{placeins}

\title{Fidelity Before Structure: Verbatim Chunks Beat Lossy Artifact\\ Extraction in Long-Conversation LLM Memory}

\author{Tao An \\
  Hawaii Pacific University \quad FIM Labs \\
  \texttt{tony@fim.ai}}

\hypersetup{pdftitle={Fidelity Before Structure: Verbatim Chunks Beat Lossy Artifact Extraction in Long-Conversation LLM Memory}, pdfauthor={Tao An}}
\begin{document}
\maketitle

\begin{abstract}
A growing class of conversational-memory systems compresses dialogue history into structured artifacts (extracted facts, decisions, or events). The premise is that distilled structure retrieves better than raw text. We test this premise with a controlled ablation. Within one fixed retrieval--rerank--reasoning pipeline, we swap only the stored representation (LLM-extracted typed artifacts versus verbatim conversation chunks) and hold the model, retriever, reranker, and judge constant. Verbatim chunks win by 15.9 points on LoCoMo (43.9\% vs.\ 28.0\%) and by 22.0 points on LongMemEval-S (67.4\% vs.\ 45.4\%). A semantic graph does not recover the gap, and six confound controls reproduce the effect. The mechanism is lossy distillation rather than structure per se. Accuracy tracks how much source text survives in the store, and the extracted-artifact pipeline does not beat naive RAG in overall accuracy (chunks do abstain worse, \S\ref{sec:limitations}). For the extraction designs we test, structured memory should augment verbatim text rather than replace it. Adding artifacts alongside chunks costs no detected accuracy, while substituting them forfeits the gap. Code and data are available at \url{https://github.com/tao-hpu/cog-canvas}.

\end{abstract}

\section{Introduction}
\label{sec:introduction}

\begin{quote}
\textit{``Please use type hints everywhere.''}
\end{quote}

This unremarkable instruction, dropped in turn 3 of a 50-turn programming discussion, is the kind of detail that long-conversation memory systems must preserve and routinely lose. Once the context window fills, truncation discards it outright. Summarization paraphrases it to ``the user prefers type hints,'' which deletes the quantifier \textit{everywhere}. In a controlled probe, summarization-based memory answers such constraint questions at 14.0\% exact match, versus 91.0\% when the verbatim wording is retrieved. This 77.0-point difference is the gap between \textbf{compression and fidelity} that the paper studies.

This fidelity problem has motivated a large family of memory systems that \textit{distill} raw dialogue into structured artifacts (extracted facts, decisions, events, or episodes) and retrieve over those~\citep{packer2023memgpt, chhikara2025mem0, zhou2025simple, evermemos2025, edge2024graphrag}. They differ in mechanism but share one premise. Distilled structure is taken to be a \emph{better} retrieval target than the raw text it replaces. Some systems state the premise directly. Mem0 credits its gains to ``capturing only the most salient facts in memory, rather than retrieving large chunk of original text''~\citep{chhikara2025mem0}. SeCom reports that compression ``can effectively serve as a denoising mechanism, enhancing memory retrieval accuracy''~\citep{pan2025secom}. The premise has not been cleanly isolated. Published comparisons are system-level and swap retrieval stacks, answerers, and judges along with the representation~\citep{packer2023memgpt, chhikara2025mem0}. Representation-level studies vary the granularity of stored \emph{text}~\citep{pan2025secom, zhu2024longmemeval} or sweep structure types across task suites~\citep{zeng2024structural}. These studies do not pair extraction against the text it replaces in one fixed pipeline. If extraction also \emph{discards} information, structure and fidelity pull in opposite directions. Which one wins is then an empirical question.

We answer it with a controlled ablation. Inside one fixed retrieval--rerank--reasoning pipeline, we change \emph{only} the stored representation (LLM-extracted artifacts versus verbatim conversation chunks). The backbone model, retriever, reranker, and LLM judge stay constant. The result inverts the premise. Verbatim chunks beat extracted artifacts by \textbf{15.9 points} on LoCoMo~\citep{maharana2024locomo} (43.9\% vs.\ 28.0\%) and by \textbf{22.0 points} on LongMemEval-S~\citep{zhu2024longmemeval} (67.4\% vs.\ 45.4\%). The ordering holds across every answerable question type. The extracted-artifact pipeline also does not outperform a naive RAG baseline in overall accuracy ($p{=}0.89$). Within this fixed pipeline, extraction does not pay off in \emph{either} memory-budget regime. When the conversation fits the window (LoCoMo, $\sim$26K tokens), artifacts fall 31.9 points below the free full-context ceiling. When it does not (LongMemEval-S, $\sim$115K tokens), chunk retrieval exceeds even the full-context baseline (67.4 vs.\ 60.6, different judge), while artifacts trail it. The scope of this result is narrow. It is a negative result about lossy \emph{extraction}. It does not claim that verbatim chunks are universally optimal, since they abstain worse (\S\ref{sec:limitations}). It is also confined to the stateless, training-free regime where representations swap cleanly (\S\ref{sec:experiments}).

The cause is \emph{lossy distillation}. Extraction throws away verbatim detail that chunks keep at no cost, and a semantic graph built over the artifacts does not recover the loss. Six confound controls reproduce the gap (graph disabled, zero chunk overlap, evidence budget matched, dense-only retrieval, session-level extraction granularity, and a sentence-verbatim store). The sentence-verbatim store attributes most of the loss to fidelity. Granularity accounts for 6.6pp of the 16.3pp gap in the gpt-4o-mini control batch. A 9-configuration GraphRAG sweep leaves its tuned ceiling far below verbatim chunks.

This result \emph{sharpens}, and does not contradict, concurrent work reporting that a carefully designed extraction can match strong baselines~\citep{zhou2025simple}. The units in that work are minimally abstractive and carry source provenance, which places them near the \emph{verbatim} end of the spectrum. One account covers both results. Retrieval accuracy tracks how far the stored representation departs from the source. We make the following contributions:

\begin{itemize}
    \item A \textbf{controlled-ablation finding}. Across two benchmarks and every answerable query type, verbatim chunks outperform \emph{lossy} typed-artifact extraction by 15.9 and 22.0 points under an otherwise identical pipeline. The artifact pipeline does not beat naive RAG overall ($p{=}0.89$; Tables~\ref{tab:locomo_results} and~\ref{tab:longmemeval_results}).

    \item A \textbf{mechanistic account}, \emph{lossy distillation}, that reconciles our negative result with concurrent positive findings on near-lossless extraction.

    \item \textbf{Evidence against the obvious objections}. We report six confound controls, a four-axis GraphRAG grid search (\S\ref{sec:graphrag_grid}), and a synthetic 77.0-point summarization-vs-verbatim probe. A stronger-extractor swap and a decontaminated-prompt re-run leave the gap intact, which makes a strawman extractor an unlikely explanation (Appendices~\ref{sec:error_analysis}, \ref{sec:prompts}).

    \item A \textbf{measured cost analysis} (Table~\ref{tab:cost_analysis}). The verbatim-chunk pipeline costs 65\% more per query. The two pipelines cost about the same per \emph{correct answer} (\$12.5 vs.\ \$11.9 per 1{,}000 correct answers), so the query cost that extraction saves is given back in accuracy.
\end{itemize}

\section{Related Work}

\looseness=-1 \textbf{Context window management} approaches include sparse attention~\citep{child2019sparsetransformer, beltagy2020longformer} and retrieval-augmented methods~\citep{borgeaud2022retro, izacard2022atlas}. We focus on \textit{what} to preserve rather than on extending capacity.

\textbf{Memory-augmented LLMs.} MemGPT~\citep{packer2023memgpt} pioneered OS-inspired virtual memory management and is now the production framework Letta~\citep{letta2024memory}. EMem~\citep{zhou2025simple} reaches strong accuracy with a training-free design whose event units are deliberately \emph{non-compressive}. These units are near-verbatim propositions with source-turn provenance (\S\ref{sec:discussion} and Appendix~\ref{sec:emem_control} reconcile its result with ours). Mem0~\citep{chhikara2025mem0} incrementally extracts and updates salient facts. A-Mem~\citep{xu2025amem} maintains an evolving note graph whose entries link and rewrite one another. SeCom~\citep{pan2025secom} keeps verbatim segments but varies their granularity. We reproduce each system's mechanism in our one fixed pipeline (\S\ref{sec:discussion}) as points on a single fidelity curve. Accuracy can then be attributed to \emph{what is stored} instead of to differences between whole stacks. Closest to our question, \citet{zeng2024structural} sweep chunks, triples, atomic facts, summaries, and mixed stores across six datasets. They find chunk-based and mixed stores strongest on dialogue understanding (LoCoMo), which is consistent with our chunks and union results. Appendix~\ref{sec:related_work_full} positions our isolation, mechanism, and replacement claim against that ranking. We test the premise in the \textit{training-free, stateless} drop-in regime that most memory libraries ship by default.

\textbf{Graph-based methods} like GraphRAG~\citep{edge2024graphrag} are designed for static documents; on LoCoMo the tuned GraphRAG baseline reaches 13.0\% (Appendix~\ref{sec:graphrag_grid}). \textbf{Summarization}~\citep{tang2021convosumm, chen2021dialogsum} is lossy compression; in our synthetic probe it trails verbatim retrieval by 77.0 points of exact match (\S\ref{sec:introduction}). See Appendix~\ref{sec:related_work_full} for extended discussion.

\section{Systems Under Test}
\label{sec:methodology}

\begin{figure*}[!t]
\centering
\includegraphics[width=\textwidth]{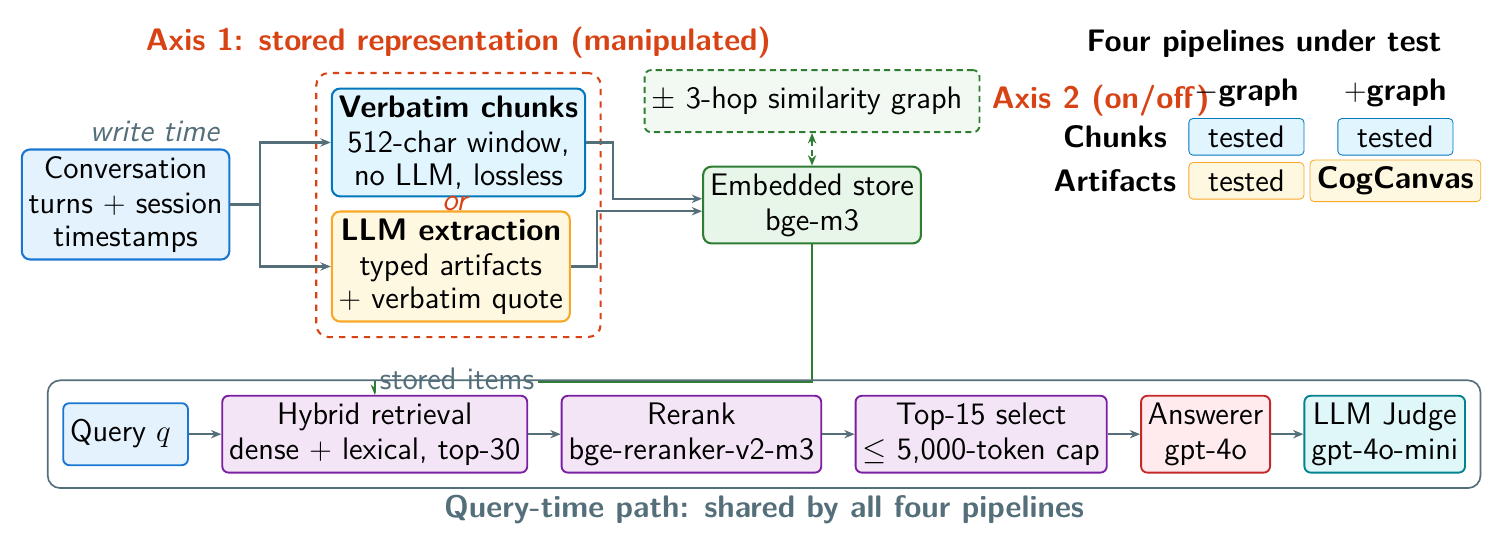}
\caption{The controlled ablation design. At write time the conversation passes through exactly one manipulated stage (Axis~1), either verbatim 512-character chunking \emph{or} LLM extraction into typed artifacts. The output enters a shared embedded store, optionally linked by a similarity graph (Axis~2). The query-time path (hybrid retrieval, reranking, token-budgeted context selection, answerer, judge) is identical across all four resulting pipelines. The Artifacts$+$Graph cell is the full CogCanvas system as originally engineered.}
\label{fig:architecture}
\end{figure*}

Our goal is to isolate one design variable. We factor the conversational-memory design space into two axes. The first is \emph{what is stored} (verbatim chunks vs.\ LLM-extracted artifacts). The second is \emph{whether a semantic graph links the stored items}. We instantiate all four combinations inside one pipeline, and everything else is shared. The artifacts-without-graph cell is reported in the component ablation (Table~\ref{tab:component_ablation}, an earlier batch). All four occupy the same \emph{stateless, training-free, infrastructure-free} cell (Table~\ref{tab:comparison}, Appendix~\ref{sec:additional_figures}). The Artifacts$+$Graph cell is the full CogCanvas system as originally engineered (the artifact-extraction memory pipeline under test; code at \url{https://github.com/tao-hpu/cog-canvas}). The chunk cells replace exactly one write-time stage of it (Figure~\ref{fig:architecture}).

\subsection{Shared Backbone}
At \textbf{write time}, the conversation is converted into a set of stored items according to the representation axis below, and every item receives a \textsf{bge-m3} embedding. At \textbf{query time}, a hybrid retriever fuses dense similarity with lexical keyword matching. The top-30 coarse candidates are re-scored by \textsf{bge-reranker-v2-m3}. The top 15 re-ranked items form the context under a 5{,}000-token cap that rarely binds. The budget-matched row of Table~\ref{tab:locomo_results} controls the resulting prompt asymmetry. A reasoning-aware (chain-of-thought) prompt then hands the selected items, with their turn indices and session timestamps, to the answerer (\textsf{gpt-4o}). All four pipelines execute this path with identical hyperparameters (\S\ref{sec:experiments}).

\subsection{Axis 1: Stored Representation}

\paragraph{Verbatim chunks.}
The transcript, with explicit turn and session-timestamp markers, is segmented by a sliding window of 512 characters with 100-character overlap, aligned to line breaks. These parameters match the standalone RAG baseline exactly. Each chunk is stored as-is. No LLM runs at write time, and no information is discarded. Write-time confound controls remove the overlap (control~ii of \S\ref{sec:experiments}) and the graph (control~i, Axis~2).

\paragraph{Extracted artifacts.}
The conversation is passed through an extraction LLM (\textsf{gpt-4o-mini}) in batches of 40 turns, which is the interval of the rolling compression. The extractor emits typed artifacts $o = (\tau, c, q, s, e, t, \gamma)$. The type $\tau$ is drawn from an eight-type inventory. Five types are task-oriented (\textsc{Decision}, \textsc{Todo}, \textsc{KeyFact}, \textsc{Reminder}, \textsc{Insight}), and three personal/social types were added for conversational data (\textsc{PersonAttribute}, \textsc{Event}, \textsc{Relationship}). The remaining fields are the content string $c$, a \emph{verbatim grounding quote} $q$ excerpted from the source message, source role $s$, embedding $e$, turn index $t$, and extraction confidence $\gamma$. The design instantiates the standard premise of extraction-based memory. The dialogue is distilled into its retrievable essence, and a quote anchors each artifact to limit hallucination. A second \emph{gleaning} pass~\citep{guo2024lightrag} re-scans each batch for missed entities (pronoun referents, omitted subjects, implicit causal and temporal expressions). Its measured contribution is negligible (Limitations, Appendix~\ref{sec:error_analysis}). The store contains whatever the extractor deemed worth keeping, and this is the property under test.

\subsection{Axis 2: Graph Expansion}
Stored items of either representation are linked at write time by identical rules. \emph{Reference edges} are added when embedding cosine similarity exceeds $\theta_{\text{ref}}{=}0.5$ or lemmatized keyword overlap exceeds $0.3$ (Jaccard). \emph{Causal edges} link type- and time-compatible pairs (similarity above $\theta_{\text{causal}}{=}0.45$, predecessor within a 20-turn window). A temporal heuristic also links recent \textsc{KeyFact}s and \textsc{Reminder}s to subsequent \textsc{Decision}s. At query time, retrieved items are expanded along graph edges, up to three hops, before reranking. Within the artifact pipeline, this expansion lifts \emph{retrieval} recall from 25.8\% to 71.8\% on LoCoMo10 (\autoref{tab:retrieval_recall} in Appendix~\ref{sec:retrieval_quality}). This recall gain makes its null effect on final \emph{answer} accuracy over chunks (Table~\ref{tab:locomo_results}) the more diagnostic.

\section{Experiments}
\label{sec:experiments}

We evaluate the four pipelines and external baselines first on a synthetic information-retention benchmark and then on two public long-conversation benchmarks. LoCoMo~\citep{maharana2024locomo} has 10 multi-session conversations of about 300 turns each, with 1{,}986 questions in five categories. LongMemEval-S~\citep{zhu2024longmemeval} is LongMemEval's 500-question split, with a $\sim$115K-token multi-session history per question. The synthetic benchmark exercises the pipeline end to end. It shows that artifacts win when planted facts align with the extraction ontology. The benchmark sections show that this alignment does not hold on open-ended dialogue, which is consistent with the inversion on LoCoMo and LongMemEval.

\subsection{Experimental Setup}

\paragraph{Benchmark Design.}
We construct synthetic multi-turn conversations in which specific facts are ``planted''. Each fact is inserted as one utterance at a designated turn. It belongs to one of three categories of the extraction ontology of \S\ref{sec:methodology} (\textsc{KeyFact}, \textsc{Decision}, \textsc{Reminder}). Two variants differ in how the questions recombine the planted facts. \textbf{Standard} ($N{=}50$) has 50 turns, rolling compression at turn 40, and single-hop recall. \textbf{Multi-hop Reasoning} ($N{=}50$) asks questions that require causal or impact-based reasoning across multiple planted facts, e.g.\ ``Why was X decided?''. Querying happens after the compression boundary discards short-term context. This isolates long-range recall from short-context lookup (walked example: \autoref{fig:synthetic_walked_example}, Appendix~\ref{sec:additional_figures}).

\paragraph{Why a synthetic probe.}
The synthetic benchmark serves two purposes. Its Standard variant is a positive control for the extraction arm. Every planted fact has a known location and a known type, so a failure to recover it would point to the extractor and not to the data. The benchmark also fixes a condition under which the premise of structured memory should hold, namely facts that arrive already shaped to the extraction schema. This condition gives the LoCoMo and LongMemEval results a reference point.

\paragraph{Conversation benchmarks.}
LoCoMo~\citep{maharana2024locomo} contains 10 two-speaker conversations (294 turns on average, spread over multiple dated sessions) and 1{,}986 questions in five categories. The categories are multi-hop (282), temporal (321), open-domain (96), single-hop (841), and adversarial questions whose correct response is a refusal (446). Categories 1--3 (699 questions) form the main protocol on which every system and control is compared. Categories 4 and 5 are reported separately as held-out. LongMemEval-S~\citep{zhu2024longmemeval} pairs each of 500 questions with its own multi-session history of about 115K tokens. It groups the questions into information extraction (150), multi-session reasoning (121), knowledge updates (72), temporal reasoning (127), and abstention (30). In both benchmarks the conversations are machine-generated and human-verified, and the questions are human-written. Both are English; a native-Chinese probe is in Appendix~\ref{sec:perltqa}.

\paragraph{Evaluation Metrics.}
The standard benchmark reports \textbf{Recall Rate} (token-level fuzzy match $\geq 0.80$) and \textbf{Exact Match}. The multi-hop benchmark reports \textbf{Keyword Coverage}, its \textbf{Pass Rate} ($\mathrm{KC}\geq 0.80$), and $\mathrm{KC}$ on causal/impact questions (definitions in Appendix~\ref{sec:reproducibility}). For LoCoMo and LongMemEval we report \textbf{LLM Judge accuracy}. The judge, \textsf{gpt-4o-mini} at temperature 0, assigns a binary CORRECT/INCORRECT label given $(q,g,a)$ (prompt in Appendix~\ref{sec:reproducibility}). It is the headline metric because gold answers admit paraphrases that substring EM penalizes unfairly. Judge--human agreement on a stratified 100-question subset is 95\% ($\kappa{=}0.897$). The agreement is near-symmetric and unbiased across representations (Appendix~\ref{sec:judge_validation}).

\paragraph{Reproducibility.}
All experiments share one configuration (full list in Appendix~\ref{sec:reproducibility}). The extractor is \textsf{gpt-4o-mini} (temp.\ 0.1). The answerer is \textsf{gpt-4o} for LoCoMo and LongMemEval and \textsf{gpt-4o-mini} for the synthetic probe, and every system shares it within a benchmark (temp.\ 0.0). The judge is \textsf{gpt-4o-mini} (temp.\ 0). The embedder \textsf{bge-m3} and the reranker \textsf{bge-reranker-v2-m3} both run locally. The final top-$k{=}15$ comes from a reranked pool of 30 under a 5{,}000-token cap that rarely binds. Rolling compression runs every 40 turns and retains the 5 most-recent turns.

\paragraph{Baselines.}
Reflecting our \textit{training-free} focus, we compare against five baselines; heavily optimized systems such as EverMemOS are discussed in Appendix~\ref{sec:related_work_full}. \textbf{Truncation} keeps the most recent 5 turns. \textbf{Summarization} compresses the history with the answerer-tier model every 40 turns and keeps the recent 5 turns verbatim, which is the core hierarchical mechanism of MemGPT~\citep{packer2023memgpt}. \textbf{RAG} uses 512-char chunks, 100-char overlap, and top-10 dense retrieval. \textbf{RAG+Rerank} uses the same chunks and re-scores the top-20 candidates with \texttt{bge-reranker-v2-m3}, the same reranker CogCanvas employs. This makes it \textbf{our primary fair baseline}. It inherits RAG's top-20$\to$10 depth, while the swap pipelines use top-30$\to$15. \textbf{GraphRAG} is the official Microsoft implementation~\citep{edge2024graphrag}. We report it at the best cell of a two-tier, 4-axis, 9-configuration grid search (tuned ceiling 13.0\% on LoCoMo, drift mode; default 11.2\%; Appendix~\ref{sec:graphrag_grid}). Stateful systems (e.g.\ Letta, 74.0\%~\citep{letta2024memory}) sit outside this stateless comparison (Appendix~\ref{sec:stateful_discussion}).

\paragraph{Fair-comparison disclosure.}
\looseness=-1 GraphRAG underwent a 9-configuration sweep, and the summarizer underwent prompt engineering. CogCanvas hyperparameters were fixed on the synthetic benchmark. Its extraction prompt contains LoCoMo-style few-shot examples, which favor the artifact pipeline and so bias \emph{against} our finding. A decontaminated-prompt re-run leaves the result unchanged (28.5\% vs.\ 28.0\%, $p{=}0.82$; Appendix~\ref{sec:prompts}). RAG and RAG+Rerank use deliberately standard settings. Further tuning of these baselines could only widen the artifact deficit (Appendices~\ref{sec:fair_comparison}, \ref{sec:rag_baseline}).

\subsection{Main Results (Standard)}

Table~\ref{tab:main_results} (Appendix~\ref{sec:additional_figures}) shows the methods clustering by storage fidelity rather than architecture (Figure~\ref{fig:recall_comparison}, Appendix~\ref{sec:additional_figures}). Every verbatim-preserving store sits at or above 89\% exact match. This includes the artifact pipeline (97.5 recall / 93.0 EM), whose grounding quotes make it lossless on this probe by design. The lossy stores collapse (summarization 19.0 / 14.0, truncation 12.5 / 7.0). Each planted fact is a single, clean, extractable statement, so extraction gives up nothing here.

\subsection{Multi-hop Reasoning Results}

The artifact pipeline achieves an \textbf{81.0\%} pass rate (Table~\ref{tab:multi_hop_results} and Figure~\ref{fig:multihop_results}, Appendix~\ref{sec:additional_figures}). This is far ahead of RAG (55.5\%) and the chunk pipelines (24.5--25.0\%). The inversion is informative in both directions. First, it indicates that the extraction machinery works as intended. A broken extractor could not dominate its own ablation by 56 points, a margin far above the 1--3.7pp cross-batch drift of Table~\ref{tab:provenance}. Second, it delimits where the premise holds. Every planted fact is authored as exactly one artifact, and each question recombines two or three of them. Open-ended conversation does not offer this pre-shaping. The spread \emph{within} the verbatim family (24.5--55.5 under shared chunking) shows that retrieval-stack interactions dominate this probe. It is a fragile basis for claims about retrieval architecture, and we base none on it.

\FloatBarrier
\subsection{Evaluation on LoCoMo}

We evaluate on \textbf{LoCoMo} and report LLM-Judge accuracy (\S\ref{sec:experiments}; the de-facto metric of memory-system evaluations~\citep{chhikara2025mem0, letta2024memory}). Every system shares one retrieval--rerank backbone and one answerer (gpt-4o), and only the stored representation changes.

\begin{table*}[t]
\centering
\small
\setlength{\tabcolsep}{5pt}
\begin{tabular}{@{}lcccc@{}}
\toprule
\textbf{Stored representation} & \textbf{Overall} & \textbf{Multi-hop} & \textbf{Temporal} & \textbf{Open-domain} \\
\midrule
\multicolumn{5}{@{}l}{\textit{Reference ceiling (no compression; conversations fit the context window)}} \\
\quad Full context (entire conversation in prompt) & 59.9 & 57.4 & 62.3 & 59.4 \\
\midrule
\multicolumn{5}{@{}l}{\textit{Verbatim conversation chunks}} \\
\quad Chunks, no graph & \textbf{43.9} & 35.5 & \textbf{50.2} & 47.9 \\
\quad Chunks $+$ graph & 43.1 & 36.5 & 47.7 & 46.9 \\
\quad Chunks, zero overlap (control ii) & 42.8 & 34.8 & 49.2 & 44.8 \\
\quad Chunks, dense-only retrieval (control iv) & 41.8 & 34.0 & 47.7 & 44.8 \\
\midrule
\multicolumn{5}{@{}l}{\textit{Mixed store}} \\
\quad Chunks $\cup$ artifacts $+$ graph (union) & 42.5 & 35.8 & 48.3 & 42.7 \\
\midrule
\multicolumn{5}{@{}l}{\textit{LLM-extracted artifacts}} \\
\quad Artifacts $+$ graph (CogCanvas) & 28.0 & 22.3 & 31.2 & 34.4 \\
\quad Artifacts $+$ graph, budget-matched ($k{=}60$) & 29.2 & 25.2 & 30.2 & 37.5 \\
\quad Artifacts $+$ graph, dense-only (control iv) & 27.5 & 21.6 & 30.5 & 34.4 \\
\midrule
\multicolumn{5}{@{}l}{\textit{External baselines}} \\
\quad RAG $+$ Rerank & 30.5 & \textbf{41.1} & 16.5 & 45.8 \\
\quad RAG ($k{=}10$) & 27.6 & 34.8 & 14.0 & \textbf{52.1} \\
\quad GraphRAG (drift)$^\dagger$ & 13.0 & 12.8 & 7.5 & 32.3 \\
\quad Summarization (rolling) & 5.7 & 5.3 & 0.9 & 22.9 \\
\quad Truncation (recent-5; no retrieval) & 5.0 & 3.9 & 0.3 & 24.0 \\
\midrule
\textit{$\Delta$ Chunks $-$ Artifacts} & \textbf{+15.9} & +13.1 & +19.0 & +13.5 \\
\bottomrule
\end{tabular}

\vspace{2pt}
{\footnotesize $^\dagger$GraphRAG uses gpt-4o-mini as the answerer (its indexing pipeline is coupled to the model); all other rows use gpt-4o. The row reports the best of a two-tier, 9-configuration grid over $\{$chunk\_size, max\_gleanings, community\_level, search\_method$\}$ (the \textit{drift} mode; Appendix~\ref{sec:graphrag_grid}). Under a matched gpt-4o-mini answerer, Chunks-no-graph scores 41.9\%, which is $+$28.9pp over GraphRAG, so the gap is not a model artifact.}
\caption{\textbf{LoCoMo (Cat 1--3; 10 conversations, 699 questions; LLM-Judge via GPT-4o-mini).} Within one fixed retrieval--rerank--reasoning pipeline we swap only the stored representation; all rows share one retriever (bge-m3), reranker (bge-reranker-v2-m3), answerer (gpt-4o), and judge. The full-context row is a reference ceiling and not a memory system. LoCoMo conversations ($\sim$26K tokens) fit gpt-4o's window, so nothing is compressed. \textbf{Bold} marks the best memory-system entry per column (baseline rows eligible). Verbatim chunks beat LLM-extracted artifacts by 15.9 points overall and on every question type. The gap is significant (McNemar exact test: 154 vs.\ 43 discordant pairs, $p<10^{-15}$; cluster bootstrap resampling whole conversations: 95\% CI $[13.0, 18.4]$pp), while artifacts vs.\ naive RAG is not (106 vs.\ 103, $p=0.89$). The graph, overlap, budget, dense-only, and union controls are analyzed in the text.}
\label{tab:locomo_results}
\end{table*}

Table~\ref{tab:locomo_results} reports the representation swap (per-type breakdown: Figure~\ref{fig:locomo_results}, Appendix~\ref{sec:additional_figures}). The verbatim-chunk pipeline reaches \textbf{43.9\%} overall. Replacing those chunks with LLM-extracted artifacts (the CogCanvas representation) drops accuracy to \textbf{28.0\%}. This is a 15.9-point loss inside an otherwise identical pipeline. The loss has the same sign on every question type (multi-hop $+$13.1pp, temporal $+$19.0pp, open-domain $+$13.5pp), and no question type favors extraction (temporal mechanism: Appendix~\ref{sec:why_temporal}). A closed-book control (same answerer, \emph{no} retrieved context) scores 2.3\% ($n{=}699$). Accuracy therefore reflects retrieved text rather than parametric priors or contamination.

The extracted-artifact pipeline (28.0\%) also does not outperform naive RAG (27.6\%). Extraction adds a write-time LLM stage (\S\ref{sec:cost_analysis}) and gains no accuracy for it. Six confound controls rule out retrieval and budget artifacts and bound the share due to granularity. Table~\ref{tab:controls_summary} summarizes controls (ii)--(v) together with the vendor, prompt, and judge controls. \emph{(i)~Disabling the graph} changes nothing for chunks (43.1\% vs.\ 43.9\%). \emph{(ii)~Removing lexical overlap}, so that adjacent chunks share no tokens, preserves the advantage (42.8\%). Chunks therefore do not win by overlapping the gold answer. \emph{(iii)~Matching the evidence budget} addresses a prompt-length asymmetry. Artifacts are short ($\sim$30--60 tokens) and chunks are long ($\sim$130), so the shared $k{=}15$ gives chunks nearly twice the prompt (2{,}127 vs.\ 1{,}191 tokens). Raising artifacts to $k{=}60$ under a 2{,}000-token cap recovers only 1.2pp (28.0\%$\to$29.2\%; $p{=}0.36$). The gap is still 14.7pp ($p{<}10^{-13}$; bootstrap 95\% CI $[12.0, 17.2]$). \emph{(iv)~Dense-only retrieval}, which drops the lexical component, leaves the gap at 14.3pp (41.8\% vs.\ 27.5\%). \emph{(v)~Session-level extraction} runs the extractor once per session instead of once per 40-turn batch. It scores 29.2\% ($p{=}0.57$ vs.\ the batch default), a gap of 14.7pp, so the loss does not depend on where the extraction windows are cut. \emph{(vi)~Granularity vs.\ fidelity} is tested with a sentence-verbatim store, which keeps the source verbatim at artifact-scale units. It beats artifacts by 5.7pp at matched granularity and budget. It trails chunks by 6.6pp at the chunk token budget ($k{=}38$, 2{,}140 tokens). Granularity thus accounts for 6.6pp of that batch's 16.3pp chunks-minus-artifacts gap (the $k{=}38$ row is a later batch), and most of the rest is fidelity (Appendix~\ref{sec:sentence_control}). The deficit is informational. It lies in what the store contains rather than in how the prompt budget is allocated. Tuned GraphRAG (13.0\%, Appendix~\ref{sec:graphrag_grid}), Summarization (5.7\%), and Truncation (5.0\%) trail far behind. The pair replicates under a second vendor for answerer and extractor (53.9\% vs.\ 33.2\%; Appendix~\ref{sec:vendor_control}).

\begin{table}[t]
\centering
\footnotesize
\setlength{\tabcolsep}{3pt}
\begin{tabular}{@{}p{0.50\columnwidth}ccc@{}}
\toprule
\textbf{Condition} & \textbf{Chunks} & \textbf{Artif.} & \textbf{Gap} \\
\midrule
Headline (Table~\ref{tab:locomo_results}) & 43.9 & 28.0 & $+$15.9 \\
Zero-overlap chunks & 42.8 & 28.0 & $+$14.7 \\
Budget-matched artifacts ($k{=}60$) & 43.9 & 29.2 & $+$14.7 \\
Dense-only retrieval & 41.8 & 27.5 & $+$14.3 \\
Session-level extraction & 43.9 & 29.2 & $+$14.7 \\
Decontaminated extraction prompt & 43.9 & 28.5 & $+$15.5 \\
Claude answerer and extractor & 53.9 & 33.2 & $+$20.7 \\
Seven judges, five vendors & \multicolumn{3}{c}{gap $+$13.0 to $+$17.0} \\
\midrule
LongMemEval-S (Table~\ref{tab:longmemeval_results}) & 67.4 & 45.4 & $+$22.0 \\
\bottomrule
\end{tabular}
\caption{Robustness of the chunks-minus-artifacts gap on LoCoMo categories 1--3 (LLM-Judge \%, 699 questions; last row: LongMemEval-S, 500 questions). Each row changes one element of the headline comparison; the Claude row changes both LLM roles. Gaps use unrounded counts. The budget-matched and session-level rows are separate runs that coincide at 204 of 699. Details: \S\ref{sec:experiments} and Appendices~\ref{sec:prompts}, \ref{sec:vendor_control}, \ref{sec:judge_validation}.}
\label{tab:controls_summary}
\end{table}

\paragraph{Union storage: replacement causes the loss.}
If the loss came from \emph{replacing} text, a union store should recover chunk-level accuracy, and it comes within 1.4 points. Indexing artifacts \emph{alongside} the chunks yields 42.5\% against 43.9\% for chunks alone. McNemar detects no significant difference ($p{=}0.39$), which does not establish equivalence. The union is far above artifacts alone (28.0\%; $p{<}10^{-12}$). Structure may coexist with verbatim text at no detected accuracy cost, while substituting it forfeits 15.9 points.

\paragraph{The held-out categories widen the gap, except where abstention is the skill.}
\looseness=-1 Categories 4--5 complete the LoCoMo picture. On held-out \emph{category 4} (single-hop factual recall; 841 questions), the gap \emph{widens} to $+$33.3pp, with chunks at 82.6\% and artifacts at 49.3\% (McNemar $p{<}10^{-60}$). These answers live in surface detail, which is what extraction discards. \emph{Category 5} (adversarial; 446 unanswerable questions, scored with a LongMemEval-style abstention template) is the one type that favors artifacts (15.0\% vs.\ 6.5\% for chunks; $-$8.5pp, $p{=}1.9{\times}10^{-6}$). LongMemEval's 30 abstention questions show no such difference (46.7\% vs.\ 43.3\%), but 30 questions cannot rule one out. We therefore claim the advantage only on LoCoMo. The verbatim store surfaces the trap fact, and the answerer asserts it. The sparser artifact store more often returns nothing assertable. This rewards storing less rather than extracting better, and both stores remain far from competent refusal. Keeping artifacts alongside the chunks restores only part of the artifact store's refusals. In a later batch where all three arms share one store, the union refuses on 7.0\% of category-5 questions, against 4.7\% for chunks and 10.3\% for artifacts. No significant difference from chunks is detected ($p{=}0.11$), and the union stays below artifacts ($p{=}0.04$; Table~\ref{tab:union_abstention}, Appendix~\ref{sec:retriever_robustness}).

\paragraph{Anchoring the absolute level.}
Published LoCoMo numbers run higher. Mem0 self-reports 66.9, with a 72.9 full-context ceiling~\citep{chhikara2025mem0}, but these numbers are computed over categories 1--4 with different answerers and judges. Re-computed over that category mix, our same-batch chunks score 65.1\% (1{,}002/1{,}540), close to Mem0's self-reported 66.9 (artifacts 39.7\%). Our full-context ceiling rises to 74.4\%, broadly consistent with Mem0's 72.9. The lower headline reflects the harder 1--3 subset rather than a weak harness.

\FloatBarrier
\subsection{Cross-Benchmark Validation (LongMemEval)}

To test whether the finding generalizes beyond LoCoMo, we evaluate on \textbf{LongMemEval-S}~\citep{zhu2024longmemeval} (the 500-question split defined in \S\ref{sec:experiments}; $\sim$115K tokens per history). The benchmark decomposes memory into five abilities: \textit{Information Extraction} (IE), \textit{Multi-Session Reasoning} (MSR), \textit{Knowledge Updates} (KU), \textit{Temporal Reasoning} (TR), and \textit{Abstention} (Abs., correctly refusing unanswerable questions). We follow the benchmark's official evaluation (GPT-4o-mini judge with question-type-specific prompts). As before, every system shares one retrieval--rerank backbone and answerer (gpt-4o), and only the stored representation changes. GraphRAG and Summarization are omitted. They would re-index or re-compress each question's $\sim$115K-token haystack 500 times, and on LoCoMo they already trail every retrieval method by 15--38 points.

\begin{table}[t]
\centering
\footnotesize
\setlength{\tabcolsep}{2pt}
\begin{tabular}{@{}p{0.165\columnwidth}ccccccc@{}}
\toprule
\textbf{Store} & \textbf{IE} & \textbf{MSR} & \textbf{KU} & \textbf{TR} & \textbf{Abs} & \textbf{All} & \textbf{T-Avg} \\
\textit{n} & \textit{150} & \textit{121} & \textit{72} & \textit{127} & \textit{30} & \textit{500} & --- \\
\midrule
Chunks & 84.7 & \textbf{62.8} & \textbf{77.8} & 50.4 & 46.7 & \textbf{67.4} & 64.5 \\
\;Union & 80.7 & 58.7 & 76.4 & \textbf{51.2} & 36.7 & 64.6 & 60.7 \\
RAG+RR & 85.3 & 61.2 & 72.2 & 38.6 & 70.0 & 64.8 & \textbf{65.5} \\
RAG & \textbf{86.7} & 58.7 & 72.2 & 37.0 & 70.0 & 64.2 & 64.9 \\
Artifacts & 50.7 & 34.7 & 69.4 & 36.2 & 43.3 & 45.4 & 46.9 \\
\;$k{=}60$ & 50.7 & 36.4 & 69.4 & 40.9 & 50.0 & 47.4 & 49.5 \\
Trunc. & 13.3 & 0.0 & 13.9 & 0.8 & \textbf{93.3} & 11.8 & 24.3 \\
\midrule
$\Delta$ C$-$A & +34.0 & +28.1 & +8.3$^{\mathrm{ns}}$ & +14.2 & +3.3$^{\mathrm{ns}}$ & \textbf{+22.0} & +17.6 \\
\bottomrule
\end{tabular}
\caption{\textbf{LongMemEval-S (500 questions; LLM-Judge via GPT-4o-mini).} The same representation swap as Table~\ref{tab:locomo_results}, on a second benchmark. Rows: Chunks = chunks, no graph; Union = chunks $\cup$ artifacts $+$ graph; RAG$+$RR = RAG $+$ Rerank; Artifacts = artifacts $+$ graph (CogCanvas); $k{=}60$ = its budget-matched variant; Trunc.\ = truncation, recent-5 turns; $\Delta$ C$-$A = chunks minus artifacts. IE = information extraction, MSR = multi-session reasoning, KU = knowledge updates, TR = temporal reasoning, Abs = abstention; Overall is question-weighted, T-Avg the unweighted category mean. Truncation's 93.3 abstention is degenerate, since it answers almost nothing. McNemar exact tests: chunks vs.\ artifacts $p<10^{-15}$ (152 vs.\ 42 discordant pairs; bootstrap 95\% CI $[17.0, 27.0]$pp); artifacts trail even naive RAG, $p<10^{-11}$ (48 vs.\ 142). $\Delta$s are computed from unrounded counts; $^{\mathrm{ns}}$ marks per-type deltas that are not significant (KU: 15 vs.\ 9 discordant, $p{=}0.31$; Abs: 4 vs.\ 3, $p{=}1.0$, a one-question margin).}
\label{tab:longmemeval_results}
\end{table}

Table~\ref{tab:longmemeval_results} reproduces the pattern on a second benchmark. Chunks reach \textbf{67.4\%} overall and artifacts \textbf{45.4\%}, a 22.0-point loss. Chunks lead on all five question types, significantly on three: IE ($+$34.0pp, $p{<}10^{-9}$), MSR ($+$28.1pp, $p{<}10^{-5}$), and temporal ($+$14.2pp, $p{=}0.011$). Knowledge updates are where structured state-tracking was most expected to help. There the lead is positive but non-significant ($+$8.3pp, $n{=}72$, $p{=}0.31$), so extraction does not win even on this type. The abstention lead ($+$3.3pp) is a one-question margin over $n{=}30$. The loss follows the representation rather than the retrieval stack. Both verbatim pipelines, chunks (67.4\%) and naive RAG (64.2\%), sit $\sim$20 points above artifacts, while the reranker moves accuracy by under a point. Both LoCoMo controls replicate. Budget-matched $k{=}60$ artifacts recover 2.0pp (47.4\%, gap 20.0pp, $p{<}10^{-12}$). A union store scores 64.6\%, which is 2.8pp below chunks with no significant difference detected ($p{=}0.14$) and $+$19.2pp over artifacts alone. Adding artifacts to verbatim text thus costs no detected accuracy. The losses concentrate where surface detail matters most (IE 50.7 vs.\ 84.7, MSR 34.7 vs.\ 62.8). These types depend on the wording and per-turn provenance that extraction normalizes away. As an external anchor, the benchmark authors report GPT-4o at 60.6\% reading the full $\sim$115k-token haystack~\citep{zhu2024longmemeval}. Chunk retrieval at 67.4\% exceeds that full-context baseline (different judge), so the representation that artifacts lose to is not a weak one.

\textbf{The finding concerns extraction; it does not show that chunks are universally best.} Chunks' one clear weakness is abstention (46.7\% vs.\ naive RAG's 70.0\% on LME's 30 abstention questions, $p{=}0.039$). The weakness replicates with far more power on LoCoMo's 446 adversarial questions (\S\ref{sec:experiments}). Handed no relevant evidence, the pipeline still answers instead of refusing (see Limitations). On every answerable question type across both benchmarks, however, the extracted-artifact representation is dominated by at least one verbatim representation. It never significantly beats either verbatim pipeline overall, and no tested carve-out shows extraction winning (case study: Appendix~\ref{sec:verbatim_case_study}). The ordering holds cross-lingually. On native-Chinese PerLTQA, both verbatim stores beat the typed-artifact store by $\sim$47--50 points (Appendix~\ref{sec:perltqa}).

\subsection{Cost and Efficiency Analysis}
\label{sec:cost_analysis}

We re-ran all instrumented methods on LoCoMo10 (categories~1--3) with per-call token logging, plus a full re-run with a gpt-4o-mini answerer (Table~\ref{tab:cost_analysis}; USD at OpenAI list prices of 2026-05-27; embedding and reranking run locally at zero marginal cost). There are three findings (Appendix~\ref{sec:cost_appendix}). First, with a gpt-4o answerer, extraction is \emph{not} the dominant cost. It accounts for \$0.13 of the artifact pipeline's \$2.33 total, and the gpt-4o answerer dominates under every architecture. Second, extraction \emph{does} buy cheaper generation, with shorter prompts and 5.2$\times$ shorter completions than RAG+Rerank. Per correct answer, however, the advantage nearly disappears (\$11.9 per 1{,}000 for artifacts vs.\ \$12.5 for chunks and \$16.4 for RAG+Rerank). Third, the ordering depends on the answerer tier. Re-priced at gpt-4o-mini (Total$^\dagger$), the fixed extraction cost makes the artifact pipeline the more expensive one (\$0.27 vs.\ \$0.23 for the run). In sum, with a gpt-4o answerer extraction lowers cost per query by about 40\% and leaves cost per correct answer about where it was, at 15.9 points lower accuracy.

\begin{table}[t]
\centering
\footnotesize
\setlength{\tabcolsep}{1.8pt}
\begin{tabular}{@{}lrrrrr@{}}
\toprule
\textbf{Method} & \textbf{Extract} & \textbf{Gen} & \textbf{Total} & \textbf{Total$^\dagger$} & \textbf{Ratio} \\
\midrule
Truncation          & \$0.00 & \$1.25 & \$1.25 & \$0.09 & 13.2$\times$ \\
RAG                 & \$0.00 & \$3.50 & \$3.50 & \$0.23 & 15.3$\times$ \\
RAG+Rerank          & \$0.00 & \$3.50 & \$3.50 & \$0.23 & 15.3$\times$ \\
Chunks (no graph)$^{\S}$ & \$0.00 & \$3.84 & \$3.84 & \$0.23 & 16.7$\times$ \\
Summarization       & \$0.67 & \$1.74 & \$2.41 & \$0.17 & 14.2$\times$ \\
Artifacts (CogCanvas)$^{\S}$ & \$0.13 & \textbf{\$2.20} & \textbf{\$2.33} & \$0.27 & 8.7$\times$ \\
GraphRAG$^\ddagger$ & ---    & ---    & ---    & ---    & ---          \\
\bottomrule
\end{tabular}
\caption{Measured cost on LoCoMo10 (699 questions, 2,938 indexed turns; USD). \textbf{Extract} and \textbf{Gen} aggregate \texttt{response.usage} tokens per phase. \textbf{Total$^\dagger$} is the gpt-4o-mini answerer tier and \textbf{Ratio} is gpt-4o\,/\,mini. \textbf{Bold} marks the cheapest retrieval-augmented method. $^{\S}$Priced from the token logs of the two Table~\ref{tab:locomo_results} runs; Total$^\dagger$ re-prices the same tokens. A separately instrumented artifact run with deeper retrieval (top-20, four hops) costs \$2.92. $^\ddagger$Not instrumented. Details in Appendix~\ref{sec:cost_appendix}.}
\label{tab:cost_analysis}
\end{table}

\subsection{Discussion}
\label{sec:discussion}

\paragraph{Extraction commits to relevance before the question exists.}
An extractor decides at \emph{write time} what survives, and whatever it discards cannot be recovered at \emph{read time}. Verbatim chunks defer the decision to query time, when the question can guide selection. The hardest-hit categories are those whose answers live in surface detail (IE $-$34.0pp, MSR $-$28.1pp). Of the 58 diagnosable chunk-pass/artifact-fail questions, 29.3\% concern facts absent from every stored artifact (write-time loss). Another 39.7\% concern facts that survive only in a raw-context field and never as a typed fact (extraction failure; Appendix~\ref{sec:error_analysis}). A temporal query such as \emph{``When did Caroline join the LGBTQ support group?''} is answered from the dated source turn that the chunk store keeps verbatim. The artifact store answers it only if the extractor logged that date. Both classes are errors \emph{avoidable by not extracting}, and only the first is irreversible loss. A paraphrase-tolerant re-test replaces token matching with an LLM judge validated against the gold evidence. It finds an extracted artifact sufficient for 2 of 42 validated questions (Appendix~\ref{sec:error_analysis}). The synthetic multi-hop probe shows the converse. When facts \emph{are} pre-shaped to the schema, artifacts win (\S\ref{sec:experiments}).

\paragraph{Neither graph expansion nor reranking is the missing ingredient.}
Graph expansion connects stored items but cannot regenerate content that is absent from all of them. It changes nothing over chunks (43.1\% vs.\ 43.9\%), and GraphRAG never exceeds 13.0\% across its 9-configuration sweep. Its community-summarization layer dissolves the per-turn timestamps that conversational temporal queries depend on (Appendix~\ref{sec:graphrag_grid}). Reranking is equally marginal. The reranker (bge-reranker-v2-m3) moves overall accuracy by $+$0.6pp on LongMemEval and $+$2.9pp on LoCoMo, well below the representation effect.

\paragraph{Fidelity, not structure, governs the ordering.}
\looseness=-1 We order this paper's designs by the distance between the stored representation and the source text. From closest to farthest, they are verbatim chunks, near-verbatim event units with provenance~\citep{zhou2025simple}, typed artifacts, and community summaries. This order reproduces the observed accuracy ordering (the EDU point is measured at matched token budget in Appendix~\ref{sec:emem_control}). Isolating fidelity supports a causal reading. Keeping a random fraction of each verbatim chunk's tokens, with the pipeline fixed, drives accuracy down monotonically (retain $1.0{\to}0.2$: $43.9{\to}35.5{\to}23.9{\to}14.6{\to}9.0$; each step McNemar $p{<}10^{-4}$; Figure~\ref{fig:fidelity_dial}). We also reproduce the defining mechanisms of four memory systems in the same pipeline, and they fall on this axis in fidelity order. Extraction stores (Mem0, A-Mem) fall $\sim$22 points below chunks ($p{<}10^{-25}$), while verbatim-preserving SeCom stays high (a within-tier segmentation effect). The \emph{official} Mem0 package reproduces this ordering, which makes a strawman an unlikely explanation (Appendix~\ref{sec:mem0_anchor}). The design question is then \emph{where} to add structure. Annotating verbatim text (e.g.\ Zep's episode-grounded graph~\citep{rasmussen2025zep}) preserves a lossless path to source, and replacing it severs that path (Appendices~\ref{sec:stateful_discussion}, \ref{sec:why_temporal}).

\section{Conclusion}

We reported a controlled negative result for \emph{lossy} extraction-based memory. With only the stored representation swapped, LLM-distilled artifacts lose to verbatim chunks by 15.9 points on LoCoMo and 22.0 on LongMemEval-S. The damage tracks fidelity. Granularity accounts for under half of the gap, and graph tuning for none of it. The verbatim store has one unrepaired weakness, abstention. On unanswerable questions, chunks refuse less often than naive RAG on LongMemEval and than the artifact store on LoCoMo, and each of three post-hoc refusers trades answerable accuracy for refusal (\S\ref{sec:limitations}). The finding concerns \emph{lossy extraction}. It is not a case against structure. Structure should \emph{augment} verbatim text, not replace it.

\section{Limitations}
\label{sec:limitations}

The result is a negative finding about lossy extraction. It does not show that verbatim chunks are universally best. Appendix~\ref{sec:limitations_full} gives each limitation in full, with the supporting controls and statistics.

\textbf{Scope.} We tested one artifact schema in one stateless, training-free pipeline family. Neither a stronger extractor (tested on a 3-conversation subset) nor near-verbatim event units~\citep{zhou2025simple} closes the gap, and removing the gleaning pass changes LoCoMo accuracy by 1.6pp (Appendices~\ref{sec:error_analysis} and~\ref{sec:emem_control}). Five confound controls each leave at least 14.3pp of the 15.9pp gap, and granularity accounts for under half of it. An untested design may still escape the pattern. Within the schema we extracted over 40-turn batches and whole sessions. Segment-level~\citep{pan2025secom} and incremental~\citep{chhikara2025mem0} extraction of the same schema remain untested, as does fine-tuned memory. A stateful-agent probe reproduces the direction of the result (Appendix~\ref{sec:stateful_probe}). Figures self-reported by stateful systems come from other evaluation stacks, and we use them only as external anchors (Appendix~\ref{sec:stateful_discussion}).

\textbf{Abstention.} Verbatim chunks refuse too rarely. On LoCoMo's 446 adversarial questions they abstain correctly 6.5\% of the time, against 15.0\% for the artifact pipeline ($p{=}1.9{\times}10^{-6}$). On LongMemEval abstention they trail naive RAG by 23.3pp ($n{=}30$, $p{=}0.039$). The three repair mechanisms we tested raise refusal only by lowering accuracy on answerable questions (Appendix~\ref{sec:retriever_robustness}). A refusal mechanism without that cost remains open.

\textbf{Extraction's one win.} Artifacts lead on our synthetic multi-hop probe (81.0, against 55.5 for RAG and 24.5--25.0 for the chunk pipelines), where each planted fact maps to exactly one artifact. The lead does not transfer to the multi-hop questions of LoCoMo or LongMemEval, where chunks lead by 13.1pp and 28.1pp.

\textbf{Benchmarks.} Both headline benchmarks are English chat-style QA over model-generated dialogue. A native-Chinese probe reproduces the ordering (PerLTQA, Appendix~\ref{sec:perltqa}). Memory uses beyond QA, such as proactive recall, personalization, and summarization-style review, remain untested.

\textbf{Models and retrieval.} Headline runs use one retrieval stack and OpenAI models. The ordering holds under two other retriever families and a weaker answerer (Appendix~\ref{sec:retriever_robustness}). It also holds under a non-OpenAI answerer and extractor (chunks 53.9\%, artifacts 33.2\%; Appendix~\ref{sec:vendor_control}). Other answerer and extractor families remain untested.

\textbf{Variance.} Each table entry is a single run at temperature 0. Three end-to-end re-runs of the headline LoCoMo cells differ by at most 0.4pp. The unpinned \texttt{gpt-4o} endpoint drifts by 1--3.7pp across months, and both arms shift in parallel (Appendix~\ref{sec:provenance}).

\section{Ethics Statement}

This work evaluates memory systems on public benchmarks of fictional, machine-generated conversations (LoCoMo, LongMemEval, PerLTQA). No real user data, personally identifying information, or human subjects are involved. The judge-validation study (Appendix~\ref{sec:judge_validation}) was annotated by the author and one independent volunteer annotator. Both labeled only model outputs on public benchmarks (no personal or human-subjects data). The volunteer participated with informed consent and received no monetary compensation, contributing as an acknowledged collaborator (see Acknowledgments). Because the task involves only public model outputs and no human-subjects data, we judged that it does not require IRB review. All model access used commercial APIs under their terms of service. We report the full token and dollar cost of the experiments (\S\ref{sec:cost_analysis}) to make the computational footprint transparent. The paper's central contribution is a cautionary, negative result about extraction-based memory and does not add a new capability. We foresee no dual-use risk beyond those already present in retrieval-augmented LLM systems.

\paragraph{Scientific artifacts and licensing.} All evaluation data are public research benchmarks, used consistently with their intended non-commercial research use. LoCoMo~\citep{maharana2024locomo} and PerLTQA~\citep{du2024perltqa} are released under CC~BY-NC~4.0, and LongMemEval~\citep{zhu2024longmemeval} under the MIT license. We redistribute none of the dataset files. Our released run artifacts contain model outputs and scores. They quote the benchmark questions and gold answers they were scored against, which remain under the original licenses. Our independently constructed synthetic planted-fact benchmark, the experiment harness, and run artifacts are released at \url{https://github.com/tao-hpu/cog-canvas} (MIT for code, CC~BY~4.0 for our own data and outputs). The release contains the pipeline and experiment-harness code, the synthetic-benchmark generator, and all extraction/answerer/judge prompts (including the decontaminated-prompt variant). It also contains cached model outputs, scoring and significance-testing scripts, and per-run manifests sufficient to reproduce every table and figure.

\paragraph{Use of AI assistance.} We used an AI programming assistant for software scaffolding and routine refactoring of the experiment harness, and an LLM for light copy-editing (grammar and concision) of author-written prose. All research questions, experimental design, analysis, and claims are the author's own. No text was generated wholesale by an LLM, and the author verified every reported number against the run artifacts.

\section*{Acknowledgments}
We thank Jin Jiang for the independent blind re-annotation of the judge-validation study (Appendix~\ref{sec:judge_validation}).

\bibliography{references}

\begin{thebibliography}{27}
\providecommand{\natexlab}[1]{#1}

\bibitem[{Beltagy et~al.(2020)Beltagy, Peters, and
  Cohan}]{beltagy2020longformer}
Iz~Beltagy, Matthew~E Peters, and Arman Cohan. 2020.
\newblock Longformer: The long-document transformer.
\newblock \emph{arXiv preprint arXiv:2004.05150}.

\bibitem[{Borgeaud et~al.(2022)Borgeaud, Mensch, Hoffmann, Cai, Rutherford,
  Millican, Van Den~Driessche, Lespiau, Damoc, Clark
  et~al.}]{borgeaud2022retro}
Sebastian Borgeaud, Arthur Mensch, Jordan Hoffmann, Trevor Cai, Eliza
  Rutherford, Katie Millican, George Van Den~Driessche, Jean-Baptiste Lespiau,
  Bogdan Damoc, Aidan Clark, and 1 others. 2022.
\newblock Improving language models by retrieving from trillions of tokens.
\newblock In \emph{International conference on machine learning}, pages
  2206--2240. PMLR.

\bibitem[{Chen et~al.(2023)Chen, Wong, Chen, and Tian}]{chen2023extending}
Shouyuan Chen, Sherman Wong, Liangjian Chen, and Yuandong Tian. 2023.
\newblock Extending context window of large language models via positional
  interpolation.
\newblock \emph{arXiv preprint arXiv:2306.15595}.

\bibitem[{Chen et~al.(2021)Chen, Liu, Chen, and Zhang}]{chen2021dialogsum}
Yulong Chen, Yang Liu, Liang Chen, and Yue Zhang. 2021.
\newblock {DialogSum}: A real-life scenario dialogue summarization dataset.
\newblock In \emph{Findings of the Association for Computational Linguistics:
  ACL-IJCNLP 2021}, pages 5062--5074.

\bibitem[{Chhikara et~al.(2025)Chhikara, Khant, Aryan, Singh, and
  Yadav}]{chhikara2025mem0}
Prateek Chhikara, Dev Khant, Saket Aryan, Taranjeet Singh, and Deshraj Yadav.
  2025.
\newblock Mem0: Building production-ready {AI} agents with scalable long-term
  memory.
\newblock \emph{arXiv preprint arXiv:2504.19413}.

\bibitem[{Child et~al.(2019)Child, Gray, Radford, and
  Sutskever}]{child2019sparsetransformer}
Rewon Child, Scott Gray, Alec Radford, and Ilya Sutskever. 2019.
\newblock Generating long sequences with sparse transformers.
\newblock \emph{arXiv preprint arXiv:1904.10509}.

\bibitem[{Du et~al.(2024)Du, Wang, Zhao, Liang, Wang, Zhong, Wang, and
  Wong}]{du2024perltqa}
Yiming Du, Hongru Wang, Zhengyi Zhao, Bin Liang, Baojun Wang, Wanjun Zhong,
  Zezhong Wang, and Kam-Fai Wong. 2024.
\newblock {PerLTQA}: A personal long-term memory dataset for memory
  classification, retrieval, and fusion in question answering.
\newblock In \emph{Proceedings of the 10th SIGHAN Workshop on Chinese Language
  Processing (SIGHAN-10)}, pages 152--164.

\bibitem[{Edge et~al.(2024)Edge, Trinh, Cheng, Bradley, Chao, Mody, Truitt,
  Metropolitansky, Ness, and Larson}]{edge2024graphrag}
Darren Edge, Ha~Trinh, Newman Cheng, Joshua Bradley, Alex Chao, Apurva Mody,
  Steven Truitt, Dasha Metropolitansky, Robert~Osazuwa Ness, and Jonathan
  Larson. 2024.
\newblock From local to global: A graph {RAG} approach to query-focused
  summarization.
\newblock \emph{arXiv preprint arXiv:2404.16130}.

\bibitem[{{EverMind AI}(2025)}]{evermemos2025}
{EverMind AI}. 2025.
\newblock {EverMemOS}: A memory operating system for {AI} agents.
\newblock \url{https://github.com/EverMind-AI/EverMemOS}.
\newblock README as of 2025-12-09; repository since renamed EverOS.

\bibitem[{Fabbri et~al.(2021)Fabbri, Rahman, Rizvi, Wang, Li, Mehdad, and
  Radev}]{tang2021convosumm}
Alexander~R. Fabbri, Faiaz Rahman, Imad Rizvi, Borui Wang, Haoran Li, Yashar
  Mehdad, and Dragomir Radev. 2021.
\newblock {ConvoSumm}: Conversation summarization benchmark and improved
  abstractive summarization with argument mining.
\newblock In \emph{Proceedings of the 59th Annual Meeting of the Association
  for Computational Linguistics and the 11th International Joint Conference on
  Natural Language Processing (Volume 1: Long Papers)}, pages 6866--6880.

\bibitem[{Guo et~al.(2024)Guo, Xia, Yu, Ao, and Huang}]{guo2024lightrag}
Zirui Guo, Lianghao Xia, Yanhua Yu, Tu~Ao, and Chao Huang. 2024.
\newblock {LightRAG}: Simple and fast retrieval-augmented generation.
\newblock \emph{arXiv preprint arXiv:2410.05779}.

\bibitem[{Izacard et~al.(2022)Izacard, Lewis, Lomeli, Hosseini, Petroni,
  Schick, Dwivedi-Yu, Joulin, Riedel, and Grave}]{izacard2022atlas}
Gautier Izacard, Patrick Lewis, Maria Lomeli, Lucas Hosseini, Fabio Petroni,
  Timo Schick, Jane Dwivedi-Yu, Armand Joulin, Sebastian Riedel, and Edouard
  Grave. 2022.
\newblock Atlas: Few-shot learning with retrieval augmented language models.
\newblock \emph{arXiv preprint arXiv:2208.03299}.

\bibitem[{Lee et~al.(2024)Lee, Chen, Furuta, Canny, and
  Fischer}]{lee2024readagent}
Kuang-Huei Lee, Xinyun Chen, Hiroki Furuta, John Canny, and Ian Fischer. 2024.
\newblock A human-inspired reading agent with gist memory of very long
  contexts.
\newblock In \emph{International Conference on Machine Learning}, pages
  26396--26415.

\bibitem[{{Letta AI}(2025)}]{letta2024memory}
{Letta AI}. 2025.
\newblock Benchmarking {AI} agent memory: Is a filesystem all you need?
\newblock \url{https://www.letta.com/blog/benchmarking-ai-agent-memory}.
\newblock Published 2025-08-12; accessed 2026-11-14.

\bibitem[{Maharana et~al.(2024)Maharana, Lee, Tulyakov, Bansal, Barbieri, and
  Fang}]{maharana2024locomo}
Adyasha Maharana, Dong-Ho Lee, Sergey Tulyakov, Mohit Bansal, Francesco
  Barbieri, and Yuwei Fang. 2024.
\newblock Evaluating very long-term conversational memory of {LLM} agents.
\newblock In \emph{Proceedings of the 62nd Annual Meeting of the Association
  for Computational Linguistics (Volume 1: Long Papers)}, pages 13851--13870.
\newblock ArXiv:2402.17753.

\bibitem[{Packer et~al.(2023)Packer, Wooders, Lin, Fang, Patil, Stoica, and
  Gonzalez}]{packer2023memgpt}
Charles Packer, Sarah Wooders, Kevin Lin, Vivian Fang, Shishir~G. Patil, Ion
  Stoica, and Joseph~E. Gonzalez. 2023.
\newblock {MemGPT}: Towards {LLMs} as operating systems.
\newblock \emph{arXiv preprint arXiv:2310.08560}.

\bibitem[{Pan et~al.(2025)Pan, Wu, Jiang, Luo, Cheng, Li, Yang, Lin, Zhao, Qiu,
  and Gao}]{pan2025secom}
Zhuoshi Pan, Qianhui Wu, Huiqiang Jiang, Xufang Luo, Hao Cheng, Dongsheng Li,
  Yuqing Yang, Chin-Yew Lin, H.~Vicky Zhao, Lili Qiu, and Jianfeng Gao. 2025.
\newblock {SeCom}: On memory construction and retrieval for personalized
  conversational agents.
\newblock In \emph{Proceedings of the International Conference on Learning
  Representations (ICLR)}.

\bibitem[{Rasmussen et~al.(2025)Rasmussen, Paliychuk, Beauvais, Ryan, and
  Chalef}]{rasmussen2025zep}
Preston Rasmussen, Pavlo Paliychuk, Travis Beauvais, Jack Ryan, and Daniel
  Chalef. 2025.
\newblock \href {https://arxiv.org/abs/2501.13956} {Zep: A temporal knowledge
  graph architecture for agent memory}.
\newblock \emph{Preprint}, arXiv:2501.13956.

\bibitem[{Wang et~al.(2023)Wang, Dong, Cheng, Liu, Yan, Gao, and
  Wei}]{wang2023longmem}
Weizhi Wang, Li~Dong, Hao Cheng, Xiaodong Liu, Xifeng Yan, Jianfeng Gao, and
  Furu Wei. 2023.
\newblock Augmenting language models with long-term memory.
\newblock In \emph{Advances in Neural Information Processing Systems},
  volume~36, pages 74530--74543.

\bibitem[{Wu et~al.(2025)Wu, Wang, Yu, Zhang, Chang, and
  Yu}]{zhu2024longmemeval}
Di~Wu, Hongwei Wang, Wenhao Yu, Yuwei Zhang, Kai-Wei Chang, and Dong Yu. 2025.
\newblock {LongMemEval}: Benchmarking chat assistants on long-term interactive
  memory.
\newblock In \emph{Proceedings of the International Conference on Learning
  Representations (ICLR)}.
\newblock ArXiv:2410.10813.

\bibitem[{Wu et~al.(2021)Wu, Ouyang, Ziegler, Stiennon, Lowe, Leike, and
  Christiano}]{wu2021recursively}
Jeff Wu, Long Ouyang, Daniel~M Ziegler, Nisan Stiennon, Ryan Lowe, Jan Leike,
  and Paul Christiano. 2021.
\newblock Recursively summarizing books with human feedback.
\newblock \emph{arXiv preprint arXiv:2109.10862}.

\bibitem[{Xu et~al.(2025)Xu, Liang, Mei, Gao, Tan, and Zhang}]{xu2025amem}
Wujiang Xu, Zujie Liang, Kai Mei, Hang Gao, Juntao Tan, and Yongfeng Zhang.
  2025.
\newblock {A-Mem}: Agentic memory for {LLM} agents.
\newblock \emph{arXiv preprint arXiv:2502.12110}.

\bibitem[{Yasunaga et~al.(2021)Yasunaga, Ren, Bosselut, Liang, and
  Leskovec}]{yasunaga2021qagnn}
Michihiro Yasunaga, Hongyu Ren, Antoine Bosselut, Percy Liang, and Jure
  Leskovec. 2021.
\newblock {QA-GNN}: Reasoning with language models and knowledge graphs for
  question answering.
\newblock In \emph{Proceedings of the 2021 Conference of the North American
  Chapter of the Association for Computational Linguistics: Human Language
  Technologies}, pages 535--546.

\bibitem[{Zeng et~al.(2024)Zeng, Fang, Liu, and Meng}]{zeng2024structural}
Ruihong Zeng, Jinyuan Fang, Siwei Liu, and Zaiqiao Meng. 2024.
\newblock \href {https://arxiv.org/abs/2412.15266} {On the structural memory of
  {LLM} agents}.
\newblock \emph{Preprint}, arXiv:2412.15266.

\bibitem[{Zhang et~al.(2022)Zhang, Bosselut, Yasunaga, Ren, Liang, Manning, and
  Leskovec}]{zhang2022greaselm}
Xikun Zhang, Antoine Bosselut, Michihiro Yasunaga, Hongyu Ren, Percy Liang,
  Christopher~D Manning, and Jure Leskovec. 2022.
\newblock {GreaseLM}: Graph {REAS}oning enhanced language models for question
  answering.
\newblock In \emph{International Conference on Learning Representations}.

\bibitem[{Zhong et~al.(2024)Zhong, Guo, Gao, Ye, and
  Wang}]{zhong2024memorybank}
Wanjun Zhong, Lianghong Guo, Qiqi Gao, He~Ye, and Yanlin Wang. 2024.
\newblock {MemoryBank}: Enhancing large language models with long-term memory.
\newblock In \emph{Proceedings of the AAAI Conference on Artificial
  Intelligence}, volume~38, pages 19724--19731.

\bibitem[{Zhou and Han(2025)}]{zhou2025simple}
Sizhe Zhou and Jiawei Han. 2025.
\newblock \href {https://arxiv.org/abs/2511.17208} {A simple yet strong
  baseline for long-term conversational memory of {LLM} agents}.
\newblock \emph{Preprint}, arXiv:2511.17208.

\end{thebibliography}

\appendix

\section{Additional Figures and Tables}
\label{sec:additional_figures}

\begin{table}[t]
\centering
\footnotesize
\setlength{\tabcolsep}{3pt}
\begin{tabular}{@{}p{0.55\columnwidth}ccc@{}}
\toprule
\textbf{Method} & \textbf{Recall} & \textbf{Exact} & \textbf{Cmplx.} \\
\midrule
\multicolumn{4}{@{}l}{\textit{Lossy stores}} \\
\quad Truncation & 12.5\% & 7.0\% & Low \\
\quad Summarization & 19.0\% & 14.0\% & Low \\
\quad GraphRAG & 83.5\% & 70.0\% & High \\
\midrule
\multicolumn{4}{@{}l}{\textit{Verbatim stores}} \\
\quad RAG (k=10) & 93.5\% & 89.5\% & Low \\
\quad Chunks, no graph & 92.0\% & 90.5\% & Low \\
\quad Chunks $+$ graph & 92.5\% & 91.0\% & Low \\
\quad Artifacts $+$ graph (CogCanvas) & \textbf{97.5\%} & \textbf{93.0\%} & Low \\
\bottomrule
\end{tabular}
\caption{Information retention on the standard synthetic benchmark (a fixed 50-conversation subset, 50 turns, 4 planted facts each; rolling compression every 40 turns; gpt-4o-mini answerer). The two chunks rows were measured in the 2026-06 batch. All other rows are the 2025-12 batch under the identical protocol (Table~\ref{tab:provenance}). Same-configuration cross-batch drift on this harness is 1--3.7pp with direction preserved (Table~\ref{tab:provenance}). The summarization-vs-chunks exact-match gap (14.0\% vs.\ 91.0\%), the 77.0-point figure quoted in \S1, is therefore a cross-batch difference, far larger than any observed drift. Cmplx.\ = qualitative implementation complexity (write-time machinery and infrastructure). Truncation keeps the most recent 5 turns (\S\ref{sec:experiments}).}
\label{tab:main_results}
\end{table}

\begin{table}[t]
\centering
\footnotesize
\setlength{\tabcolsep}{2.5pt}
\begin{tabular}{@{}p{0.36\columnwidth}cccc@{}}
\toprule
\textbf{Method} & \textbf{Pass} & \textbf{KW} & \textbf{Causal} & \textbf{Impact} \\
\midrule
Summarization & 0.0\% & 42.5\% & 48.3\% & 35.8\% \\
Chunks $+$ graph & 24.5\% & 60.8\% & 50.6\% & 71.0\% \\
Chunks, no graph & 25.0\% & 60.8\% & 50.6\% & 71.0\% \\
GraphRAG & 40.0\% & 64.8\% & 62.5\% & 67.0\% \\
RAG+Rerank & 45.0\% & 68.2\% & 64.2\% & 73.3\% \\
RAG & 55.5\% & 74.8\% & 68.8\% & 79.5\% \\
Artifacts $+$ graph (CogCanvas) & \textbf{81.0\%} & \textbf{90.2\%} & \textbf{87.5\%} & \textbf{92.6\%} \\
\midrule
\textit{$\Delta$ Artifacts $-$ RAG} & +25.5 & +15.4 & +18.7 & +13.1 \\
\bottomrule
\end{tabular}
\caption{Multi-hop causal reasoning on the synthetic probe (50 conversations, rolling compression every 40 turns, gpt-4o-mini answerer). The two chunks rows and the RAG+Rerank row are the 2026-06 batch. All other rows, including both terms of the $\Delta$ row, are the 2025-12 batch under the identical protocol (Table~\ref{tab:provenance}). The RAG configuration re-measures at 52.5\% in 2026-06 (Appendix~\ref{sec:rag_baseline}). KW = Keyword Coverage. This is the one setting in the paper where the artifact pipeline wins, and it is extraction's best case (see text). The advantage inverts on the public benchmarks (Tables~\ref{tab:locomo_results} and~\ref{tab:longmemeval_results}).}
\label{tab:multi_hop_results}
\end{table}

This appendix collects the design-space positioning table and the visualizations of the main results referenced from the body of the paper.

\begin{figure*}[t]
\centering
\includegraphics[width=\textwidth]{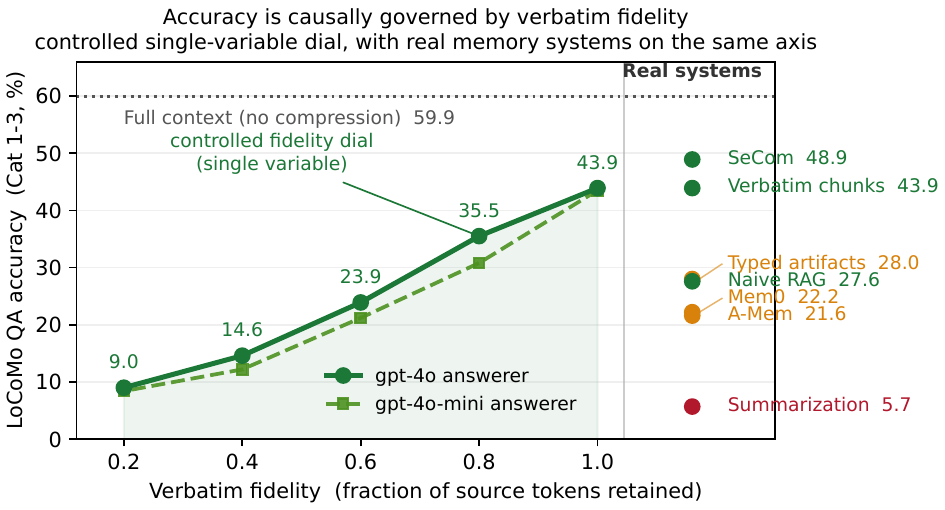}
\caption{\textbf{A controlled fidelity dial} (\S\ref{sec:discussion}). Keeping a random fraction of each verbatim chunk's tokens (solid line; pipeline held fixed) drives LoCoMo Cat~1--3 accuracy down monotonically. At retain$=$1.0 the dial reproduces the verbatim-chunks anchor (43.9\%). The dashed line repeats the dial under a gpt-4o-mini answerer (accuracy $43.5{\to}30.8{\to}21.2{\to}12.2{\to}8.4$ as retain goes $1.0{\to}0.2$). The near-coincident curve is consistent with a collapse that belongs to the stored representation and does not depend on the strong answerer model. Real memory systems (right column) land on the same accuracy axis in fidelity order. Extraction-based Mem0/A-Mem fall in the dial's low-fidelity regime, verbatim-preserving SeCom/chunks fall in the high one, and full context (no compression) sits above all. SeCom's edge over fixed chunks is a segmentation effect \emph{within} the verbatim tier and does not reverse the fidelity order.}
\label{fig:fidelity_dial}
\end{figure*}

\begin{table}[t]
\centering
\footnotesize
\setlength{\tabcolsep}{3pt}
\begin{tabular}{@{}p{0.45\columnwidth}ccccc@{}}
\toprule
\textbf{Method} & \textbf{TF} & \textbf{State} & \textbf{Infra} & \textbf{P\&P} & \textbf{Incr.} \\
\midrule
Truncation & \ding{51} & \ding{55} & \ding{55} & \ding{51} & \ding{51} \\
Summarization & \ding{51} & \ding{55} & \ding{55} & \ding{51} & \ding{51} \\
RAG & \ding{51} & \ding{55} & \ding{55} & \ding{51} & \ding{51} \\
RAG+Rerank & \ding{51} & \ding{55} & \ding{55} & \ding{51} & \ding{51} \\
GraphRAG & \ding{51} & \ding{55} & \ding{55} & \ding{51} & \ding{55} \\
Letta~\citep{letta2024memory}     & \ding{51} & \ding{51} & \ding{51} & \ding{55} & \ding{51} \\
EverMemOS~\citep{evermemos2025}   & \ding{51} & \ding{51} & \ding{51} & \ding{55} & \ding{51} \\
EMem~\citep{zhou2025simple}       & \ding{51} & \ding{55} & \ding{55} & \ding{51} & \ding{51} \\
Pipelines under test (\S\ref{sec:methodology}) & \ding{51} & \ding{55} & \ding{55} & \ding{51} & \ding{51} \\
\bottomrule
\end{tabular}
\caption{Design-space positioning across five operational axes. \textbf{TF} = training-free (no fine-tuning required). \textbf{State} = maintains memory state across queries. \textbf{Infra} = requires persistent infrastructure (e.g.\ PostgreSQL, Docker). \textbf{P\&P} = plug-and-play (drop-in for any answerer LLM, no API contract). \textbf{Incr.} = supports incremental updates without re-indexing the full conversation. All four pipelines under test (chunks/artifacts $\times$ graph on/off) occupy the same \textit{stateless, training-free, infrastructure-free} cell. Our comparison therefore swaps representations within one corner of this space and does not promote any one system. GraphRAG's lack of incremental support stems from its community-summarization stage, which must re-cluster the entity graph when new turns arrive. Letta's stateful design places it outside our stateless evaluation pipeline (\S\ref{sec:experiments}, Appendix~\ref{sec:stateful_discussion}). We cite its 74.0\% LoCoMo result only as an external anchor.}
\label{tab:comparison}
\end{table}

\begin{figure*}[!t]
\centering
\includegraphics[width=\textwidth]{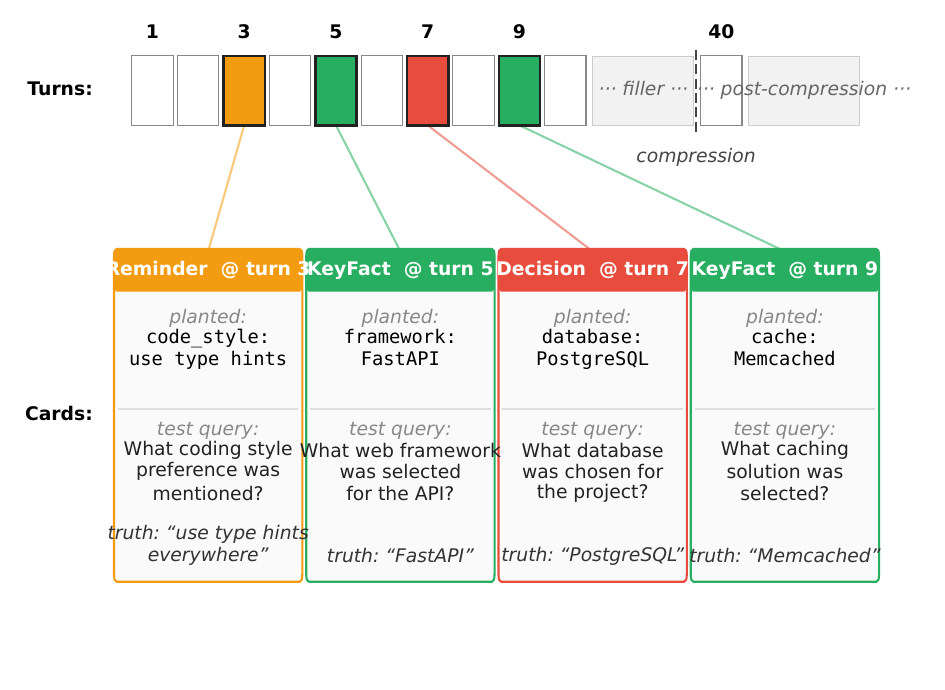}
\caption{Walked example of the synthetic benchmark protocol on \texttt{conv\,922f651b} (50 turns, ``hard'' difficulty). \textit{Turns row.} A 50-turn conversation has four facts planted at turns 3, 5, 7, 9 (colored cells). Filler dialogue (turns 11--39) and the post-compression query window (turns 41--50) are collapsed into ellipsis blocks for space. The dashed line at turn 40 marks the rolling-compression boundary that discards short-term context. \textit{Cards row.} Each planted fact carries a cognitive type drawn from the broader extraction ontology of \S\ref{sec:methodology} (here \textsc{Reminder} in orange, \textsc{KeyFact} in green, or \textsc{Decision} in red). Each card also shows the fact's structured \texttt{key:\,value} content, the post-compression test query, and the ground-truth answer used to score Recall and Exact Match.}
\label{fig:synthetic_walked_example}
\end{figure*}

\begin{figure*}[!t]
\centering
\includegraphics[width=\textwidth]{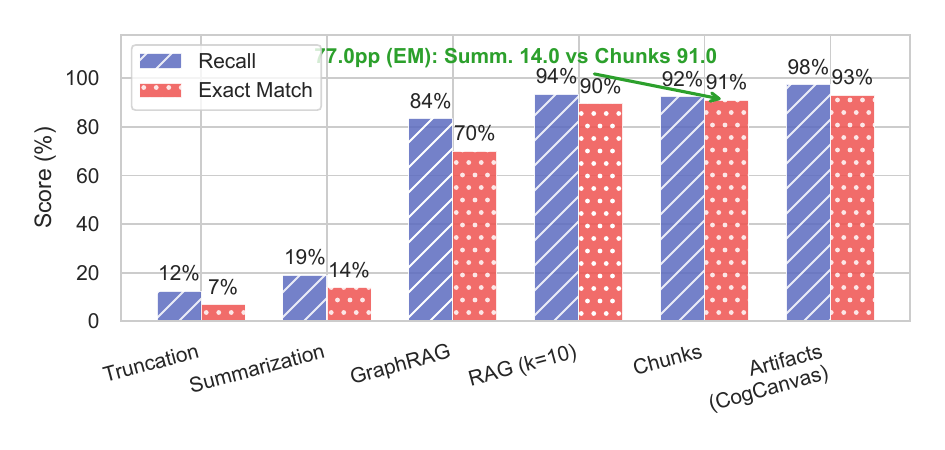}
\caption{Planted-fact retention on the synthetic benchmark (protocol of Table~\ref{tab:main_results}). Methods cluster by storage fidelity. Verbatim stores (RAG, chunks, quote-grounded artifacts) all exceed 90\% recall (89.5--93.0\% exact match), and lossy stores collapse. The annotated 77.0-point exact-match gap is the figure quoted in \S1.}
\label{fig:recall_comparison}
\end{figure*}

\begin{figure*}[!t]
\centering
\includegraphics[width=\textwidth]{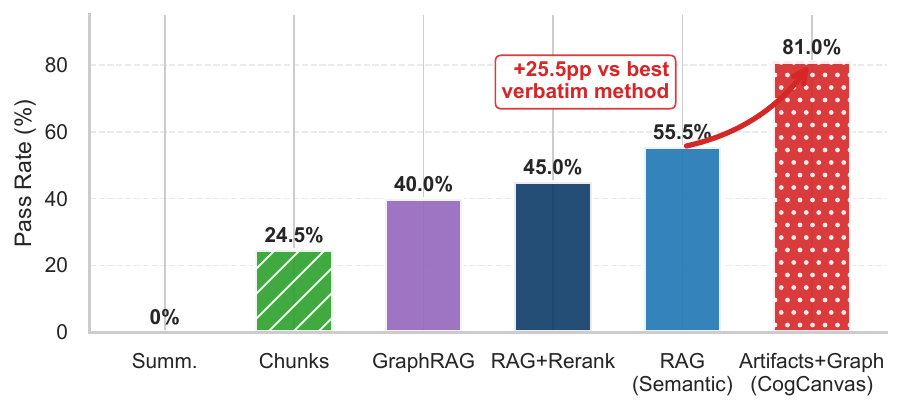}
\caption{The synthetic multi-hop probe is the one setting where the artifact pipeline wins (81.0\% vs.\ 55.5\% for the best verbatim method). It is extraction's best case, where every fact is pre-shaped into exactly one artifact. The advantage inverts on both public benchmarks.}
\label{fig:multihop_results}
\end{figure*}

\begin{figure*}[!t]
\centering
\includegraphics[width=\textwidth]{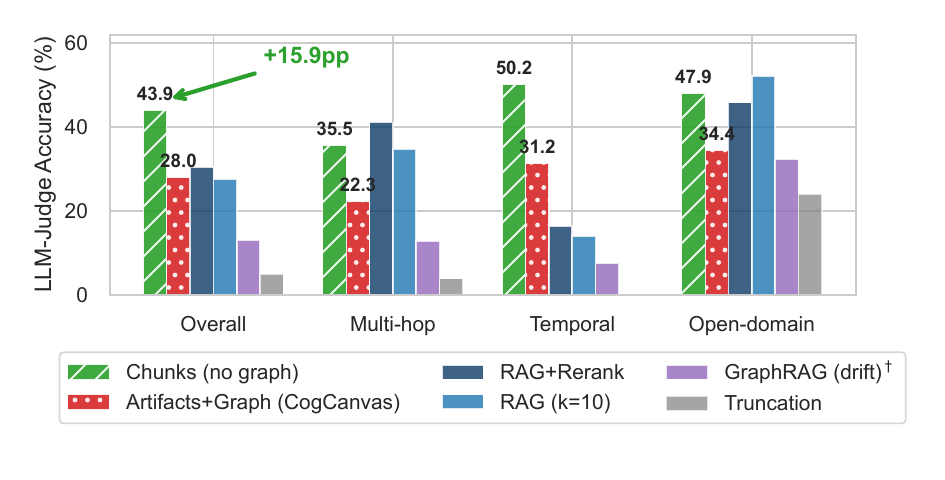}
\caption{LoCoMo breakdown by question type (per-question micro-averages, LLM Judge). Swapping LLM-extracted artifacts (red) for verbatim chunks (green) inside the same pipeline costs accuracy on every question type. $^\dagger$GraphRAG uses a gpt-4o-mini answerer (Table~\ref{tab:locomo_results}, footnote).}
\label{fig:locomo_results}
\end{figure*}

\section{Case Study: Verbatim Preservation}
\label{sec:verbatim_case_study}

The exact-match gap on the synthetic probe illustrates the fidelity mechanism at the scale of a single fact. Consider a \textsc{Reminder} planted in turn 3: \textbf{``use type hints everywhere.''}  When queried about coding style preferences after compression, the two lossy representations respond differently.

\vspace{0.5em}
\noindent\textbf{Summarization:} ``The user prefers a consistent code style with type hints.'' \hfill \textit{Exact: \ding{55}}

\noindent\textbf{Artifact pipeline:} ``please use type hints everywhere'' \hfill \textit{Exact: \ding{51}}
\vspace{0.5em}

\noindent The summarizer correctly identifies that type hints are relevant but loses the qualifier ``everywhere''. The artifact's verbatim \texttt{citation} field preserves the exact constraint. This is the regime where extraction is at its best. Each planted fact is a single self-contained statement that maps onto exactly one artifact, so nothing depends on the extractor's judgment of salience. The case therefore illustrates that the extraction machinery works when a fact is stated once and cleanly, and it leaves the extraction premise untested. A verbatim chunk store preserves the same wording \emph{unconditionally}, without requiring the extractor to notice the fact at all. The benchmark gap of Table~\ref{tab:locomo_results} opens where that condition fails. Of the artifact pipeline's diagnosable discordant failures, 29.3\% involve facts absent from every stored artifact, and a further 39.7\% involve facts the extractor never wrote down as a typed fact (Appendix~\ref{sec:error_analysis}).

\section{Stateful vs. Stateless Memory: A Trade-off Analysis}
\label{sec:stateful_discussion}

\textbf{Stateful} and \textbf{stateless} memory architectures for long conversations are commonly described as trading off as follows. Apart from the probe at the end of this section, we did not measure the stateful side.

\paragraph{Stateful Systems (e.g., Letta, EverMemOS).}
These methods maintain persistent memory state across interactions, typically backed by databases or specialized storage. Letta achieves 74.0\% on LoCoMo~\citep{letta2024memory}, while EverMemOS reaches 92.3\%~\citep{evermemos2025}. Their advantages include:
\begin{itemize}[leftmargin=*,itemsep=2pt]
    \item \textbf{Superior recall}: Persistent storage enables exhaustive memory retention without context length constraints.
    \item \textbf{Incremental learning}: Agents can accumulate knowledge over extended sessions.
\end{itemize}

However, these benefits come with costs:
\begin{itemize}[leftmargin=*,itemsep=2pt]
    \item \textbf{Infrastructure overhead}: Requires databases (PostgreSQL), containerization (Docker), and server management.
    \item \textbf{Deployment complexity}: Non-trivial setup prevents rapid prototyping and integration into existing systems.
\end{itemize}

\paragraph{Stateless Systems (this work).}
All four pipelines under test are stateless. The conversation is indexed once, queries are answered independently, and no persistent agent state is maintained. The regime's practical appeal is real (zero infrastructure with drop-in use of standard LLM APIs, no fine-tuning, full portability). Within it, our controlled comparison identifies verbatim chunks as the best stored representation (43.9\% on LoCoMo, 67.4\% on LongMemEval-S; Tables~\ref{tab:locomo_results} and~\ref{tab:longmemeval_results}), ahead of extracted artifacts.

\paragraph{When to Use Each Approach.}
Our findings suggest:
\begin{itemize}[leftmargin=*,itemsep=2pt]
    \item \textbf{Stateful systems}: Production deployments with long-lived agents, where infrastructure investment is justified by sustained interactions.
    \item \textbf{Stateless verbatim pipelines}: Rapid prototyping, research experiments, or applications requiring portability and simplicity. Within the stateless budget, our results indicate that the spend belongs in retrieval over verbatim text instead of extraction machinery.
\end{itemize}

The gap between the best stateless pipeline (43.9\%) and self-reported stateful results (Letta 74.0\%, EverMemOS 92.3\%) is large. It is still not a controlled comparison, because those figures come from different answerer models and evaluation stacks. Our controlled data support a narrower and firmer claim. In the stateless pipelines and the one stateful probe we ran, rewriting conversation text into distilled artifacts is dominated by keeping the text.

\paragraph{A controlled probe inside a stateful agent (referenced from \S\ref{sec:limitations}).}
\label{sec:stateful_probe}
The comparisons above swap representations in a \emph{stateless} pipeline. To test whether statefulness overturns the ordering, we ran the same one-variable swap \emph{inside} a stateful agent. The agent is a MemGPT-style controller that keeps the most recent turns in a fixed-size core memory. On rolling compression, it moves older turns into an embedded, keyword-indexed \emph{archival} store retrieved at query time. We hold the entire stateful loop fixed (core-memory size, compression cadence, archival top-$k$, retrieval, answerer, judge) and swap only the archival representation. The two arms are \emph{verbatim} archived turns and LLM-\emph{extracted} artifacts (the extractor of \S\ref{sec:methodology}). The extractor yields $\sim$1.3 artifacts per turn, several hundred per conversation, so the store is well populated. On LoCoMo (Cat~1--3, 699 questions), verbatim archival beats artifact archival on \emph{every} question type (Table~\ref{tab:stateful_probe}). The overall gain is $+$10.9pp (McNemar exact, 100 vs.\ 24 discordant pairs, $p{<}10^{-11}$), the same direction as the stateless headline. The absolute level is lower than our full stateless pipeline, because this lite controller is a weaker retrieval stack. The \emph{representation} ordering is unchanged, and adding state does not rescue lossy extraction.

\begin{table}[t]
\centering
\footnotesize
\setlength{\tabcolsep}{3pt}
\begin{tabular}{@{}p{0.37\columnwidth}cccc@{}}
\toprule
\textbf{Archival representation} & \textbf{Overall} & \textbf{M-hop} & \textbf{Temp.} & \textbf{O-dom.} \\
\midrule
Verbatim turns & \textbf{24.0} & \textbf{27.7} & \textbf{13.4} & \textbf{49.0} \\
Extracted artifacts & 13.2 & 14.5 & 7.2 & 29.2 \\
\midrule
\textit{$\Delta$ (verbatim $-$ artifacts)} & +10.9 & +13.1 & +6.2 & +19.8 \\
\bottomrule
\end{tabular}
\caption{\textbf{Statefulness does not overturn the ordering.} We swap only the archival representation inside a stateful MemGPT-style agent (core memory $+$ rolling compression $+$ archival retrieval) and hold everything else fixed. Evaluation is on LoCoMo Cat~1--3 (699 questions, gpt-4o answerer, gpt-4o-mini judge). Verbatim archival dominates extracted-artifact archival on every question type (overall $+$10.9pp, McNemar exact $p{<}10^{-11}$, 100 vs.\ 24 discordant). The $\Delta$ row is computed from unrounded counts (168 vs.\ 92 of 699 correct), and cells are rounded to one decimal.}
\label{tab:stateful_probe}
\end{table}

\section{Error Analysis: Where the Artifact Pipeline's Answers Go Missing}
\label{sec:error_analysis}

We analyzed 3,121 missing keywords across the artifact pipeline's failed LoCoMo questions and found three primary failure modes. (1) \textbf{Extraction gaps} (78.8\%) occur when the LLM extractor fails to identify certain facts as salient (e.g., ``Caroline is single'' is not extracted as a key\_fact). (2) \textbf{Temporal blindness} (14.9\%) covers dates mentioned in passing (``last Tuesday,'' ``May 2023'') that are not preserved as structured temporal attributes. (3) \textbf{Reasoning chain breaks} (6.4\%) are multi-hop questions that require inference across multiple artifacts and fail when intermediate links are missing. This typology is assigned by inspecting answers. It therefore cannot by itself separate \emph{write-time} losses (the store never contained the fact) from \emph{query-time} misses (the store contained it but retrieval or generation failed). The store-level probe below makes that separation directly.

\paragraph{Store-Level Survival Probe.}
We examine the 147 questions that the chunk pipeline answers correctly and the budget-matched artifact pipeline misses (the discordant pairs of \S\ref{sec:experiments}). For each, we test whether the gold answer's content tokens co-occur within any \emph{single} stored artifact (its content, quote, and context fields). We take co-occurrence within a single conversation turn as the fairness ceiling. Golds that never co-occur verbatim even in one turn are analyzed separately, and two golds with no content tokens at all (e.g.\ ``Yes'') are excluded as undiagnosable. Among the 58 questions whose gold appears verbatim in a single turn, 17 (29.3\%) survive in \emph{no} artifact at all. Another 23 (39.7\%) survive \emph{only} inside the raw-text window that temporal extraction copies verbatim as context, and never as a deliberately extracted fact. ``Caroline is single'' is one of these. The answer token appears in no key\_fact, only mid-window, and it survives at all because that field keeps text verbatim. We report the two classes separately rather than as one figure. In the strict sense, 29.3\% (17/58) are write-time losses (no lexical trace anywhere in the artifact store). Another 39.7\% (23/58) are extraction failures (the fact survives in a stored raw-context field but was never extracted as a typed fact). The remaining 31.0\% (18/58) are query-time misses on facts that were extracted but not surfaced. Only the first class supports ``irreversible loss''; the second supports ``failed to reach a retrievable item''.
\paragraph{Paraphrase-tolerant re-test.}
Token co-occurrence counts a fact as lost when an artifact restates it in other words. It also cannot diagnose the 87 questions whose gold never co-occurs in one turn. We therefore repeated the survival test on all 145 non-trivial questions without string matching. For each question we embed the question together with its gold answer (bge-m3) and shortlist the ten most similar stored units of each kind (extracted artifacts and raw-text windows). A gpt-4o-mini judge then decides whether a unit, read on its own, states the information needed to give the gold answer; paraphrases are accepted. The judge is validated per question against LoCoMo's gold evidence annotations. Shown the annotated evidence utterances with their session dates, it must find them sufficient. It does so for 115 of 145 questions (42 of the 58 diagnosable ones). We exclude the other 30, because a judge that rejects the evidence itself cannot certify loss; many of them need an inference across several evidence utterances. As a negative control, the ten best units of a \emph{different} conversation's store are accepted for 2 of 145 questions.

On the 115 validated questions, an extracted artifact is judged sufficient for 5 (4.3\%), only a raw-text window for 25 (21.7\%), and no shortlisted unit for 85 (73.9\%). On the validated diagnosable subset the counts are 2, 11, and 29 of 42. Paraphrase tolerance therefore does not move the answers back into the extracted artifacts. Of the token-level write-time losses that pass validation (14 of 17), 12 remain without a sufficient unit and the other two are recovered from a raw-text window. We keep the token-based classes above as reported, since they measure lexical trace. The judge measures a stricter property, whether a unit states the fact. Under it, most questions that token matching classed as extracted have no sufficient artifact (10 of the 12 validated ones). The two tests thus bracket the estimate, and both leave most of these answers outside the deliberately extracted artifacts. Two limits apply. A sufficient unit ranked below the shortlists (at most 20 units per question) would be missed, and the judge is a single model.

The other 87 questions have golds that appear verbatim in no single turn. These are typically dates that must be resolved from a relative expression plus a session header (``yesterday'' $+$ ``8 May, 2023''). The chunk pipeline wins them by handing the answerer the surface context that enables the inference, which rewriting strips; it does not need to contain the answer string. In 25.3\% of them the artifact store's best single-object coverage falls below even the single-turn ceiling. A verbatim-chunk store cannot commit the write-time class at all. Its failures are confined to query-time retrieval misses and over-answering (abstention). For this reason the Discussion characterizes these as errors avoidable by not extracting.

\textbf{What Gets Missed?} Qualitative analysis of extraction failures reveals systematic patterns in what the LLM extractor overlooks.
\begin{itemize}[nosep]
    \item \textbf{Implicit personal attributes} (e.g., ``Caroline is single,'' relationship status inferred from context rather than stated directly). The extractor prioritizes explicit statements over implicit facts.
    \item \textbf{Emotional preferences} (e.g., ``she felt anxious about...''). Affective states are often not recognized as \textsc{KeyFact}s worth preserving.
    \item \textbf{Casual temporal references} (``last Tuesday,'' ``a few months ago''). Relative time expressions are harder to anchor than absolute dates.
    \item \textbf{Negations and constraints} (``does NOT want X''). Negative preferences are sometimes missed or inverted during extraction.
\end{itemize}
These patterns suggest that extraction prompts could be improved by explicitly instructing the LLM to attend to implicit attributes, emotional states, and negative constraints.

\textbf{Gleaning contribution.} The gleaning pass (\S\ref{sec:methodology}) is designed to address these gaps. It explicitly targets omitted subjects and implicit relationships in a second extraction scan. The main LoCoMo and LongMemEval configurations run the initial extraction plus one gleaning pass. Its measured contribution is negligible. In an ablation under an earlier evaluation protocol (January batch), disabling gleaning \emph{raised} LoCoMo accuracy from 28.6\% to 30.2\% (200 vs.\ 211 of 699 questions), a 1.6pp difference. The extraction gaps documented above therefore persist with or without the second pass. We did not tune the number of gleaning passes per-dataset.

\textbf{Stronger-extractor swap.} A natural objection is that the extractor itself (gpt-4o-mini in the main configuration) is too weak, and that a stronger model would close the gap. It does not. We re-ran extraction with gpt-4o on a 3-conversation LoCoMo subset ($n{=}185$ questions). Artifact accuracy is statistically unchanged (30.3\% with gpt-4o vs.\ 29.7\% with gpt-4o-mini on the same subset; McNemar $p{\approx}1$). The verbatim-chunk advantage stays intact at $-$17.8pp on this subset ($p{<}10^{-4}$). Extractor strength is therefore not the missing ingredient. The loss lies in what extraction \emph{discards} relative to the source, rather than in how well a given model performs the extraction.

\textbf{External-system anchor with the official Mem0 package.} \label{sec:mem0_anchor}
A second objection targets the in-pipeline reproductions of \S\ref{sec:discussion}. Because we re-implement each system's defining mechanism inside our fixed pipeline, we might misrepresent the real system and argue against a strawman. We rule this out for Mem0~\citep{chhikara2025mem0} by running its \emph{official} package end-to-end on LoCoMo (Cat~1--3, all 10 conversations, $n{=}699$). The run uses Mem0's own extraction loop (\texttt{add(infer=True)}), its own vector search, and its native OpenAI embedder (\textsf{text-embedding-3-small}, cosine retrieval). Three stores share one harness: (1)~official Mem0 extracted memory, (2)~verbatim 512-character chunks over the same embedder, and (3)~the full transcript (reference ceiling). All three use a \emph{shared} \textsf{gpt-4o-mini} answerer and the \emph{same} LLM judge used for the headline tables. Both memory stores receive per-item session timestamps, and Mem0 is granted twice the retrieval budget of chunks (top-30 vs.\ top-15). Even so, official Mem0 trails verbatim chunks overall and on temporal questions (\textbf{Mem0 36.6\%} $<$ \textbf{chunks 47.9\%}, $-$11.3pp), and Mem0 also trails full context (40.5\%). The deficit concentrates where the fidelity account predicts, on temporal questions (27.1\% vs.\ 51.1\%, $-$24.0pp). Mem0's extraction rewrites or drops the surface dates that the chunk store keeps verbatim. Mem0's own LoCoMo report shows the same extracted-vs-full-context direction (66.9 $<$ 72.9).

\textbf{Strong-answerer re-test (the strawman closed at full strength).} The harness above answers with \textsf{gpt-4o-mini}. A residual worry is that a weak answerer depresses every extraction system below its deployed level (Mem0's 36.6\% trails its self-reported 66.9), so the ordering might be an answerer artifact. We therefore re-ran the identical three-store anchor with the \emph{strong} answerer of our headline tables (\textsf{gpt-4o}). Mem0 also used \textsf{gpt-4o} for its \emph{own} extraction loop, and the run covers the wider Cat~1--4 split (all 10 conversations, $n{=}1{,}540$). Official Mem0 climbs to \textbf{54.7\%}, far above its stripped-harness 36.6\%, so the harness does not cripple it. Verbatim chunks still reach \textbf{69.9\%}, a \textbf{$+$15.2pp} margin (exact McNemar test, 343 vs.\ 109 discordant pairs, $p{=}3.0{\times}10^{-29}$). The margin holds in \emph{every} category (multi-hop $+$2.1, 46.5 vs.\ 44.3; temporal $+$39.9, 59.2 vs.\ 19.3; open-domain $+$6.2, 60.4 vs.\ 54.2; single-hop $+$11.2, 82.9 vs.\ 71.7). The temporal gap is the widest, as the fidelity account predicts. Mem0's write-time rewriting discards the surface dates that a verbatim store keeps for free (\S\ref{sec:why_temporal}). Full context for this split reaches 74.4\% and sits above both; with no retrieval, it is harness-independent given the answerer. The strong answerer thus restores the expected fidelity ordering (full $74.4 >$ chunks $69.9 >$ Mem0 $54.7$). It also dissolves the chunks-above-full inversion that the \textsf{gpt-4o-mini} answerer produces. That inversion is consistent with a lost-in-the-middle effect of handing $\sim$26K tokens to a small model; we did not test this explanation. At its own full strength, then, a deployed extraction system still loses to verbatim chunks by the same $\sim$15-point margin as our headline, within one controlled comparison.

{\sloppy Two further controls localize the cause to the stored representation rather than our reproduction. \emph{(i)~Not a strawman extractor.} Our \emph{own} typed-artifact extractor, run in this same stripped harness, scores 42.5\%, \emph{above} official Mem0 (36.6\%). The headline artifact deficit therefore does not come from an extractor weaker than a deployed system. In this harness our extractor in fact stores more items per conversation than Mem0. \emph{(ii)~Not a reranker artifact.} With the bge reranker removed and both stores matched at top-15 under symmetric timestamps, verbatim chunks still beat our artifacts by $+$4.6pp (47.1\% vs.\ 42.5\%). The full rerank pipeline of Table~\ref{tab:locomo_results} amplifies this to the $+$15.9pp headline. The verbatim advantage therefore keeps its direction across pipelines. It is present without the reranker and widened by it. The lossy-extraction deficit is also present in a deployed system (Mem0) and is not an artifact of our reproduction.\par}

The high proportion of temporal markers (14.9\%) is consistent with temporal questions remaining hard for the artifact pipeline (31.2\% in the same-batch protocol of Table~\ref{tab:locomo_results}), even though its date attributes and quote fields lead the timestamp-less baselines. Extraction preserves well what it notices, and what it misses is lost at write time (Appendix~\ref{sec:why_temporal}). Better extraction prompts (explicit date/time targeting, temporal attribute modeling) could shrink this class. They cannot reach the unconditional retention of a verbatim store.

\textbf{Performance by question type.} In the same-batch protocol of Table~\ref{tab:locomo_results}, the artifact pipeline reaches 22.3\% on multi-hop, 31.2\% on temporal, and 34.4\% on open-domain questions. It leads RAG (14.0\%) and RAG$+$Rerank (16.5\%) on temporal reasoning, the category where its date attributes and quote fields matter most. It trails RAG$+$Rerank by a wide margin on multi-hop (22.3\% vs.\ 41.1\%). The category that pre-joined artifacts are supposed to serve is instead won by retrieval over verbatim text. This echoes the finding of \S\ref{sec:discussion} that extraction's synthetic multi-hop advantage does not transfer. The chunk pipeline leads the artifact pipeline in every category.

\section{Temporal Questions as a Fidelity Probe}
\label{sec:why_temporal}

{\sloppy Temporal questions separate the stored representations more sharply than any other category. The ordering tracks a single write-time quantity, namely how much temporal source information (session timestamps and the surface strings that carry them) survives in the store. In the same-batch protocol of Table~\ref{tab:locomo_results}, timestamp-marked verbatim chunks reach \textbf{50.2\%}. Extracted artifacts, whose schema captures dates as structured attributes alongside verbatim quotes, reach 31.2\%. The standalone chunk baselines trail far behind (RAG$+$Rerank 16.5\%, RAG 14.0\%). They store verbatim text but lack the pipeline's explicit session-timestamp markers and adaptive retrieval depth. Representations that pass through summarization collapse (GraphRAG 7.5\%, Truncation 0.3\%). The artifact pipeline leads RAG$+$Rerank by 14.7pp (the two share the same BGE reranker; $+$13.1pp under the decontaminated-prompt control of Appendix~\ref{sec:prompts}) and trails verbatim chunks by 19.0pp. These are two halves of one account. Extraction preserves the timestamps the extractor notices, a timestamp-marked verbatim store preserves all of them, and a summarization layer preserves almost none. This section traces that mechanism through failure cases and component ablations.\par}

\paragraph{Failure Cases: GraphRAG vs.\ the Artifact Pipeline.}
A representative instance of summarization-style compression losing specific dates is \textit{``When did Caroline go to the LGBTQ support group?''} (ground truth May~7,~2023). The verbatim-chunk pipeline retrieves the original turn and returns ``May~7,~2023''. GraphRAG has summarized the relevant turns into a community-level description and returns ``the specific date of her attendance is not provided''. The event survives summarization, but its timestamp does not. A second illustrative case from the LoCoMo benchmark shows the same failure mode under a different surface form.

\begin{itemize}
    \item \textbf{Example (When did Melanie paint a sunrise?)}
    \begin{itemize}
        \item \textbf{GraphRAG:} offered only a vague temporal reference, stating, ``the specific date... is not provided... mentioned that she created a painting... last year.'' The exact date was generalized away during community summarization.
        \item \textbf{Artifact pipeline:} returned the exact date, ``8 May 2022''. The extractor captured this event, and the artifact's verbatim quote field carried the date string through to retrieval. A verbatim chunk store preserves the same string with no extraction step at all.
    \end{itemize}
\end{itemize}

The contrast places the three representations on the fidelity spectrum of \S\ref{sec:discussion} without declaring a winner. Community summarization discards the date. Extraction keeps it \emph{when the extractor notices it}; the 14.9\% temporal-blindness class in Appendix~\ref{sec:error_analysis} consists of the cases it does not. Verbatim chunks keep it unconditionally.

\paragraph{Ablation Study: Component Contributions.}
\textit{Scope note.} This ablation is internal to the \emph{artifact} pipeline. It measures how much each component contributes \emph{given} that one has already committed to extracted artifacts as the stored representation. It also predates the chunks comparison. All rows are from the January batch, the protocol of the gleaning ablation in Appendix~\ref{sec:error_analysis}. In the same-batch protocol of Table~\ref{tab:locomo_results}, the full configuration measures 28.0\%, consistent with the 28.6\% here. Its findings therefore do not contradict the main result. Graph expansion recovers some of what extraction loses \emph{relative to artifacts without a graph}, yet the completed system still trails verbatim chunks by 15.9 points. Structure mitigates the damage of extraction without repairing it.

To quantify the individual contributions of the artifact pipeline's components, we remove each component from the full system and measure the performance drop.

\begin{table}[t]
\centering
\footnotesize
\setlength{\tabcolsep}{4pt}
\begin{tabular}{@{}lcccc@{}}
\toprule
\textbf{Configuration} & \textbf{All} & \textbf{M-hop} & \textbf{Temp.} & \textbf{O-dom.} \\
\midrule
\textbf{Full System} & \textbf{28.6\%} & \textbf{20.9\%} & \textbf{30.5\%} & \textbf{44.8\%} \\
\midrule
$-$ Graph Expansion & 25.8\% & 19.5\% & 29.0\% & 33.3\% \\
$-$ Reranking & 20.9\% & 16.7\% & 23.1\% & 26.0\% \\
\midrule
Minimal (Graph only) & 14.3\% & 11.7\% & 13.7\% & 24.0\% \\
\bottomrule
\end{tabular}
\caption{Ablation study on LoCoMo (Cat 1/2/3, all 10 conversations, LLM Judge, question-micro; January batch). Full system achieves 28.6\%. Reranking is the largest contributor ($-$7.7pp when removed), followed by graph expansion ($-$2.8pp overall, $-$11.5pp on open-domain).}
\label{tab:component_ablation}
\end{table}

\textbf{Key findings.}
\begin{itemize}
    \item \textbf{Reranking is the largest contributor.} Removing BGE reranking causes a --7.7pp drop (28.6\% $\to$ 20.9\%), the largest of any component within the artifact pipeline. The same reranker added to vanilla RAG yields 30.5\% overall in the same-batch protocol (Table~\ref{tab:locomo_results}). Reranking is thus a general-purpose improvement rather than an artifact-specific one. On temporal questions, RAG+Rerank reaches only 16.5\% against the artifact pipeline's 31.2\% (and the chunk pipeline's 50.2\%). Temporal accuracy is driven by how much source wording, especially timestamps, survives in the store, consistent with the fidelity ordering of \S\ref{sec:discussion}.
    \item \textbf{Graph expansion helps only within the artifact representation.} Removing graph expansion reduces overall accuracy by --2.8pp (28.6\% $\to$ 25.8\%), with the largest impact on open-domain questions (--11.5pp, 44.8\% $\to$ 33.3\%). This is a within-representation effect. The graph recovers part of what extraction loses relative to artifacts without a graph. The same graph over verbatim chunks is a no-op (43.1\% vs.\ 43.9\%, Table~\ref{tab:locomo_results}), since structure cannot regenerate content the store never kept (\S\ref{sec:discussion}).
    \item \textbf{Combined system effects.} The full system (28.6\%) improves +14.3pp over the minimal baseline (14.3\%).
    \item \textbf{Temporal accuracy is carried by preserved wording rather than edges.} The artifact pipeline's temporal lead over the timestamp-less baselines (31.2\% vs.\ RAG's 14.0\%, Table~\ref{tab:locomo_results}) comes from the schema's date attributes and verbatim quote fields, which are the least lossy parts of the representation. It is more plausibly attributable to these fields than to temporal heuristic edges, which the ablation does not isolate. The same quantity bounds what the pipeline cannot do. Dates the extractor fails to notice (Appendix~\ref{sec:error_analysis}) are unrecoverable at query time. This is why timestamp-marked chunks, which keep every date unconditionally, lead by a further 19.0pp.
\end{itemize}

\paragraph{Retrieval Quality Analysis.}
To verify that graph expansion improves \textit{retrieval quality} rather than merely leveraging LLM reasoning capabilities, we run a retrieval-only evaluation. It measures whether the retrieved context contains ground-truth keywords, \textbf{without invoking the LLM for answer generation}. As shown in Table~\ref{tab:retrieval_recall}, graph expansion increases retrieval recall from 25.8\% to 71.8\% (\textbf{+45.9pp}). This indicates that the performance gains in Table~\ref{tab:component_ablation} stem from \textit{improved context retrieval} rather than from LLM answering ability. The graph structure enables retrieving related artifacts that pure semantic similarity matching would otherwise miss.

\paragraph{Why Baselines Fall Short on Temporal Reasoning.}
The low performance of baselines on temporal questions (RAG 14.0\%, GraphRAG 7.5\%, Summarization 0.9\%; Table~\ref{tab:locomo_results}) can be attributed to several factors.
\begin{itemize}
    \item \textbf{RAG (Chunk-based).} Standard RAG systems typically chunk conversations or documents into fixed-size segments. Temporal information, such as specific dates or sequences of events, can be fragmented across chunk boundaries, which makes it difficult for retrieval to reconstruct the full temporal context. Without explicit temporal indexing or relational understanding, chunks are retrieved by semantic similarity, which often overlooks the temporal dimension.
    \item \textbf{GraphRAG (Community Summarization).} GraphRAG constructs a knowledge graph, but its reliance on community detection and summarization for higher-level queries often generalizes information. Specific dates, times, or precise event sequences are frequently summarized away or absorbed into broader thematic summaries. The granular temporal detail needed for precise temporal questions is then lost. The emphasis on conceptual relationships can also obscure the temporal ordering of events.
    \item \textbf{Summarization.} Directly compressing conversation history into summaries is lossy by construction. The summarization LLM often treats temporal details as less ``important'' than core facts or decisions, so they are omitted or represented vaguely. Iterative summarization exacerbates this issue, since temporal markers are progressively lost with each compression pass. Time-sensitive data thus decays recursively.
\end{itemize}
All three failure modes are instances of the same write-time quantity, namely how much temporal source information survives in the store. They place each system on the fidelity spectrum of \S\ref{sec:discussion}. Artifact extraction sits in the middle of that spectrum. Its structured date attributes and verbatim quotes preserve the temporal markers the extractor notices, hence its lead over the timestamp-less and summarization-based baselines. Everything it fails to notice is lost before any query arrives, hence its 19.0pp deficit against timestamp-marked verbatim chunks (Table~\ref{tab:locomo_results}). The category-level lesson matches the paper's headline result. \emph{Preserving} temporal wording matters more than \emph{connecting} it through graph structure, and keeping all of it beats extracting some of it.

\section{A Real-Extraction Fidelity Dial: SeCom's Compression Sweep}
\label{sec:secom_dial}
The controlled dial of \S\ref{sec:discussion} degrades fidelity by deleting \emph{random} tokens. This invites the objection that it measures the damage of random corruption rather than the selective loss of a real extractor. We replicate the dial with a real extractive-compression mechanism, SeCom's LLMLingua-2 denoiser, and sweep its token-retention rate over $\{1.0, 0.75, 0.5\}$ with the pipeline otherwise fixed. The setup is backbone-routed \texttt{cogcanvas-secom-nograph} with a gpt-4o answerer on LoCoMo Cat~1--3 ($n{=}10$, 699 questions), scored by micro-accuracy from per-question LLM-Judge decisions.

\begin{table}[h]
\centering
\footnotesize
\setlength{\tabcolsep}{3pt}
\begin{tabular}{@{}p{0.38\columnwidth}cccc@{}}
\toprule
\textbf{compress\_rate} & \textbf{Overall} & \textbf{M-hop} & \textbf{Temp.} & \textbf{O-dom.} \\
\midrule
1.00 (verbatim segments) & 47.4 & 37 & 56 & 47 \\
0.75 (SeCom default)     & 46.9 & 39 & 53 & 49 \\
0.50 (aggressive)        & \textbf{36.9} & 30 & 41 & 44 \\
\midrule
\textit{ref:} verbatim chunks & \textit{45.2} & & & \\
\textit{ref:} full context    & \textit{59.9} & & & \\
\bottomrule
\end{tabular}
\caption{SeCom accuracy as a function of its own extractive-compression rate (LoCoMo Cat~1--3, micro-accuracy). Accuracy falls monotonically as less verbatim text is retained. At aggressive compression a real extractive compressor drops \emph{below} the fixed-chunk baseline. \emph{Ref.}\ rows re-measure the chunks and full-context anchors in this batch. Chunks re-measures at 45.2 against the 43.9 headline of Table~\ref{tab:locomo_results}, a same-configuration run-to-run difference of 1.3pp (Appendix~\ref{sec:provenance}).}
\label{tab:secom_dial}
\end{table}

Accuracy falls monotonically as SeCom retains less verbatim text. At aggressive compression (0.5) it drops to 36.9\%, below the fixed-chunk baseline (45.2\%), which places a real extractive compressor in the dial's low-fidelity regime. The 0.75 default (46.9\%) sits 2.0pp below the SeCom mechanism-reproduction anchor of Figure~\ref{fig:fidelity_dial} (48.9\%). The two runs are a same-configuration repeat, and this is the widest gap we observed (disclosed in Table~\ref{tab:provenance}), so the sweep is on the same scale. Two further observations follow. (i) The curve \emph{plateaus} between 0.75 and 1.0 ($+0.5$pp, within noise). Mild denoising loses no measurable accuracy, while keeping \emph{more} verbatim text adds no accuracy, so SeCom saturates near 47\%, still about 12.5 points below full context. (ii) At full retention, SeCom's topical segmentation edges fixed-window chunks by $\sim$2pp (47.4 vs.\ 45.2), the segmentation effect noted in \S\ref{sec:discussion}. The load-bearing point is the monotone collapse under a \emph{real} extractor rather than random deletion. The fidelity$\to$accuracy ordering is a property of how much source text survives, under both random deletion and a real extractive compressor.

\section{Separating Fidelity from Granularity: A Sentence-Verbatim Control}
\label{sec:sentence_control}
Verbatim chunks ($\sim$512-char windows) differ from typed artifacts on two axes at once. They differ in granularity (a coarse window vs.\ a short item) and in fidelity (raw source vs.\ lossy distillation), so the chunks-vs-artifacts gap could in principle be a granularity effect. We isolate the two axes with a \emph{sentence-verbatim} store. It keeps the transcript fully verbatim but cuts it to artifact-scale units (one or more whole sentences packed to $\le$160 characters; no extraction, no compression, sentences never split). The store runs through the same nograph backbone as every other anchor (gpt-4o-mini answerer, LoCoMo Cat~1--3, 699 questions, question-micro accuracy). The budget-matched $k{=}38$ row was run in 2026-09 and the other rows in one 2026-06 batch; the ordering is answerer-invariant (\S\ref{sec:limitations}). Table~\ref{tab:sentence_control} reports the decomposition.

\begin{table}[t]
\centering
\small
\setlength{\tabcolsep}{4pt}
\begin{tabular}{@{}lrr@{}}
\toprule
Store (all same backbone) & Tok/q & Acc. \\
\midrule
Verbatim chunks (coarse) & 2{,}127 & 42.8 \\
Sentence-verbatim, $k{=}90$ (2.2$\times$ budget) & 4{,}674 & 39.1 \\
Sentence-verbatim, $k{=}38$ (budget match) & 2{,}140 & 36.2 \\
Sentence-verbatim, $k{=}15$ & 1{,}010 & 32.2 \\
Typed artifacts (lossy, fine) & 1{,}159 & 26.5 \\
\bottomrule
\end{tabular}
\caption{Sentence-verbatim control (question-micro accuracy over 699 questions; gpt-4o-mini answerer). The chunks, $k{=}90$, $k{=}15$ and artifacts rows come from one batch; the $k{=}38$ row was added in a later batch under the same configuration. ``Tok/q'' is mean answerer-prompt tokens per question. At matched fine granularity and budget, verbatim beats lossy by 5.7pp (32.2 vs.\ 26.5). At the chunk token budget ($k{=}38$, 2{,}140 vs.\ 2{,}127 tokens), the sentence store trails chunks by 6.6pp, which is granularity's share of the gap. The $k{=}90$ row (2.2$\times$ the chunk budget) is kept for reference. This batch's chunks and artifacts anchors (42.8, 26.5) re-measure the mini-answerer anchors reported elsewhere (41.9, Appendix~\ref{sec:limitations_full}; 28.5--29.9, Appendix~\ref{sec:retriever_robustness}). The two sets agree within 0.9--3.4pp, the cross-batch range documented in Appendix~\ref{sec:provenance}.}
\label{tab:sentence_control}
\end{table}

Three readings isolate the axes. \emph{(1)~Fidelity, controlling for granularity and budget.} At the same fine granularity and a slightly \emph{smaller} answerer budget (1{,}010 vs.\ 1{,}159 tokens), the sentence-verbatim store beats artifacts by 5.7pp (32.2\% vs.\ 26.5\%). Keeping the source verbatim helps even when unit size matches the artifacts'. \emph{(2)~Granularity, controlling for fidelity and budget.} Raising the sentence store's $k$ to 38 matches the chunk token budget (2{,}140 vs.\ 2{,}127 tokens) and lifts the store to 36.2\%, 6.6pp below verbatim chunks. Over-budgeting it further ($k{=}90$, 4{,}674 tokens) reaches 39.1\%. At equal budget, coarse-vs-fine granularity therefore accounts for 6.6pp of the 16.3pp gap. \emph{(3)~Budget-responsiveness as a fidelity signature.} Added budget lifts the \emph{verbatim} sentence store by 4.0pp at matched budget (32.2$\to$36.2) and by 6.9pp when over-budgeted (39.1). It lifts the \emph{lossy} artifact store by only 1.2pp (the $k{=}60$ control, \S\ref{sec:experiments}). Retrieval budget thus converts to accuracy only when the source survives in the store. Of the 16.3pp gap, 6.6pp is granularity at equal budget. The remaining 9.7pp separates two fine-grained stores that differ in both fidelity and budget. Since budget alone moves the lossy store by 1.2pp, most of that remainder is fidelity. Fidelity is the larger term and the one that isolates cleanly at matched granularity (reading 1).

\section{Answerer and Extractor Vendor Control}
\label{sec:vendor_control}
Every headline cell uses gpt-4o as answerer and gpt-4o-mini as extractor, so the chunks-over-artifacts ordering could in principle be a property of one vendor's models. We re-ran the Table~\ref{tab:locomo_results} pair with both LLM roles moved to a second vendor and everything else fixed. The answerer is \textsf{claude-sonnet-4-6} (temperature 0), and the extractor and graph builder is \textsf{claude-haiku-4-5}. The bge-m3 retriever, bge-reranker-v2-m3, gpt-4o-mini judge, and LoCoMo categories 1--3 (699 questions, 10 conversations) are unchanged. Table~\ref{tab:vendor_control} reports both pairs as question-level micro accuracy, the Table~\ref{tab:locomo_results} convention, with the mean over conversations in parentheses for the new pair.

\begin{table}[t]
\centering
\footnotesize
\setlength{\tabcolsep}{3pt}
\begin{tabular}{@{}p{0.33\columnwidth}cc@{}}
\toprule
\textbf{Store (same pipeline)} & \textbf{GPT pair} & \textbf{Claude pair} \\
\midrule
Verbatim chunks & 43.9 & 53.9 (54.7) \\
Artifacts $+$ graph & 28.0 & 33.2 (34.0) \\
\midrule
$\Delta$ Chunks $-$ Artifacts & $+$15.9 & $+$20.7 \\
Prompt tok/q, chunks / artifacts & 2{,}127 / 1{,}191 & 2{,}400 / 1{,}518 \\
\bottomrule
\end{tabular}
\caption{The Table~\ref{tab:locomo_results} pair under two vendors for answerer and extractor (LoCoMo Cat 1--3, 699 questions; question-micro, mean over the 10 conversations in parentheses; judge gpt-4o-mini in both). The GPT pair is gpt-4o / gpt-4o-mini, and the Claude pair is claude-sonnet-4-6 / claude-haiku-4-5. On the Claude pair, the McNemar exact test counts 199 chunks-only correct and 54 artifacts-only correct ($p{=}1.1{\times}10^{-20}$). Token counts are the vendors' own accounting and are not directly comparable across columns. The within-column asymmetry is the one analysed in Table~\ref{tab:locomo_results}.}
\label{tab:vendor_control}
\end{table}

Both stores improve under the stronger answerer and the ordering survives. The gap widens from 15.9 to 20.7pp. This is what the fidelity account predicts. A better reader raises accuracy in proportion to what reaches it, and less reaches it from the artifact store. Extraction cost for the new pair was 81 extractor calls and 413K prompt tokens over the 2{,}938 indexed turns; answerer cost was 2.74M prompt tokens.

\section{A Near-Verbatim EDU Control: Measuring the High-Fidelity End of Extraction}
\label{sec:emem_control}
Concurrent work reports that a training-free memory built from ``enriched elementary discourse units'' (EDUs) reaches strong LoCoMo accuracy~\citep{zhou2025simple}. EDUs are self-contained statements with normalized entities and source-turn attributions, deliberately constructed to preserve information ``in a non-compressive form''. Our fidelity account predicts this result (\S\ref{sec:discussion}). Here we test the prediction by measuring the EDU \emph{representation} inside the fixed pipeline, the same mechanism-reproduction treatment given to Mem0, A-Mem, and SeCom. A gpt-4o-mini extractor decomposes each utterance into EDUs under three constraints that match the published design. These are full coverage (no salience filtering), near-verbatim wording (only pronoun-to-name normalization plus minimal glue words), and the same dated turn marker every other anchor prints. The store is dense ($\sim$21k units over LoCoMo10, $\sim$7 per turn, against $\sim$1.3 typed artifacts). A parse failure falls back to sentence-verbatim units, so no turn is silently dropped (the decomposition prompt ships with the released code). Because EDUs are fine-grained, we sweep retrieval depth. Depth $k{=}15$ matches the headline protocol, $k{=}60$ approximately matches the chunks token budget, and $k{=}90$ overshoots it.

\begin{table}[t]
\centering
\footnotesize
\setlength{\tabcolsep}{2.5pt}
\begin{tabular}{@{}lrrrrrr@{}}
\toprule
\textbf{Store} & $k$ & \textbf{Tok/q} & \textbf{All} & \textbf{M-hop} & \textbf{Temp.} & \textbf{O-dom.} \\
\midrule
Chunks & 15 & 2{,}127 & \textbf{47.4} & 37.9 & 53.9 & 53.1 \\
EDU & 90 & $\sim$2{,}700$^e$ & 36.5 & 23.4 & 46.7 & 40.6 \\
EDU & 60 & $\sim$1{,}900$^e$ & 36.3 & 22.0 & 47.4 & 41.7 \\
EDU & 15 & $\sim$700$^e$ & 29.8 & 17.0 & 40.5 & 31.2 \\
Artifacts & 15 & 1{,}159 & 30.2 & 26.2 & 30.8 & 39.6 \\
\bottomrule
\end{tabular}
\caption{EMem-style EDU control (LoCoMo Cat~1--3, question-micro accuracy; Chunks = verbatim chunks, EDU = EMem-style EDU store, Artifacts = typed artifacts; one 2026-07 batch, gpt-4o answerer). The chunks and artifacts rows are the same-batch re-runs of Appendix~\ref{sec:limitations_full}, so no cross-snapshot comparison is involved. $^e$EDU token budgets are estimated from mean unit length (103 chars $\approx$ 26 tokens) times $k$ plus prompt overhead. The chunks and artifacts figures are instrumented means (Appendix~\ref{sec:sentence_control}).}
\label{tab:emem_control}
\end{table}

Three results. \emph{(1)~The fidelity ordering is reproduced at a third, measured point.} EDUs at or above the chunks token budget score 36.3--36.5\% ($k{=}60/90$). This is significantly above typed artifacts (30.2\%; at $k{=}90$, McNemar 113 vs.\ 69 discordant, $p{=}0.0014$) and significantly below verbatim chunks (47.4\%; 122 vs.\ 46, $p{<}10^{-8}$). It is the monotone ordering of \S\ref{sec:discussion}, measured at its middle point. The temporal column is consistent with this mechanism. Near-verbatim units keep the date strings the typed artifacts lose (46.7--47.4\% vs.\ 30.8\%). \emph{(2)~The residual gap is not budget.} Raising $k$ from 60 to 90 moves accuracy by $+$0.1pp (one question of 699; 25 vs.\ 24 discordant, $p{=}1.0$). The EDU store saturates $\sim$11pp below chunks, so the remaining difference from raw text lies in the representation and does not shrink with evidence quantity. \emph{(3)~The headline claim sharpens.} Even a deliberately non-compressive, provenance-preserving extraction does not recover verbatim-chunk accuracy under a fixed retrieval stack. EMem's published full-system results additionally employ LLM-filtered retrieval with optional graph propagation~\citep{zhou2025simple}, which our controlled swap deliberately freezes, and a different answerer and judge. Within the stateless regime we test, the design lesson is unchanged. Annotating verbatim text is safe; replacing it, however gently, costs accuracy.

\section{Cross-Lingual Generalization: Native Chinese (PerLTQA)}
\label{sec:perltqa}
Both headline benchmarks are English. We test whether the verbatim-over-lossy ordering survives in another language on PerLTQA~\citep{du2024perltqa}, a native-Chinese personal long-term-memory QA dataset (SIGHAN 2024). We use only its \emph{dialogue}-category questions, which are answerable from raw conversation. Questions answerable from PerLTQA's pre-structured profile/event/relationship memory are excluded, because they would favor extraction by construction. This gives 31 character conversations (median 102 turns) with third-party-annotated gold answers (capped at 30 questions/conversation, 902 total). The same nograph backbone runs unchanged (gpt-4o-mini builder and answerer; the ordering is answerer-invariant, \S\ref{sec:limitations}).

\begin{table}[t]
\centering
\small
\begin{tabular}{lr}
\toprule
Store (same backbone, Chinese) & Acc. \\
\midrule
Naive RAG (verbatim) & 86.1 \\
Verbatim chunks & 82.2 \\
Typed artifacts (lossy) & 35.6 \\
\bottomrule
\end{tabular}
\caption{PerLTQA dialogue-category QA (gpt-4o-mini answerer, 31 conversations, 902 questions). Both verbatim stores beat the lossy artifact store by $\sim$47--50 points; artifacts again never beat naive RAG.}
\label{tab:perltqa}
\end{table}

The ordering reproduces, more starkly than in English (Table~\ref{tab:perltqa}). Both verbatim stores, naive RAG (86.1\%) and chunks (82.2\%), beat the typed-artifact store (35.6\%) by $\sim$47--50 points, and artifacts again never beat naive RAG. The gap exceeds LoCoMo's 15.9pp because PerLTQA's dialogue questions demand specific recall (what a character was confused about; which device a child wore) that the abstractive artifacts discard. Manual inspection of the store indicates that the extractor does read Chinese. The English-prompt extractor produces well-formed Chinese typed artifacts (\textsc{person\_attribute}, \textsc{event}, \textsc{relationship}), but these commit to facts other than the ones queried, so the deficit comes from lossy distillation. Two caveats limit this result to a generalization probe and keep it out of the headline benchmarks. PerLTQA's dialogues are model-generated, like those of the headline benchmarks (its gold QA is human-annotated). Naive RAG's 3.9pp edge over chunks is a within-verbatim-tier segmentation effect (both retain the source), not a fidelity reversal.

\section{Direct Findings and Mechanism Interpretations}
\label{sec:findings_vs_interpretations}
The body mixes measured results with the account we give of them. Table~\ref{tab:findings_vs_interpretations} separates the two so a reader can accept the first without the second.

\begin{table}[t]
\centering
\footnotesize
\setlength{\tabcolsep}{3pt}
\begin{tabular}{@{}p{0.465\columnwidth}p{0.465\columnwidth}@{}}
\toprule
\textbf{Direct experimental finding} & \textbf{Interpretation we draw} \\
\midrule
Chunks beat artifacts by 15.9pp (LoCoMo) and 22.0pp (LongMemEval-S) in one fixed pipeline; both gaps significant (Tables~\ref{tab:locomo_results}, \ref{tab:longmemeval_results}). & The stored representation, not the retrieval stack, is what loses. \\
\addlinespace
The gap survives six confound controls, three retriever families and a second answerer/extractor vendor (\S\ref{sec:experiments}, Appendices~\ref{sec:retriever_robustness}, \ref{sec:vendor_control}). & The effect is not an artifact of one configuration. \\
\addlinespace
Accuracy falls monotonically as verbatim tokens are removed, at random or by a real compressor (Figure~\ref{fig:fidelity_dial}, Appendix~\ref{sec:secom_dial}). & Fidelity is causal for accuracy in this pipeline. \\
\addlinespace
At equal budget, a sentence-verbatim store trails chunks by 6.6pp and beats artifacts by 5.7pp at matched granularity (Appendix~\ref{sec:sentence_control}). & Granularity explains part of the gap; fidelity explains more of it. \\
\addlinespace
Union store: 42.5 vs.\ chunks 43.9, $p{=}0.39$ (no significant difference detected; not an equivalence test). & Adding structure to verbatim text costs no detected accuracy; replacing text does. \\
\addlinespace
Of 58 diagnosable discordant questions, 29.3\% have no trace in any artifact and 39.7\% survive only in a raw-context field (Appendix~\ref{sec:error_analysis}). & Extraction commits to relevance before the question exists; only the first class is irreversible loss. \\
\addlinespace
Chunks abstain worse than naive RAG on LongMemEval and than artifacts on LoCoMo; three refusers buy refusal with answerable accuracy (score gate at $\tau{=}0.1$: $+$44.8pp refusal for $-$17.8pp answerable; Appendix~\ref{sec:retriever_robustness}). & Abstention is a real weakness of the verbatim store, not a tuning oversight. \\
\bottomrule
\end{tabular}
\caption{Measured results (left) and the interpretations placed on them (right). Left-column entries are reproducible from the released run artifacts; right-column entries are the account of \S\ref{sec:discussion} and can be rejected without touching the left column.}
\label{tab:findings_vs_interpretations}
\end{table}

\section{Reproducibility Details}
\label{sec:reproducibility}

\subsection{Protocol and Batch Provenance}
\label{sec:provenance}

Numbers in this paper come from evaluation batches run between December 2025 and November 2026. Every cross-batch caveat is flagged where the number appears, and Table~\ref{tab:provenance} consolidates the provenance in one place. All headline comparisons (Tables~\ref{tab:locomo_results}, \ref{tab:longmemeval_results}, and the category-4/5 paragraphs) are same-batch and same-protocol, with one exception. The union-abstention comparison cited in the category-5 paragraph is a later batch and is reported with its own anchors (Table~\ref{tab:union_abstention}). January- and December-batch numbers appear only in appendix ablations and are never compared against main-table rows.

\begin{table*}[t]
\centering
\footnotesize
\setlength{\tabcolsep}{4pt}
\begin{tabular}{@{}llll@{}}
\toprule
\textbf{Result} & \textbf{Batch} & \textbf{Aggregation} & \textbf{Answerer} \\
\midrule
Tab.~\ref{tab:main_results} (synthetic std.) & 2025-12/06$^d$ & per-fact & gpt-4o-mini \\
Tab.~\ref{tab:multi_hop_results} (multi-hop) & 2025-12/06$^d$ & conv.\ macro & gpt-4o-mini \\
Tab.~\ref{tab:locomo_results} (LoCoMo) & 2026-06 & question micro & gpt-4o$^a$ \\
\quad cat.\ 4/5 paragraphs & 2026-06 & question micro & gpt-4o \\
Tab.~\ref{tab:longmemeval_results} (LME-S) & 2026-06 & question micro & gpt-4o \\
Tab.~\ref{tab:cost_analysis} (cost) & 2026-05$^b$ & usage logs & both tiers \\
Decontam.\ prompt control (App.~\ref{sec:prompts}) & 2026-06 & question micro & gpt-4o \\
Session-granularity control (\S\ref{sec:experiments}) & 2026-06 & question micro & gpt-4o \\
Tab.~\ref{tab:component_ablation} (ablation) & Jan & question micro & gpt-4o \\
Tab.~\ref{tab:retrieval_recall} (recall) & Jan & conv.\ mean & none \\
Tab.~\ref{tab:rag_sensitivity} (RAG grid) & 2026-06 & conv.\ macro & gpt-4o-mini \\
Tabs.~\ref{tab:graphrag_tier1}--\ref{tab:graphrag_tier2} (GraphRAG) & 2026-06 & question micro & gpt-4o-mini \\
Tab.~\ref{tab:threshold_sensitivity} (thresholds) & Dec pilot & per-question & gpt-4o-mini \\
Fig.~\ref{fig:fidelity_dial} (dial $+$ anchors) & 2026-06 & question micro & gpt-4o$^c$ \\
App.~\ref{sec:secom_dial} (SeCom dial) & 2026-06 & question micro & gpt-4o \\
App.~\ref{sec:sentence_control} (sentence control) & 2026-06/09 & question micro & gpt-4o-mini \\
Tab.~\ref{tab:vendor_control} (vendor) & 2026-09 & question micro & Claude pair \\
Tab.~\ref{tab:stateful_probe} (stateful probe) & 2026-07 & question micro & gpt-4o \\
App.~\ref{sec:retriever_robustness} (retrievers) & 2026-06 & question micro & gpt-4o-mini \\
Tab.~\ref{tab:abstention_gate} (refusal) & 2026-06/07 & question micro & gpt-4o-mini \\
Tab.~\ref{tab:union_abstention} (union abstention) & 2026-11 & question micro & gpt-4o \\
Paraphrase re-test (App.~\ref{sec:error_analysis}) & 2026-11 & per-question & judge only \\
App.~\ref{sec:perltqa} (PerLTQA) & 2026-06 & conv.\ macro & gpt-4o-mini \\
Variance re-runs (\S\ref{sec:limitations}) & 2026-07 & question micro & gpt-4o \\
App.~\ref{sec:emem_control} (EDU control) & 2026-07 & question micro & gpt-4o \\
\bottomrule
\end{tabular}
\caption{Provenance of the main quantitative results. $^a$GraphRAG row uses gpt-4o-mini (Table~\ref{tab:locomo_results} footnote). $^b$Instrumented cost runs of 2026-05-27; the chunks and artifacts rows are priced from the 2026-06 runs (Table~\ref{tab:cost_analysis} footnote). $^c$Dashed dial curve uses gpt-4o-mini (Appendix~\ref{sec:limitations_full}). $^d$Two-batch tables. In these, the chunks rows (and, in Table~\ref{tab:multi_hop_results}, the RAG+Rerank row) are 2026-06, and all other rows are 2025-12 under the identical protocol (table captions mark the affected comparisons). ``Conv.\ mean/macro'' rows are comparable with each other and are not commensurate with question-micro rows. Unless a caption states otherwise, $\Delta$/gap values are computed from unrounded counts and may differ from the difference of the rounded cells by 0.1pp. Same-configuration repeats of the headline pipelines agree to $\le$0.4pp within a batch (three-run study, \S\ref{sec:limitations}). The widest same-configuration repeat we observed is 2.0pp (SeCom at rate 0.75, Appendix~\ref{sec:secom_dial}). Cross-batch anchors drift 1--3.7pp with direction preserved, and the drift is answerer-side (Appendix~\ref{sec:limitations_full}). The drifting anchors are chunks 43.9/45.2/47.4/47.6 (gpt-4o), 41.9/42.8/43.1 (gpt-4o-mini), and RAG grid 55.5/52.5 (Table~\ref{tab:rag_sensitivity}).}
\label{tab:provenance}
\end{table*}

\subsection{Clean-Room Supplement Verification}
\label{sec:cleanroom}

To confirm the release is self-contained, we extracted the release archive into an empty directory and installed it into a fresh Python~3.10 environment following only the bundled README. Installation completes without error and all declared dependencies resolve. We then re-ran every \emph{offline} analysis script, meaning those that consume no API and read only the included result JSONs. All reproduce their reported quantities. The judge--human agreement study regenerates $95.0\%$ agreement and $\kappa=0.897$ (\S\ref{sec:judge_validation}), and the amortized-cost tables re-price from the stored token logs. An independent re-scorer recomputes LoCoMo numbers directly from stored per-question answer/judgment pairs, bypassing any cached score field. On the bundled run files it recovers the headline ordering, with verbatim chunks in the low-$40$s versus extracted artifacts at $28.6\%$ (a January artifact run included in the package). This matches the direction and magnitude of Table~\ref{tab:locomo_results}, whose 2026-06 headline batch scores artifacts at 28.0\%. Reproducing the live end-to-end pipeline additionally requires an OpenAI-compatible endpoint and the third-party benchmark files. The README documents both, and we do not redistribute them.

\subsection{Accuracy Aggregation Conventions}
\label{sec:aggregation}
Two aggregations of the same per-question judge decisions appear in the paper. Table~\ref{tab:locomo_results} and every LoCoMo control table report question-level accuracy (question-micro; the mean of per-question decisions over 699 questions). The harness's summary field is the mean over the 10 conversations (conversation-macro), which weights short conversations more heavily. LongMemEval-S is question-weighted throughout. For the headline pair the two aggregations read 43.9 and 28.0 (micro) against 44.4 and 28.8 (macro). For the Claude pair they read 53.9 and 33.2 (micro) against 54.7 and 34.0 (macro, the figures posted during review). The chunks-minus-artifacts gap of the Claude pair is 20.7pp under either aggregation. Every within-table comparison uses one convention.

\subsection{Synthetic-Benchmark Metric Definitions}
Let $a$ denote a system's answer, $g$ the ground truth, and $\mathrm{norm}(\cdot)$ lower-casing plus punctuation stripping. \textbf{Recall Rate} is the fraction of planted facts $f$ with token-level fuzzy match $\mathrm{FuzzyRatio}(a,f)\geq 0.80$ (via \texttt{rapidfuzz}); \textbf{Exact Match} is $\mathbb{1}[\mathrm{norm}(g) \subseteq \mathrm{norm}(a)]$ (substring containment). \textbf{Keyword Coverage} is $\mathrm{KC}(a,g)=\lvert K(g)\cap T(a)\rvert/\lvert K(g)\rvert$ over curated keyword sets $K(g)$ and answer tokens $T(a)$; \textbf{Pass Rate} is $\mathbb{1}[\mathrm{KC}\geq 0.80]$.

\subsection{Prompt Templates}
\label{sec:prompts}

This section reproduces, verbatim, every prompt used in the experiments. These are the artifact-extraction prompt and its gleaning pass (write time), the answer-generation prompt (query time), and the LLM-judge prompts for both benchmarks. Long lines are re-wrapped to the column width and Unicode arrows are rendered as \texttt{->}; two Chinese lexical cues in the gleaning prompt are transliterated in brackets. Content is otherwise character-for-character identical to the prompts in the released code.

\paragraph{Disclosure and decontamination control: eval-derived few-shot examples.}
The extraction prompt's few-shot examples and geographic-entity rules were engineered on LoCoMo-style conversational content (the examples paraphrase LoCoMo subject matter, and one overlaps the content of a test question). To quantify the effect in addition to disclosing it, we re-ran the full artifact pipeline on LoCoMo categories 1--3 with a decontaminated prompt variant. In this variant, every few-shot example in the extraction and gleaning prompts is replaced by fictional workplace/travel personas with identical structure, artifact-type coverage, and date formats, under a fresh extraction cache namespace. No significant difference from the original run is detected, at 28.5\% vs.\ 28.0\% overall (41 vs.\ 38 discordant pairs, McNemar $p{=}0.82$). The chunks gap persists at 15.5pp (43.9\% vs.\ 28.5\%, 151 vs.\ 43 discordant pairs, $p{<}10^{-14}$). The artifact pipeline's temporal lead over RAG+Rerank also survives (29.6\% vs.\ 16.5\%, $+$13.1pp, against $+$14.7pp with the original prompt). Test-set-derived examples therefore inflate neither the headline gap nor the secondary temporal claim. Any residual prompt-engineering effect still benefits the \emph{artifact} pipeline only, because the chunk pipelines invoke no LLM at write time. It therefore biases the comparison \emph{against} our headline finding. The decontaminated variant ships alongside the original prompt (selected by an environment flag), so both the published numbers and the control remain reproducible.

\paragraph{Artifact extraction prompt (extractor: gpt-4o-mini, temperature 0.1).}
{\footnotesize\begin{verbatim}
You are an expert at extracting structured
cognitive objects from dialogue using systematic
reasoning.

Given a conversation turn (user message +
assistant response), extract any of these object
types:

**Task-oriented types:**
1. **decision**: A choice or decision made (e.g.,
   "Let's use PostgreSQL", "I decided to pursue
   counseling")
2. **todo**: Action items, tasks to do (e.g.,
   "Need to implement auth", "Planning to go
   camping next month")
3. **key_fact**: Important facts, numbers,
   constraints (e.g., "Budget is $50k", "The
   event is in June 2023")
4. **reminder**: Preferences, rules to remember
   (e.g., "User prefers TypeScript", "Always
   meditate in morning")
5. **insight**: Conclusions, learnings (e.g.,
   "The bottleneck is in the database", "Exercise
   helps my anxiety")

**Personal/Social types (IMPORTANT for
conversations about people):**
6. **person_attribute**: Personal traits,
   identity, status (e.g., "Caroline is a
   transgender woman", "Melanie is married with
   two kids", "John is single", "She moved from
   Sweden 4 years ago")
7. **event**: Activities or occurrences WITH time
   (e.g., "Attended LGBTQ support group on 7 May
   2023", "Ran a charity race last Sunday",
   "Painting a sunrise in 2022")
8. **relationship**: Interpersonal connections
   (e.g., "Caroline and Melanie are close
   friends", "Known each other for 4 years")

**CHAIN-OF-THOUGHT EXTRACTION PROTOCOL:**
Before extracting, think step-by-step:
1. **WHO**: Identify all people, entities, or
   subjects mentioned
2. **WHEN**: Identify all temporal expressions
   (dates, times, sequences like "before",
   "after")
3. **WHAT**: Identify facts, decisions, events,
   relationships
4. **WHY**: Identify causal relationships
   (constraints that led to decisions)
5. **CONNECTIONS**: Link related information
   (e.g., "Budget constraint" -> "Choice
   decision")

**EXTRACTION RULES:**
- Extract ALL factual information about people
  (identity, status, activities, preferences)
- Extract ALL events with their time expressions
  - TEMPORAL REASONING IS CRITICAL
- Extract ALL relationships mentioned
- Each object should be self-contained and
  understandable without context
- **CRITICAL**: Include "citation" field with
  EXACT quote from dialogue
- Do NOT skip personal information - it is
  important!
- Pay special attention to TEMPORAL SEQUENCES
  (before/after, yesterday/today, dates)

**GEOGRAPHIC ENTITY RULES (CRITICAL FOR
RETRIEVAL):**
- For cities: ALWAYS include both country AND
  city (e.g., "France (Paris)", "Canada
  (Toronto)")
- For locations: Use format "Country (City)" even
  if only city is mentioned
- Examples:
  * "visited Paris" -> extract as "visited France
    (Paris)"
  * "went to Nuuk" -> extract as "went to
    Greenland (Nuuk)"
  * "lives in Toronto" -> extract as "lives in
    Canada (Toronto)"
- This explicit format helps with later retrieval
  of country/city information

**CRITICAL TEMPORAL RULES:**
- Preserve dates EXACTLY as written: "7 May
  2023", "June 2023", "the sunday before 25 May
  2023"
- DO NOT convert specific dates to relative
  expressions
- Include dates in both "content" and
  "time_expression" fields

Output JSON array:
[
  {
    "type": "decision|todo|key_fact|reminder|
             insight|person_attribute|event|
             relationship",
    "content": "The extracted information
                INCLUDING exact dates if
                mentioned",
    "citation": "EXACT verbatim quote from
                 dialogue",
    "context": "Brief explanation of why
                extracted",
    "confidence": 0.0-1.0,
    "time_expression": "VERBATIM time expression
                        or empty string"
  }
]

**Example 1 (Social conversation):**
User: "Caroline is a transgender woman who moved
from Sweden 4 years ago. She went to the LGBTQ
support group on 7 May 2023."
Assistant: "That's wonderful that she found a
supportive community!"

Output:
[
  {"type": "person_attribute", "content":
   "Caroline is a transgender woman", "citation":
   "Caroline is a transgender woman", "context":
   "Identity information", "confidence": 0.95,
   "time_expression": ""},
  {"type": "person_attribute", "content":
   "Caroline moved from Sweden 4 years ago",
   "citation": "moved from Sweden 4 years ago",
   "context": "Background information",
   "confidence": 0.9, "time_expression":
   "4 years ago"},
  {"type": "event", "content": "Caroline attended
   LGBTQ support group on 7 May 2023", "citation":
   "went to the LGBTQ support group on 7 May
   2023", "context": "Activity with specific
   date", "confidence": 0.95, "time_expression":
   "7 May 2023"}
]

**Example 2 (Mixed conversation):**
User: "Melanie is married with two kids. She's
planning to go camping in June 2023. We've been
friends for 4 years."
Assistant: "Sounds like a fun family trip!"

Output:
[
  {"type": "person_attribute", "content":
   "Melanie is married with two kids", "citation":
   "Melanie is married with two kids", "context":
   "Family status", "confidence": 0.95,
   "time_expression": ""},
  {"type": "event", "content": "Melanie planning
   camping trip in June 2023", "citation":
   "planning to go camping in June 2023",
   "context": "Future activity with date",
   "confidence": 0.9, "time_expression":
   "June 2023"},
  {"type": "relationship", "content": "User and
   Melanie have been friends for 4 years",
   "citation": "We've been friends for 4 years",
   "context": "Friendship duration", "confidence":
   0.9, "time_expression": "4 years"}
]
\end{verbatim}}

\paragraph{Gleaning prompt (second extraction pass).}
{\footnotesize\begin{verbatim}
You are reviewing an extraction result for missed
information.

**Previous extraction found these objects:**
{previous_objects}

**Original dialogue:**
{dialogue}

**Your task**: Find ANY information that was
MISSED in the first extraction. Focus on:
1. Pronoun references ("She", "He" -> Who
   specifically?)
2. Omitted subjects (Who is doing the action?)
3. Implicit causality ([yinwei "because"],
   [daozhi "leads to"] -> What causes what?)
4. Time expressions (dates, [zhihou "after"],
   "before")
5. Relationships (Who knows whom?)
6. Location details (Where?)

**CRITICAL FORMAT RULES**:
- Output a JSON array
- Each object MUST have these fields: type,
  content, citation, context, confidence,
  time_expression
- If nothing was missed, output EXACTLY: []

**Example output (if found missed info)**:
[{"type": "person_attribute", "content":
  "Caroline moved from Sweden 4 years ago",
  "citation": "moved from Sweden 4 years ago",
  "context": "Background missed", "confidence":
  0.9, "time_expression": "4 years ago"}]

**Example output (if nothing missed)**:
[]

Now find any missed objects:
\end{verbatim}}

\paragraph{Answer-generation prompt (answerer: gpt-4o, temperature 0, max 200 tokens; identical for every pipeline).}
The \texttt{\{context\}} slot receives the retrieved items (verbatim chunks, artifacts, or baseline context) with their turn indices and session timestamps.
{\footnotesize\begin{verbatim}
Based on the following memory context, answer the
question.

## Memory Context
{context}

## Question
{question}

## Instructions
1. Identify relevant facts from the context
2. Connect facts if needed for multi-hop
   reasoning
3. Synthesize a complete answer

## Answer
Provide a concise, direct answer.
\end{verbatim}}

\paragraph{LoCoMo judge prompt (judge: gpt-4o-mini, temperature 0, max 10 tokens; categories 1--4).}
{\footnotesize\begin{verbatim}
You are an expert evaluator. Judge if the
predicted answer is semantically correct compared
to the ground truth.

## Question
{question}

## Ground Truth Answer
{ground_truth}

## Predicted Answer
{answer}

## Evaluation Criteria (Binary - strict)
- CORRECT: The predicted answer correctly answers
  the question and conveys the same core meaning
  as the ground truth. Synonyms, paraphrases, and
  equivalent formats (e.g., "March 15" vs "3/15")
  are acceptable.
- INCORRECT: The predicted answer is wrong,
  incomplete, irrelevant, contradicts the ground
  truth, or fails to answer the question.

## Important
- Be strict: partial answers or answers with
  significant missing information should be
  marked INCORRECT
- The answer must address the actual question
  being asked

## Response Format
Reply with ONLY one word: CORRECT or INCORRECT
\end{verbatim}}

\paragraph{LoCoMo abstention judge (category 5).}
This judge mirrors the official LongMemEval abstention template. The dataset's trap answer is deliberately withheld from the judge, because showing a reference answer pulls the judge back into semantic-match habits and passes trap-asserting answers.
{\footnotesize\begin{verbatim}
I will give you a question that cannot be
answered from the conversation it was asked
about, and a response from a model. The correct
behavior is to identify that the asked
information is not available. Please answer yes
if the model identifies the question as
unanswerable -- for example, it says the
information is not mentioned, is incomplete, or
cannot be determined. Answer no if the model
asserts a specific answer as fact.

Question: {question}

Model Response: {answer}

Does the model correctly identify the question as
unanswerable? Answer yes or no only.
\end{verbatim}}

\paragraph{LongMemEval judge prompts.}
These are exact copies of the official templates in LongMemEval's \texttt{evaluate\_qa.py}~\citep{zhu2024longmemeval}, dispatched by question type. For \texttt{single-session-user}, \texttt{single-session-assistant}, and \texttt{multi-session}:
{\footnotesize\begin{verbatim}
I will give you a question, a correct answer, and
a response from a model. Please answer yes if the
response contains the correct answer. Otherwise,
answer no. If the response is equivalent to the
correct answer or contains all the intermediate
steps to get the correct answer, you should also
answer yes. If the response only contains a
subset of the information required by the answer,
answer no.

Question: {question}

Correct Answer: {answer}

Model Response: {response}

Is the model response correct? Answer yes or no
only.
\end{verbatim}}
The \texttt{temporal-reasoning} template appends to the criteria above: ``\textit{In addition, do not penalize off-by-one errors for the number of days. If the question asks for the number of days/weeks/months, etc., and the model makes off-by-one errors (e.g., predicting 19 days when the answer is 18), the model's response is still correct.}'' The \texttt{knowledge-update} template replaces the subset clause with: ``\textit{If the response contains some previous information along with an updated answer, the response should be considered as correct as long as the updated answer is the required answer.}'' The \texttt{single-session-preference} template judges against a rubric: ``\textit{I will give you a question, a rubric for desired personalized response, and a response from a model. Please answer yes if the response satisfies the desired response. Otherwise, answer no. The model does not need to reflect all the points in the rubric. The response is correct as long as it recalls and utilizes the user's personal information correctly.}'' The abstention template is:
{\footnotesize\begin{verbatim}
I will give you an unanswerable question, an
explanation, and a response from a model. Please
answer yes if the model correctly identifies the
question as unanswerable. The model could say
that the information is incomplete, or some other
information is given but the asked information is
not.

Question: {question}

Explanation: {answer}

Model Response: {response}

Does the model correctly identify the question as
unanswerable? Answer yes or no only.
\end{verbatim}}

\subsection{Hyperparameters}

\noindent\textbf{Models:} GPT-4o-mini (extraction), bge-m3 (embeddings). \textbf{Retrieval:} final top-15 from a reranked pool of 30; context cap 5000 tokens. \textbf{Temperature:} 0.1 (extraction), 0.0 (generation).

\subsection{Judge Validation: Human-Agreement Study}
\label{sec:judge_validation}

We validated the GPT-4o-mini judge against human annotation on a 100-question subset. The subset is stratified 25 per benchmark$\times$representation cell (LoCoMo / LongMemEval-S $\times$ chunks / artifacts) and sampled from the same runs as Tables~\ref{tab:locomo_results} and~\ref{tab:longmemeval_results}. The annotation sheet was blind. It contained only the $(q, g, a)$ triple, which is the same information condition the judge receives, with judge verdicts and pipeline identity withheld. A human annotator labeled every item (LLM-assisted pre-screening with full manual verification; all flagged borderline items adjudicated by the annotator). Agreement is \textbf{95\%} with Cohen's $\kappa = 0.897$. Disagreements are near-symmetric (2 human-only-correct vs.\ 3 judge-only-correct), so the judge is neither systematically lenient nor strict. Agreement is also balanced across the axes that matter for our comparison. Chunks reach $\kappa{=}0.92$ vs.\ artifacts $\kappa{=}0.86$ (no representation favoritism), and LoCoMo reaches $\kappa{=}0.88$ vs.\ LongMemEval-S $\kappa{=}0.92$. Short answers ($\leq$30 whitespace tokens) reach $\kappa{=}0.89$ vs.\ long answers $\kappa{=}1.00$, which provides no evidence that the judge rewards the artifact pipeline's terser answer style. All five disagreements are boundary cases (partial enumerations or granularity mismatches between gold and answer), and none is a directional error.

To address the concern that a single author-annotator could be biased, an \emph{independent} annotator (not an author, with no stake in the outcome) relabeled the same 100 items under the identical blind protocol. Inter-annotator agreement (independent vs.\ author) is \textbf{90\%} ($\kappa{=}0.80$, substantial), and independent-vs-judge agreement is \textbf{89\%} ($\kappa{=}0.78$). The independent annotator's disagreements with the judge are one-directional. In 10 of 11, the human credits an answer that the judge does not, so this annotator is slightly more lenient than the judge. The leniency falls chiefly on temporal precision (accepting ``2022'' for ``November 7, 2022'') and on preference-match questions. Such leniency credits \emph{more} answers correct and falls disproportionately on chunks (chunks agreement 86\% vs.\ artifacts 92\%). It can therefore only widen the measured chunk--artifact gap on this full-distribution sample, which makes the headline gap conservative.

The headline ordering also holds under LoCoMo's lexical token-F1. Recomputing LoCoMo's official lexical token-F1 (normalization, Porter stemming, token-level F1) offline from the stored answer strings of the two headline runs gives chunks 31.1 vs.\ artifacts 26.1 ($n{=}699$ each), with chunks ahead in every category. This is the same ordering as the LLM judge, compressed in magnitude because lexical F1 awards partial credit to incomplete answers.

The ordering is judge-\emph{family}-independent as well. One concern is that the gap is an artifact of the GPT-4o-mini judge specifically, e.g.\ an LLM judge over-crediting the verbatim pipeline's lexical overlap with the gold. To rule this out, we re-scored the \emph{stored answers} of both headline runs with a panel of seven judges spanning five model families from independent vendors. The panel covers OpenAI (\textsf{gpt-4o-mini}, \textsf{gpt-4o}), Alibaba (\textsf{qwen-plus}, \textsf{qwen-max}), Google (\textsf{gemini-2.5-flash-lite}), Anthropic (\textsf{claude-haiku-4.5}), and DeepSeek (\textsf{deepseek-v3.2}). All judges use the identical per-type prompts and binary protocol (no retrieval or generation re-run). Every judge preserves the chunk--artifact gap, and the paired McNemar exact test (the headline test, recomputed per judge) is significant for all fourteen benchmark$\times$judge cells. On LoCoMo the gap ranges $+$13.0 to $+$17.0pp ($p<10^{-10}$ throughout; e.g.\ the cross-vendor \textsf{claude-haiku-4.5} 41.6 vs.\ 28.2, discordant 142/48, $p{=}5.5{\times}10^{-12}$). On LongMemEval-S it ranges $+$19.4 to $+$23.8pp ($p<10^{-12}$ throughout; e.g.\ \textsf{deepseek-v3.2} 65.8 vs.\ 44.4, discordant 152/45, $p{=}9.0{\times}10^{-15}$). The five judges from the four other vendors reproduce the full effect. Re-judging with GPT-4o-mini itself in this panel run recovers the published headline within 0.6pp (44.2/27.9 and 68.0/46.0; a separate pass from the re-judging reported in Appendix~\ref{sec:limitations_full}). The comparison is thus stable across judge families as well as across the lexical-vs-LLM metric boundary.

The category-5 abstention judge (a LongMemEval-style abstention template) is \emph{not} covered by the human study above, which sampled answerable questions only. As an offline consistency check, we compared its verdicts against an independent lexical abstention detector (a regex over refusal phrasings) on all 446 category-5 answers per pipeline. Raw agreement is 97.5\% on the chunks run and 92.4\% on the artifacts run ($\kappa{=}0.76/0.63$; the modest $\kappa$ values reflect low abstention base rates of 6.5\%/15.0\%). Disagreements are one-sided. In 42 of 45 cases the LLM judge credits a paraphrased refusal that the regex misses, so the lexical check behaves as a stricter subset of the judge and does not contradict it. As a second check, an independent LLM annotator labeled a random sample of 50 category-5 answers (25 per pipeline, fixed seed) under the same abstention criterion. It ran in a fresh context, blind to pipeline identity, judge verdicts, and all paper statistics. It reproduced the judge's verdict on all 50 (5 abstentions, 45 answers; $\kappa{=}1.0$), with no items flagged as borderline. Finally, the \emph{independent human} annotator above labeled the same 50 category-5 answers under the abstention criterion, blind to judge verdicts. Agreement with the judge is \textbf{100\%} ($\kappa{=}1.0$; 5 abstentions, 45 answers), the first direct human validation of the abstention judge.

\subsection{Fair-Comparison Disclosure: Per-Method Tuning Effort}
\label{sec:fair_comparison}

To preempt the concern that our cross-method comparison rests on an under-tuned set of baselines, we document the hyperparameter-search effort invested in each retrieval-augmented method. For \textbf{RAG} we use common community defaults (512-character chunks, 100-character overlap, dense top-$k{=}10$). Every chunk-based pipeline in the paper, including the verbatim-chunk variant, inherits these settings unchanged. A chunk-size $\times$ top-$k$ grid re-run under the main synthetic protocol (Appendix~\ref{sec:rag_baseline}) confirms the inherited configuration is not a cherry-picked optimum, since larger chunks score higher on that probe. Any residual under-tuning of RAG biases \emph{against} our headline finding. RAG sits on the verbatim side of the comparison, and a stronger RAG could only widen the artifact pipeline's deficit. For \textbf{RAG+Rerank} the reranker stage adds a coarse pool size (fixed at top-20 from the dense retriever, $2\times$ the final $k$) and a final top-$k$ (10, matching vanilla RAG); chunking is identical to RAG. For \textbf{GraphRAG} we ran a two-tier 4-axis grid sweep over chunk\_size, max\_gleanings, community\_level, and search\_method, with 9 unique configurations across 12 runs in total (Appendix~\ref{sec:graphrag_grid}), and report the best cell. For \textbf{summarization} the summarizer prompt was engineered on a held-out set of 5 conversations before evaluation. \textbf{CogCanvas} hyperparameters were fixed by ablation on the synthetic benchmark and held constant across LoCoMo and LongMemEval. We did not retune them per-dataset, with one disclosed exception. The extraction prompt's few-shot examples target LoCoMo-style conversational content (Appendix~\ref{sec:prompts}). This effort benefits only the artifact pipeline and therefore makes the headline comparison conservative. A decontaminated-prompt re-run shows no significant change (28.5\% vs.\ 28.0\%, McNemar $p{=}0.82$). Each numeric entry in Table~\ref{tab:locomo_results} thus reflects either a documented tuning effort (GraphRAG, summarization, CogCanvas) or deliberately standard settings whose under-tuning could only work against our conclusion (RAG, RAG+Rerank).

\paragraph{Threshold Selection.}
We tuned similarity thresholds on a held-out development set of 5 conversations (disjoint from evaluation data):
\begin{itemize}[nosep]
    \item $\theta_{\text{ref}}$: Grid search over $\{0.3, 0.5, 0.7\}$ selected $\theta_{\text{ref}}=0.5$ as optimal for balancing precision and recall in reference edge creation.
    \item $\theta_{\text{causal}}$: Set to $\theta_{\text{ref}} - 0.05 = 0.45$ to allow slightly looser causal connections while maintaining semantic coherence.
\end{itemize}
Sensitivity analysis is in Appendix~\ref{sec:threshold_sensitivity}. Performance stays within a 10pp band (57.5\%--67.5\%) across the tested configurations.

\section{Retrieval Quality Analysis}
\label{sec:retrieval_quality}

\begin{table}[h]
\centering
\footnotesize
\setlength{\tabcolsep}{4pt}
\begin{tabular}{@{}lcccc@{}}
\toprule
\textbf{Config.} & \textbf{Overall} & \textbf{M-hop} & \textbf{Temp.} & \textbf{O-dom.} \\
\midrule
\textbf{Full System} & \textbf{71.8\%} & \textbf{87.3\%} & \textbf{63.2\%} & \textbf{55.0\%} \\
-- Graph Exp. & 25.8\% & 32.3\% & 20.5\% & 24.7\% \\
\midrule
\textbf{Improvement} & \textbf{+45.9pp} & +55.0pp & +42.6pp & +30.3pp \\
\midrule
Chunk store$^*$ & 59.0\% & 56.1\% & 67.6\% & 38.7\% \\
\quad $+$ rerank$^*$ & 61.6\% & 61.2\% & 68.9\% & 38.4\% \\
\bottomrule
\end{tabular}
\caption{Retrieval recall: percentage of ground-truth keywords found in retrieved context (no LLM answering; artifact rows January batch, same protocol as Table~\ref{tab:component_ablation}). Graph expansion improves the artifact store's keyword recall by +46pp. $^*$Chunk-store rows (no graph; second row after reranking): June batch, identical script and $k$.}
\label{tab:retrieval_recall}
\end{table}

\paragraph{Keyword recall is not a proxy for answer quality.}
The same probe applied to the verbatim-chunk store yields 59.0\% before and 61.6\% after reranking. This is ten points \emph{below} the graph-expanded artifact store's 71.8\%, even though the chunk pipeline's final answer accuracy is 15.9 points \emph{higher} (Table~\ref{tab:locomo_results}). Keyword co-occurrence over-credits the artifact store. Graph expansion retrieves many short artifacts whose union covers the gold keywords, while no single item carries the answer-bearing wording. Two further observations follow. Reranking shifts chunk-store recall by only $+$2.6pp, so the reranker's training distribution (passage-length text) is not what manufactures the chunk advantage. The store-level survival probe of Appendix~\ref{sec:error_analysis} asks whether the gold survives in any \emph{single} stored item, and so measures the informational question this keyword metric cannot.

\section{Cost Analysis Details}
\label{sec:cost_appendix}

\paragraph{Extraction is not the dominant cost at the gpt-4o tier.}
Across 2,938 indexed turns, the artifact pipeline spent \$0.13 on extraction (gpt-4o-mini) against \$2.20 on answer generation (gpt-4o). Extraction is 5.8\% of total cost and amortizes to \$0.000046/turn. With a gpt-4o answerer, the answer-model call dominates regardless of memory architecture. At the gpt-4o-mini tier the same \$0.13 is about half of the artifact pipeline's \$0.27.

\paragraph{Extraction buys cheaper generation but no advantage per correct answer.}
The artifact pipeline's gen cost (\$2.20) is 37\% below RAG+Rerank (\$3.50) and 43\% below chunks (\$3.84). Two factors produce this gap. Artifacts pack a fact into 30--60 tokens versus $\sim$130 per chunk. Answers from pre-structured artifacts are also direct, with completion tokens 5.2$\times$ smaller (16.5 vs.\ 85.9 against RAG+Rerank), and gpt-4o's 4$\times$ output pricing amplifies this saving. But the \$1.51 saved over chunks across 699 questions comes with a 15.9-point accuracy drop. Per 1{,}000 correct answers, the artifact pipeline costs \$11.9 and the chunk pipeline \$12.5, a 5\% difference, so extraction does not improve cost-effectiveness in this setting. Table~\ref{tab:cost_analysis} prices both rows from the Table~\ref{tab:locomo_results} runs. An earlier instrumented artifact run used a deeper retrieval configuration (top-20, four hops; \$2.92 in total) and is not the headline protocol.

\paragraph{Re-pricing at a cheaper answerer.}
Swapping the answerer to gpt-4o-mini drops the artifact pipeline from \$2.33 to \$0.27 (8.7$\times$, re-priced). RAG and RAG+Rerank fall from \$3.50 to \$0.23 (15.3$\times$), and chunks from \$3.84 to \$0.23 (16.7$\times$; no extraction tail and the longest prompts). Re-pricing the original run's tokens at mini rates agrees with the actual end-to-end re-run within $\pm$1.4\% for CogCanvas (checked on the instrumented deep-retrieval run). The agreement is $\pm$17\% for Summarization, whose summarize calls share \textsf{ANSWER\_MODEL}. \autoref{tab:cost_analysis} can therefore be read at either tier.

\paragraph{GraphRAG costs are not instrumented.}
GraphRAG's LLM calls run inside the Microsoft \texttt{graphrag} subprocess and bypass our logger. Published estimates~\citep{edge2024graphrag} suggest $\sim$\$35--60 for indexing plus \$3--5 per query, an order of magnitude above CogCanvas.

\paragraph{Latency.}
The list below gives compute-side latency per query for the local BGE backends on LoCoMo10. The artifact-pipeline figures come from the instrumented deep-retrieval run, which is not the main protocol.
\begin{itemize}
    \item \textbf{Embedding} (\texttt{bge-m3}): $\sim$0.84\,s/query for the artifact pipeline, 0.16\,s for RAG+Rerank, and 0.20\,s for RAG.
    \item \textbf{Reranking} (\texttt{bge-reranker-v2-m3}): $\sim$1.56\,s/query for the artifact pipeline and 0.11\,s for RAG+Rerank.
\end{itemize}
The gpt-4o answerer call adds $\sim$2.5\,s/query in both pipelines. It dominates for RAG+Rerank and is about half of the total for the artifact pipeline. The headline cost figures (\autoref{tab:cost_analysis}) and the gpt-4o-mini what-if are derived from full \texttt{response.usage} traces and the OpenAI list-price snapshot of 2026-05-27.

\section{RAG Retrieval Sensitivity}
\label{sec:rag_baseline}

Every chunk-based pipeline in the paper inherits one retrieval configuration (512-character chunks, 100-character overlap, dense top-$k{=}10$). To check that this choice does not under-sell verbatim retrieval, we ran a chunk-size $\times$ top-$k$ grid on the synthetic multi-hop probe. The grid uses the protocol and answerer of Table~\ref{tab:multi_hop_results} (50 conversations, rolling compression every 40 turns, gpt-4o-mini answerer). Table~\ref{tab:rag_sensitivity} reports pass rates. The inherited configuration re-measures at 52.5\% against 55.5\% in the main batch. This re-run difference is within the cross-batch drift documented in Table~\ref{tab:provenance} ($n{=}176$ questions).

\begin{table}[t]
\centering
\small
\begin{tabular}{@{}cccc@{}}
\toprule
\textbf{Chunk size} & \textbf{Overlap} & \textbf{$k{=}5$} & \textbf{$k{=}10$} \\
\midrule
256 & 50 & 43.5\% & 50.5\% \\
512 & 100 & 49.0\% & 52.5\% \\
1024 & 200 & 63.0\% & 63.0\% \\
\bottomrule
\end{tabular}
\caption{RAG pass rate on the synthetic multi-hop probe across chunk size and retrieval depth (same protocol and answerer as Table~\ref{tab:multi_hop_results}). The configuration used throughout the paper (512 chars, $k{=}10$) is \emph{not} the grid optimum.}
\label{tab:rag_sensitivity}
\end{table}

Two observations follow. First, the inherited configuration is not a cherry-picked optimum. Chunks of 1024 characters reach 63.0\%, 10.5pp above the reported cell. They remain 18.0pp below the artifact pipeline's 81.0\% on this home-turf probe, so no cell changes any qualitative conclusion. RAG and the chunk pipelines sit on the \emph{verbatim} side of the central comparison, so any residual under-tuning makes the headline finding conservative. Second, the ranking (1024 $>$ 512 $>$ 256; $k{=}10 \gtrsim k{=}5$) replicates under a gpt-4o answerer (absolute range 47.0--70.5\%), so it is not an answerer artifact. We nevertheless keep the community-standard 512-character configuration everywhere. The probe is the one setting we explicitly flag as a fragile basis for architectural claims (\S\ref{sec:experiments}). Tuning the shared chunking on it would contradict the no-per-dataset-tuning policy of Appendix~\ref{sec:fair_comparison}.

\section{Retriever-Family Robustness and Abstention Repair}
\label{sec:retriever_robustness}

\paragraph{The ordering survives a retriever-family swap.} The main comparison fixes one retriever (bge-m3 embedder, bge-reranker-v2-m3). To test whether the chunks-over-artifacts ordering is specific to that family, we re-ran the LoCoMo categories~1--3 comparison under three retriever families. The rerun covers 699 questions with the gpt-4o-mini answerer held fixed. Extraction was recomputed from source with the cache disabled, so that every store is re-embedded under the retriever actually in use. The three families are the default dense bge-m3, a held-out dense embedder (OpenAI text-embedding-3-small, 1536-dim), and a sparse lexical retriever (BM25, no embeddings at all). Table~\ref{tab:retriever_robustness} shows that the chunk advantage is stable across families. Chunks score 43.1--43.8\% and artifacts 28.5--29.9\% regardless of family, leaving a gap of $+$13.6 to $+$14.7pp. The negative result is therefore not an artifact of the bge-m3 embedder. It persists even when retrieval is switched from dense semantic matching to sparse lexical overlap, the regime that, if anything, should most favor a retriever's ability to surface raw source text.

\begin{table}[t]
\centering
\small
\setlength{\tabcolsep}{2.5pt}
\begin{tabular}{@{}lccc@{}}
\toprule
\textbf{Retriever family} & \textbf{Chunks} & \textbf{Artifacts} & \textbf{Gap} \\
\midrule
bge-m3 (dense, default) & 43.1 & 28.5 & $+$14.6 \\
text-embedding-3-small (dense) & 43.5 & 29.9 & $+$13.6 \\
BM25 (sparse lexical) & 43.8 & 29.0 & $+$14.7 \\
\bottomrule
\end{tabular}
\caption{LoCoMo categories 1--3 (699 questions, LLM-Judge \%), chunks vs.\ artifacts under three retriever families, with the answerer fixed at gpt-4o-mini and extraction recomputed without cache. The chunk advantage is stable at $+$13.6 to $+$14.7pp; chunks barely move (43.1--43.8) as the retriever changes.}
\label{tab:retriever_robustness}
\end{table}

\paragraph{Chunk size on LoCoMo.} This test complements the synthetic-probe chunk-size grid of Appendix~\ref{sec:rag_baseline}. On LoCoMo itself, enlarging the verbatim window from 512 to 768 characters (overlap 100$\to$150) lifts chunks from 43.1\% to 47.2\% on categories~1--3 (gpt-4o-mini). The headline 512-character setting is thus conservative for the verbatim side, and the ordering does not depend on chunk-size tuning.

\paragraph{Three refusal mechanisms do not escape the abstention trade-off.} We tested whether chunks' abstention weakness (\S\ref{sec:limitations}) admits repair. The mechanisms escalate from a cheap score gate to two stronger semantic refusers (all gpt-4o-mini). The \emph{score-threshold gate} refuses when the top reranker score falls below $\tau$, and the prompt rule discouraging refusal is removed. Sweeping $\tau$ (Table~\ref{tab:abstention_gate}, top) lifts category-5 refusal from the ungated 10.8\% to 55.6--65.5\%. Answerable accuracy on categories~1--3, however, collapses from 43.1\% to 25.3--17.9\%, below even the artifact pipeline (28.5\%). A raw similarity score cannot separate a topically-relevant-but-unanswerable retrieval from an answerable one. We therefore added two mechanisms with a \emph{semantic} signal. (M1) An \emph{LLM evidence-sufficiency gate} asks the answerer model whether the retrieved verbatim evidence actually contains the answer before responding. (M2) \emph{Answer-then-verify} answers normally, then asks whether the answer is supported by the retrieved evidence and overrides to a refusal if not. Neither escapes the frontier (Table~\ref{tab:abstention_gate}, bottom; matched gpt-4o-mini no-gate 44.6/10.8). M1 only modestly Pareto-improves on the score gate, with $+$3.7pp answerable at a matched $\sim$59\% abstention (25.6\% vs.\ 21.9\%). M2 lands on the score-gate curve (22.3\%/58.1\%). Reaching $\sim$59\% category-5 refusal still costs $\sim$19pp answerable accuracy (44.6$\to$25.6\%; McNemar 146 answerable questions newly refused vs.\ 13 recovered, $p{<}10^{-24}$). The pattern \textbf{replicates on LongMemEval-S} (M1 vs.\ matched no-gate, gpt-4o-mini). Abstention rises 53.3$\to$93.3\%, past naive RAG's 70\%, but answerable accuracy falls 63.2$\to$50.2\% (McNemar 65 lost vs.\ 4 gained, $p{<}10^{-12}$). A similarity gate, an LLM sufficiency judge, and an answer verifier thus all fail to install competent refusal without sacrificing answerable questions. The refusal rate \emph{can} be driven past naive RAG, but not for free. A refusal mechanism that avoids this trade remains open. The verbatim representation lacks competent refusal under the three mechanisms tested.

\begin{table}[t]
\centering
\small
\setlength{\tabcolsep}{3pt}
\begin{tabular}{@{}lcc@{}}
\toprule
 & \textbf{Cat-5} & \textbf{Cat 1--3} \\
\textbf{Refusal mechanism} & \textbf{abstention} & \textbf{answerable} \\
\midrule
\multicolumn{3}{@{}l}{\emph{Score-threshold gate} (no gate: 10.8 / 43.1)} \\
\quad $\tau{=}0.1$ & 55.6 & 25.3 \\
\quad $\tau{=}0.2$ & 59.2 & 21.9 \\
\quad $\tau{=}0.3$ & 63.5 & 17.9 \\
\quad $\tau{=}0.4$ & 65.5 & --- \\
\midrule
\multicolumn{3}{@{}l}{\emph{Semantic refusers} (no gate: 10.8 / 44.6)} \\
\quad M1 sufficiency, $\tau{=}0.0$ & 59.4 & 25.6 \\
\quad M1 sufficiency, $\tau{=}0.1$ & 62.1 & 19.7 \\
\quad M1 sufficiency, $\tau{=}0.2$ & 64.8 & 17.2 \\
\quad M2 answer-then-verify & 58.1 & 22.3 \\
\bottomrule
\end{tabular}
\caption{Refusal mechanisms on the chunk pipeline (LoCoMo, gpt-4o-mini, LLM-Judge \%). \emph{Top:} an evidence-sufficiency gate on the top reranker score; as $\tau$ rises, category-5 refusal climbs but answerable accuracy (categories~1--3) falls monotonically below even the artifact pipeline (the $\tau{=}0.4$ answerable cell is omitted because abstention already saturates). \emph{Bottom:} two stronger \emph{semantic} refusers, an LLM evidence-sufficiency gate (M1, swept over a coarse score pre-filter $\tau$) and answer-then-verify (M2). M1 modestly Pareto-dominates the score gate at matched abstention, but no mechanism escapes the trade-off. Both blocks use gpt-4o-mini no-gate anchors: category-5 refusal 10.8 in both, answerable accuracy 43.1 and 44.6 from two runs (equal within run-to-run noise). The 6.5\% of \S\ref{sec:experiments} is the gpt-4o answerer.}
\label{tab:abstention_gate}
\end{table}

\paragraph{Union storage recovers part of the artifact store's refusals.} The union result of \S\ref{sec:experiments} shows that adding artifacts to verbatim text costs no detectable accuracy on answerable questions. Table~\ref{tab:union_abstention} asks whether it recovers the artifact store's higher refusal rate on category 5. To make the three arms comparable object for object, the chunk arm runs on the union store's chunk objects, which are identical to the standalone chunk store in all ten conversations. The artifact arm runs on the union store's remaining objects, so that union $=$ chunks $+$ artifacts holds exactly. The union refuses on 7.0\% of category-5 questions. Chunks refuse on 4.7\% (McNemar exact, 21 vs.\ 11 discordant, $p{=}0.11$) and artifacts on 10.3\% (16 vs.\ 31, $p{=}0.040$). Chunks and artifacts differ by 5.6pp in this batch (12 vs.\ 37, $p{=}0.0005$), against 8.5pp in the 2026-06 headline batch (6.5\% vs.\ 15.0\%). The union thus recovers 10 of the 25 additional refusals (2.2pp of 5.6pp), and no significant difference from chunks is detected. The 2026-09 union run gives the same ordering. An artifact arm built from an independent extraction run gives 11.7\%. This rate is not distinguishable from the subset arm (16 vs.\ 22, $p{=}0.42$), so the ordering does not depend on one extraction sample. This is consistent with the sparser store refusing more because it retrieves less that can be asserted. It is also consistent with the answerer asserting the trap fact once verbatim text is present, whether or not artifacts sit beside it.

\begin{table}[t]
\centering
\footnotesize
\setlength{\tabcolsep}{4pt}
\begin{tabular}{@{}p{0.52\columnwidth}cc@{}}
\toprule
\textbf{Store} & \textbf{Cat 1--3} & \textbf{Cat 5} \\
\midrule
Chunks (union's chunk subset) & 47.6 & 4.7 \\
Artifacts (union's artifact subset) & 30.3 & 10.3 \\
Union (chunks $+$ artifacts) & 46.4 & 7.0 \\
\midrule
Artifacts, independent extraction & 29.3 & 11.7 \\
\bottomrule
\end{tabular}
\caption{Same-store three-way comparison on LoCoMo (question-micro LLM-Judge \%, gpt-4o answerer; categories 1--3: 699 answerable questions; category 5: 446 unanswerable questions, where the score is the correct-refusal rate). All arms are 2026-11 runs on the same code. This batch sits above the 2026-06 headline batch on categories 1--3 (chunks 43.9, artifacts 28.0, union 42.5), the answerer-side drift documented in Appendix~\ref{sec:limitations_full} and Table~\ref{tab:provenance}. The chunks-minus-artifacts gap here is 17.3pp (172 vs.\ 51 discordant, $p<10^{-15}$). Union vs.\ chunks is not significant (62 vs.\ 71, $p{=}0.49$). A 2026-09 run of the union arm gave 45.9 and 6.3.}
\label{tab:union_abstention}
\end{table}

\section{GraphRAG Baseline Sensitivity}
\label{sec:graphrag_grid}

To check whether the GraphRAG baseline in Table~\ref{tab:locomo_results} is under-tuned, we conducted a 4-axis grid search over the principal configuration knobs of the Microsoft \texttt{graphrag} library. We report the per-cell accuracy on LoCoMo. \textit{Aggregation note.} All GraphRAG numbers in this paper, in the grid tables below and in the main results table, are per-question micro-averages, matching the main-table protocol. (The sweep harness additionally logs per-conversation macro-averages, e.g.\ 12.7\% for the drift winner. Cell rankings are identical under either aggregation.)

\paragraph{Search space.} The search covers four axes. The index-time axes are (i) \textbf{chunk\_size} $\in \{400, 800, 1200\}$ and (ii) \textbf{max\_gleanings} $\in \{1, 2, 3\}$, the number of additional entity-extraction passes per chunk. The query-time axes are (iii) \textbf{community\_level} $\in \{1, 2, 3\}$, the depth in the hierarchical community summary tree, and (iv) \textbf{search\_method} $\in \{\text{local, global, drift}\}$.\footnote{The \texttt{drift} mode is the library's dynamic retrieval-informed token-level filtering method introduced in graphrag 2.x. It most closely approximates a ``hybrid'' search semantically. We exclude \texttt{basic} mode because it falls back to plain vector retrieval and does not exercise the GraphRAG pipeline.} The full Cartesian product spans 81 configurations.

\paragraph{Two-tier protocol.} To make the sweep tractable, we ran a two-tier procedure. \textbf{Tier 1} enumerated the default configuration plus all 1-D excursions ($1 + 4\times2 = 9$ configurations) on a 3-conversation subset of LoCoMo (locomo\_000, \_001, \_002). This tier identified which axes carry signal. \textbf{Tier 2} ran the three Tier-1 leaders on the full 10-conversation LoCoMo benchmark (699 questions, categories 1--3, LLM Judge evaluation) to produce the final tuned-baseline numbers. Total Tier 1 wall time was 6.2\,h. Tier 2 took 10.9\,h, dominated by the slower \texttt{drift} mode (519\,min for one cell).

\paragraph{Tier 1 (3 conversations, 9 cells).} Table~\ref{tab:graphrag_tier1} reports the 1-D sweep. Three findings emerge. (1) chunk\_size and community\_level have only modest effects ($-0.6$ to $-2.2$pp overall vs.\ the default). (2) max\_gleanings = 3 is the strongest index-time signal, lifting open-domain by 14.3pp on the 3-conversation subset. (3) \texttt{global} mode collapses to 5.4\% overall and should not be used for conversational memory. The temporal-accuracy column stays at or below 2.2\% in 8 of 9 cells. Only \texttt{drift} mode raises it, and only to 4.4\% (the full-benchmark comparison follows in Tier 2).

\begin{table}[t]
\centering
\footnotesize
\setlength{\tabcolsep}{2pt}
\begin{tabular}{@{}p{0.435\columnwidth}cccc@{}}
\toprule
\textbf{Configuration} & \textbf{Overall} & \textbf{M-hop} & \textbf{Temp.} & \textbf{O-dom.} \\
\midrule
cs=800, g=3, cl=2, local & \textbf{13.5} & 16.2 & 2.2 & \textbf{52.4} \\
cs=800, g=1, cl=2, local$^\dagger$ & 13.0 & \textbf{18.9} & 2.2 & 38.1 \\
cs=800, g=2, cl=2, local & 12.4 & 16.2 & 2.2 & 42.9 \\
cs=800, g=1, cl=3, local & 12.4 & 16.2 & 2.2 & 42.9 \\
cs=800, g=1, cl=2, drift & 11.9 & 13.5 & \textbf{4.4} & 38.1 \\
cs=400, g=1, cl=2, local & 11.9 & 16.2 & 1.1 & 42.9 \\
cs=1200, g=1, cl=2, local & 11.4 & 13.5 & 2.2 & 42.9 \\
cs=800, g=1, cl=1, local & 10.8 & 13.5 & 2.2 & 38.1 \\
cs=800, g=1, cl=2, global & 5.4 & 6.8 & 0.0 & 23.8 \\
\bottomrule
\end{tabular}
\caption{Tier 1 grid search on a 3-conversation LoCoMo subset, 185 questions (LLM Judge, \%). The default configuration ($^\dagger$) ranks second; the top row is the max\_gleanings = 3 leader. The open-domain column shows substantial variance across cells (23.8--52.4) reflecting the small denominator (21 open-domain questions on the subset).}
\label{tab:graphrag_tier1}
\end{table}

\paragraph{Tier 2 (full LoCoMo, 3 cells).} The Tier-1 leaders were re-run on the full 10-conversation LoCoMo (Table~\ref{tab:graphrag_tier2}). They are max\_gleanings = 3 (overall and open-domain winner), the default (multi-hop winner), and \texttt{drift} mode (sole temporal mover). Two findings follow. (1) The open-domain advantage of max\_gleanings = 3 disappears at scale, regressing to 31.2\%. This is a small-sample artifact of the 3-conversation subset. (2) \texttt{drift} mode emerges as the winner on overall, temporal, and open-domain accuracy (overall 13.0\%, temporal 7.5\%, open-domain 32.3\%), reversing its mid-pack Tier 1 ranking. We therefore report the \texttt{drift} cell as the GraphRAG row in Table~\ref{tab:locomo_results}.

\begin{table}[t]
\centering
\footnotesize
\setlength{\tabcolsep}{2pt}
\begin{tabular}{@{}>{\raggedright\arraybackslash}p{0.435\columnwidth}cccc@{}}
\toprule
\textbf{Configuration} & \textbf{Overall} & \textbf{M-hop} & \textbf{Temp.} & \textbf{O-dom.} \\
\midrule
cs=800, g=1, cl=2, drift (tuned) & \textbf{13.0} & 12.8 & \textbf{7.5} & 32.3 \\
cs=800, g=3, cl=2, local & 11.4 & \textbf{13.8} & 3.4 & 31.2 \\
cs=800, g=1, cl=2, local$^\dagger$ & 11.2 & 13.5 & 3.1 & 31.2 \\
\midrule
Verbatim chunks, no graph$^*$ & \textbf{43.9} & \textbf{35.5} & \textbf{50.2} & \textbf{47.9} \\
Artifacts $+$ graph$^*$ & 28.0 & 22.3 & 31.2 & 34.4 \\
\textit{Gap: tuned GraphRAG $-$ chunks} & --30.9 & --22.7 & --42.7 & --15.6 \\
\bottomrule
\end{tabular}
\caption{Tier 2 grid search on the full LoCoMo benchmark (10 conversations, 699 questions, LLM Judge \%). The \texttt{drift} cell is the tuned ceiling reported in the main results table. $^\dagger$Default configuration. $^*$Reference rows reproduce the main-table numbers (Table~\ref{tab:locomo_results}); all rows are per-question micro-averages (see aggregation note above).}
\label{tab:graphrag_tier2}
\end{table}

\paragraph{Interpretation.} Even after two-tier tuning across four axes (9 unique configurations, 12 runs) spanning index-time and query-time hyperparameters, GraphRAG's temporal accuracy on LoCoMo remains capped at 7.5\%. This is a 42.7pp deficit relative to verbatim chunks and 23.7pp below even the extracted-artifact pipeline. The narrowness of this band across 9 distinct configurations is consistent with a limitation that is \textit{paradigmatic} rather than configurational. GraphRAG's community-level summarization pipeline can abstract away the verbatim timestamps that LoCoMo's temporal questions demand. The May 7, 2023 case study in \S\ref{sec:why_temporal} illustrates this failure mode. The entity ``LGBTQ support group'' survives summarization, but the date does not. Hyperparameters cannot recover information the indexing layer never preserved.

\vspace{1.5em}

\section{Threshold Sensitivity}
\label{sec:threshold_sensitivity}

Table~\ref{tab:threshold_sensitivity} shows that the CogCanvas pass rate on the synthetic multi-hop probe varies by only 10pp (57.5\%--67.5\%) across threshold configurations. The sweep is a December pilot (10-conversation subset, single compression at turn 40 rather than the main rolling protocol, gpt-4o-mini answerer). Its rows are internally comparable, but the absolute level is not commensurate with the 81.0\% of Table~\ref{tab:multi_hop_results}.

\begin{table}[t]
\centering
\small
\begin{tabular}{@{}lccc@{}}
\toprule
\textbf{Config} & \textbf{Ref.} & \textbf{Causal} & \textbf{Pass} \\
\midrule
Low & 0.3 & 0.25 & 62.5\% \\
\textbf{Default} & \textbf{0.5} & \textbf{0.45} & \textbf{67.5\%} \\
High & 0.7 & 0.6 & 60.0\% \\
Very High & 0.8 & 0.7 & 57.5\% \\
\bottomrule
\end{tabular}
\caption{Threshold sensitivity on the synthetic multi-hop probe (10-conversation pilot, pre-rolling protocol, gpt-4o-mini answerer; rows internally comparable only). Performance is stable across values.}
\label{tab:threshold_sensitivity}
\end{table}

\section{Extended Related Work}
\label{sec:related_work_full}

This section provides detailed discussion of related work summarized in Section 2.

\subsection{Context Window Management}

The finite context window of transformer-based LLMs has motivated research into efficient context utilization. Sparse attention mechanisms \citep{child2019sparsetransformer, beltagy2020longformer} reduce the quadratic complexity of self-attention. They enable longer sequences but do not remove the capacity constraint. Retrieval-augmented approaches \citep{borgeaud2022retro, izacard2022atlas} maintain external datastores that are queried at inference time. These systems typically assume static document collections rather than dynamically evolving conversational state. Recent work on context extension through positional interpolation \citep{chen2023extending} pushes the practical context limit but does not eliminate the trade-off between context length and computational cost. Our approach is orthogonal to these efforts. Rather than extending raw context capacity, we focus on \textit{what} to preserve when compression becomes necessary.

\subsection{Memory-Augmented Language Models}

\looseness=-1 MemGPT~\citep{packer2023memgpt} pioneered a hierarchical memory system inspired by operating system design. It introduced virtual context management, in which LLMs perform explicit memory management operations (moving data between ``fast'' and ``slow'' memory tiers). The original MemGPT project has since evolved into Letta,\footnote{\url{https://letta.ai}} a production-ready framework that extends the core architecture with filesystem-based memory storage, PostgreSQL persistence, and enhanced tool integration. Letta's recent LoCoMo evaluation achieved 74.0\% accuracy using GPT-4o mini~\citep{letta2024memory} by storing conversation histories as files with semantic search capabilities. This result shows that stateful, database-backed memory systems can perform well, but the approach requires persistent infrastructure (PostgreSQL, Docker) that introduces deployment complexity. This architectural approach represents a different paradigm from our stateless, inference-time retrieval.

\looseness=-1 Several concurrent approaches explore alternative memory architectures. MemoryBank~\citep{zhong2024memorybank} incorporates a memory updating mechanism inspired by the Ebbinghaus Forgetting Curve, allowing selective memory preservation based on temporal decay and significance. LongMem~\citep{wang2023longmem} employs a decoupled architecture with a frozen backbone LLM as memory encoder and a side-network for retrieval, caching attention key-value pairs from previous segments. ReadAgent~\citep{lee2024readagent} takes a human-inspired approach, compressing long documents into ``gist memories'' while retaining the ability to look up original passages when details are needed. These methods address the memory problem in different ways. They typically either require architectural modifications or focus on document comprehension rather than dynamic conversational state.

More recent systems improve memory-augmented LLMs through dedicated optimization. EverMemOS~\citep{evermemos2025} reports 92.3\% on LoCoMo through categorical memory extraction (situational context, semantics, user profiling) and MemCell atomic storage with rich metadata (timestamps, source, tags, relational links). This is the highest self-reported LoCoMo figure we are aware of; it is measured under a different stack and we did not reproduce it.

In contrast, the concurrent EMem~\citep{zhou2025simple} reaches 78.0\% on LoCoMo (GPT-4o-mini answerer) while remaining entirely training-free. It decomposes each session into enriched Elementary Discourse Units, which are short, near-verbatim event propositions carrying normalized entities and source-turn attributions. It retrieves over these units and explicitly aims to preserve information ``in a non-compressive form.'' Its strong result with a representation near the verbatim end of the fidelity spectrum is the outcome our account in \S\ref{sec:discussion} predicts. We treat it as the near-verbatim data point in that reconciliation rather than as a competing system claim.

Two further neighbors need explicit positioning. \citet{zeng2024structural} is the closest prior comparison. Holding the model and embedder fixed, it sweeps chunks, knowledge triples, atomic facts, summaries, and mixed stores across four tasks and six datasets (including LoCoMo). It finds that different structures suit different tasks, that chunk-based stores excel on lengthy-context dialogue understanding, and that mixed stores are the most resilient. Our results agree with that ordering where the studies overlap. What we add is the controlled isolation of the \emph{extraction-vs-source-text} premise on long-conversation benchmarks. This isolation comprises paired McNemar statistics over a single fixed retrieval--rerank--reasoning pipeline, six confound controls, mechanism probes (store-level survival, keyword-recall audit), decontaminated-prompt and session-granularity reruns, and per-correct-answer cost accounting. A further addition is the negative-result framing, in which the replacement premise itself fails. Zep~\citep{rasmussen2025zep} reports strong LongMemEval results with the Graphiti temporal knowledge graph. Its lowest tier is an \emph{episode subgraph} that retains the raw interaction data beneath the entity and community layers. Zep therefore instantiates structure-annotating-text rather than structure-replacing-text, which is consistent with the fidelity ordering of \S\ref{sec:discussion}. Its absolute numbers come from a different answerer and harness and are not commensurable with our tables.

Heavily-optimized systems such as EverMemOS occupy a different design space. They report results under their own retrieval and answering stacks, which we did not reproduce. Such systems suit high-stakes deployments where maximum accuracy justifies the engineering investment. They are less suitable for practitioners who need a drop-in component.

Our study deliberately occupies the \textit{training-free, stateless} corner of this design space. It is the setting where the stored representation can be swapped in isolation, without fine-tuning or infrastructure confounds. We do not compare absolute numbers against systems evaluated on different answerer models and evaluation stacks (EverMemOS, EMem). Our contribution is the controlled within-pipeline comparison. It finds that the replacement premise common to lossy-extraction designs does not survive isolation. Verbatim chunks dominate extracted artifacts, and the artifact pipeline does not beat naive RAG (Tables~\ref{tab:locomo_results} and~\ref{tab:longmemeval_results}).

\subsection{Graph-Based Knowledge Organization}

\looseness=-1 GraphRAG \citep{edge2024graphrag} and related approaches \citep{yasunaga2021qagnn, zhang2022greaselm} use graph structures to organize and retrieve knowledge. These systems excel at capturing relational information and enabling multi-hop reasoning over structured knowledge bases. However, existing graph-based methods are predominantly designed for \textit{static document collections}, that is, corpora that are processed offline and queried at inference time. The dynamic, incremental nature of conversational knowledge poses distinct challenges. Stored items must be produced in real time, any graph structure must evolve with each turn, and retrieval must balance recency with relevance. The artifact pipeline under test meets these requirements through online extraction and incremental graph construction. Our ablation shows that meeting them does not rescue the representation. The same incremental machinery over verbatim chunks performs better at every question type (Table~\ref{tab:locomo_results}).

\subsection{Conversation Summarization}

\looseness=-1 Summarization-based approaches \citep{tang2021convosumm, chen2021dialogsum} offer an intuitive solution to context overflow. They compress older turns into concise summaries that preserve essential information within reduced token budgets. Recent work has explored hierarchical summarization \citep{wu2021recursively}, where summaries are themselves summarized as conversations extend. Summarization captures high-level narrative flow, but it operates on a lossy compression principle. Our experiments show that even an LLM-based summarizer loses specific details (Table~\ref{tab:locomo_results}). Typed artifact extraction, despite its verbatim quote fields, only \emph{mitigates} this loss (\S\ref{sec:discussion}).

\section{Limitations in Full}
\label{sec:limitations_full}

This appendix gives the complete account of the limitations summarized in \S\ref{sec:limitations}, with the supporting controls and statistics.

\textbf{Scope of the negative result.} Our finding concerns extraction that \emph{replaces} source text with distilled artifacts. We tested one artifact schema within one pipeline family. The schema's extraction runs as an initial pass plus one gleaning pass (\S\ref{sec:methodology}). An ablation found that removing the gleaning pass leaves LoCoMo accuracy statistically unchanged (Appendix~\ref{sec:error_analysis}). The second pass therefore neither drives nor masks the gap, and we did not tune it per-dataset. Extractor strength is unlikely to be the missing ingredient either. Swapping the gpt-4o-mini extractor for gpt-4o leaves both artifact accuracy and the chunk advantage statistically unchanged (Appendix~\ref{sec:error_analysis}). We also tested the strongest candidate for an escape, near-verbatim and provenance-preserving event units~\citep{zhou2025simple}. Reproduced inside the fixed pipeline, the EDU store lands between typed artifacts and chunks ($30.2{<}36.3$--$36.5{<}47.4$; both gaps significant, saturated in $k$; Appendix~\ref{sec:emem_control}). This is consistent with the fidelity ordering, and the EDU store still trails raw text. Its published full-system numbers additionally rely on LLM-filtered, graph-propagated retrieval, which our swap holds fixed. We cannot rule out that some untested design escapes the pattern.

\textbf{Extraction's one win does not generalize.} Artifacts win in one setting, our synthetic multi-hop probe (Table~\ref{tab:multi_hop_results}, 81.0 vs.\ 55.5). By construction this is extraction's best case, since every planted fact is pre-shaped into exactly one artifact. The inversion does not transfer to the benchmarks' multi-hop questions. Verbatim chunks still lead on LoCoMo's multi-hop category (35.5 vs.\ 22.3; $N{=}282$, $+13.1$pp, cluster-bootstrap 95\% CI $[10.2, 15.9]$, McNemar $p{=}2.2{\times}10^{-5}$). They also lead on LongMemEval multi-session reasoning (62.8 vs.\ 34.7; $N{=}121$, $+28.1$pp, 95\% CI $[18.2, 38.0]$, $p{=}1.2{\times}10^{-6}$). The artifact advantage therefore does not carry over to the multi-hop questions of LoCoMo and LongMemEval. Each of the two benchmark multi-hop cells is individually significant, so the reversal does not stem from small samples.

\textbf{What the controls do and do not rule out.} Our controls do not support rescue by the remedies listed below. The first three are \emph{(i)}~graph expansion, \emph{(ii)}~chunk-overlap artifacts, and \emph{(iii)}~evidence-budget asymmetry (the budget-matched control recovers 1.2pp of the 15.9pp gap). The controls also cover \emph{(iv)}~lexical-retrieval bias toward source text (the dense-only control leaves a 14.3pp gap) and \emph{(v)}~the placement of extraction windows (the session-granularity control leaves a 14.7pp gap). For \emph{(vi)}~coarse-vs-fine granularity, a sentence-verbatim store at the chunk token budget trails chunks by 6.6pp, under half of the gap (Appendix~\ref{sec:sentence_control}). The remaining two are \emph{(vii)}~baseline under-tuning (Appendix~\ref{sec:graphrag_grid}) and \emph{(viii)}~cost arguments (\S\ref{sec:cost_analysis}). Within the schema we tested two extraction granularities, 40-turn batches and whole sessions, with statistically indistinguishable results (\S\ref{sec:experiments}). Segment-level construction~\citep{pan2025secom} and incremental update schemes~\citep{chhikara2025mem0} remain untested as extraction regimes for this schema.

\textbf{Benchmarks and language.} Our two headline benchmarks are English chat-style QA. LoCoMo contributes 10 conversations / 699 questions in the main protocol plus 1,287 held-out category-4/5 questions, and LongMemEval-S contributes 500 questions. A native-Chinese probe (PerLTQA, Appendix~\ref{sec:perltqa}) reproduces the ordering more starkly still, on dialogues that are likewise model-generated. Memory uses beyond QA (proactive recall, personalization, summarization-style review) remain untested.

\textbf{One retrieval stack; neither the answerer nor its vendor is load-bearing.} All pipelines share one embedding and reranker family (bge-m3, bge-reranker-v2-m3) and one answerer (gpt-4o). The ordering is not specific to that retriever. We re-ran the chunks-vs-artifacts comparison with the answerer fixed at gpt-4o-mini under a held-out dense embedder (OpenAI text-embedding-3-small) and a sparse lexical retriever (BM25). The chunk advantage stays essentially unchanged. Across all three retriever families, chunks score 43.1--43.8\% and artifacts 28.5--29.9\%, a gap of $+$13.6 to $+$14.7pp (Appendix~\ref{sec:retriever_robustness}). The answerer is also not what drives the result. Swapping gpt-4o for the much weaker gpt-4o-mini costs verbatim chunks only 2.0pp (43.9\% to 41.9\%). Re-running the controlled fidelity dial (\S\ref{sec:discussion}) under the gpt-4o-mini answerer reproduces the same monotone collapse (accuracy $43.5{\to}30.8{\to}21.2{\to}12.2{\to}8.4$ as retain goes $1.0{\to}0.2$; Figure~\ref{fig:fidelity_dial}). The fidelity$\to$accuracy ordering is therefore not an artifact of the strong model. Nor is it an artifact of \emph{random} token deletion. Sweeping SeCom's own extractive compression reproduces the same monotone collapse (accuracy $47.4{\to}36.9$ over $1.0{\to}0.5$; Appendix~\ref{sec:secom_dial}), and under aggressive compression accuracy falls below fixed chunks. The ordering thus also holds when a real extractor, instead of random deletion, removes fidelity. It also survives a change of vendor for both LLM roles. We replaced gpt-4o and gpt-4o-mini with answerer \textsf{claude-sonnet-4-6} and extractor \textsf{claude-haiku-4-5}, holding everything else fixed. On the same 699 questions, chunks score 53.9\% and artifacts 33.2\% (gap 20.7pp; McNemar 199 vs.\ 54 discordant, $p{=}1.1{\times}10^{-20}$; Appendix~\ref{sec:vendor_control}). Both stores improve with the stronger answerer, and the gap widens. We nonetheless keep every headline comparison same-model for cleanliness. The GraphRAG row is the one exception, because its indexing is coupled to gpt-4o-mini; the matched-model control in Table~\ref{tab:locomo_results} bounds it.

\textbf{Run-to-run variance.} Every table entry is a single run at temperature 0. Paired McNemar tests and cluster-bootstrap CIs capture question- and conversation-level sampling uncertainty. To bound API nondeterminism directly, we re-ran both headline LoCoMo cells three times end-to-end, with fresh extraction caches for the artifact arm. Within-batch spread is negligible (chunks 47.4/47.4/47.5, artifacts 30.2/30.3/30.6, paired gap $+$17.2/$+$17.0/$+$16.9pp). Across months the unpinned \texttt{gpt-4o} endpoint drifts. Both arms shift upward in parallel against the 2026-06 headline batch (43.9/28.0). Re-judging the stored June answers with the same judge tier reproduces the published numbers within 0.2pp on LoCoMo (44.1/28.2) and 0.8pp on LongMemEval (67.4/46.2 vs.\ 67.4/45.4). The drift therefore comes from the answerer endpoint, since the judge and protocol reproduce the stored results. The representation gap is, if anything, larger under the newer snapshot (17.0 vs.\ 15.9pp). Appendix~\ref{sec:provenance} consolidates every batch assignment.

\textbf{The winning representation has a real weakness.} Verbatim chunks trail naive RAG on LongMemEval abstention by 23.3pp (46.7\% vs.\ 70.0\%; $n{=}30$, McNemar exact $p{=}0.039$). The better-powered evidence is LoCoMo's 446 adversarial questions, where chunks reach only 6.5\% against the artifact pipeline's 15.0\% ($p{=}1.9{\times}10^{-6}$). Handed plausible but irrelevant evidence, the chunk pipeline answers when it should refuse. The result is a negative finding about extraction and makes no claim that chunks are universally optimal. Relatedly, we validated the category-5 abstention judge against an independent lexical detector (92.4--97.5\% agreement), a context-isolated blinded LLM annotator, and an independent human annotator. The latter two each reach 50/50 agreement ($\kappa{=}1.0$; Appendix~\ref{sec:judge_validation}). \emph{Repairing} abstention is also harder than it looks. We added an evidence-sufficiency gate that refuses when the top reranker score falls below a threshold $\tau$. Under the gpt-4o-mini answerer used for this sweep, the gate lifts category-5 refusal from 10.8\% to 55.6--65.5\% as $\tau$ rises from 0.1 to 0.4. Answerable accuracy on categories 1--3 collapses in lockstep (43.1\% down to 25.3--17.9\%), dropping below even the artifact pipeline. No threshold gains abstention without a large loss of answerable questions. Two stronger \emph{semantic} refusers fare no better (an LLM evidence-sufficiency gate and an answer-then-verify check). On LongMemEval-S they can drive abstention from 53\% \emph{past} naive RAG's 70\% (to 93\%), but only at a $-$13pp answerable-accuracy cost. On LoCoMo they only match or modestly improve the score-gate frontier (Appendix~\ref{sec:retriever_robustness}). The verbatim representation therefore lacks competent refusal under the three repair mechanisms we tested. We found no tuning that repairs it. The refusal rate \emph{can} be lifted past naive RAG, but at a cost. A refusal mechanism that does not trade answerable accuracy remains open.

\textbf{Training-free scope.} Stateful and infrastructure-backed systems (e.g., Letta at a self-reported 74.0\%, EverMemOS at 92.3\%) sit outside our controlled comparison. Their numbers come from different answerer models and evaluation stacks and are not commensurable with Table~\ref{tab:locomo_results} (Appendix~\ref{sec:stateful_discussion}). We therefore treat every such figure strictly as an \emph{external anchor}. None appears as a baseline in our tables, and we make no head-to-head claim against it. Only the within-pipeline rows of Tables~\ref{tab:locomo_results}--\ref{tab:longmemeval_results} constitute a controlled representation comparison, since they share one retriever, reranker, answerer, and judge. Notably, the strongest of these external systems are themselves extraction- or classification-based. A controlled probe indicates the ordering persists under \emph{statefulness}. Inside a stateful agent (a MemGPT-style core-memory-plus-archival loop with incremental compression), we swapped \emph{only} the archival representation from verbatim turns to extracted artifacts and held the stateful machinery fixed. The swap costs 10.9 points on LoCoMo (24.0\% vs.\ 13.2\%; McNemar exact 100 vs.\ 24 discordant, $p{<}10^{-11}$). It reproduces the direction of our stateless headline on every question type (Appendix~\ref{sec:stateful_probe}). Whether the ordering also survives fine-tuning remains untested.

\end{document}